\documentclass[journal]{IEEEtran}
\usepackage{cite}
\usepackage{times}
\usepackage{epsfig}
\usepackage{epstopdf}
\usepackage{graphicx}
\usepackage{amsmath}
\usepackage{amssymb}
\usepackage{makecell}
\usepackage{multirow}
\usepackage{threeparttable}
\usepackage[pagebackref=true,breaklinks=true,letterpaper=true,colorlinks,bookmarks=false]{hyperref}
\usepackage{graphicx}
\usepackage{algorithm}  
\usepackage{algorithmic} 
\usepackage{caption}
\usepackage{subfigure}
\begin{document}
\title{iDARTS: Improving DARTS by Node Normalization and Decorrelation Discretization}

\author{Huiqun Wang, Ruijie Yang, Di Huang, \textit{Senior Member, IEEE},  and Yunhong Wang, \textit{Fellow, IEEE}

}% <-this % stops a space

%\ifCLASSOPTIONpeerreview
%\markboth{Journal of \LaTeX\ Class Files}%

\maketitle


\begin{abstract}
Differentiable ARchiTecture Search (DARTS) uses a continuous relaxation of network representation and dramatically accelerates Neural Architecture Search (NAS) by almost thousands of times in GPU-day. However, the searching process of DARTS is unstable, which suffers severe degradation when training epochs become large, thus limiting its application. In this paper, we claim that this degradation issue is caused by the imbalanced norms between different nodes and the highly correlated outputs from various operations. We then propose an improved version of DARTS, namely iDARTS, to deal with the two problems. In the training phase, it introduces node normalization to maintain the norm balance. In the discretization phase, the continuous architecture is approximated based on the similarity between the outputs of the node and the decorrelated operations rather than the values of the architecture parameters. Extensive evaluation is conducted on CIFAR-10 and ImageNet, and the error rates of 2.25\% and 24.7\% are reported within 0.2 and 1.9 GPU-day for architecture search respectively, which shows its effectiveness. Additional analysis also reveals that iDARTS has the advantage in robustness and generalization over other DARTS-based counterparts.

\end{abstract} 


\IEEEpeerreviewmaketitle

\section{Introduction}
\label{S.Intro}
Despite the great success of neural networks in a large number of areas, the design of neural architectures is still a tedious task which needs rich experience and repeated adjustment of human experts. Many efforts have been made to automate this Neural Architecture Search (NAS) process. Some studies \cite{DBLP:conf/iclr/BakerGNR17,DBLP:conf/iclr/ZophL17,DBLP:conf/cvpr/ZhongYWSL18,DBLP:conf/cvpr/ZophVSL18} formulate it as a Reinforcement Learning (RL) problem. The agents build neural architectures based on specific search spaces, and their rewards are estimated in terms of corresponding accuracies. Some methods \cite{DBLP:conf/icml/RealMSSSTLK17, DBLP:conf/iccv/XieY17, DBLP:conf/ijcai/SuganumaSN18, DBLP:conf/iclr/LiuSVFK18, DBLP:conf/iclr/ElskenMH18, DBLP:conf/pkdd/Wistuba18, DBLP:conf/aaai/RealAHL19} use Evolutionary Algorithms (EA) to search effective solutions and form the architectures by mutation and recombination of different populations. There are also some attempts to explore improvements with Sequential Model-based Optimization (SMBO) \cite{DBLP:conf/eccv/DongCJWS18,DBLP:conf/eccv/LiuZNSHLFYHM18}, Bayesian optimization \cite{DBLP:conf/nips/KandasamyNSPX18}, and Monte Carlo Tree Search (MCTS) \cite{DBLP:journals/corr/NegrinhoG17}. These methods deliver better models than the handcrafted ones; however, they demand a huge amount of computational cost.

\begin{figure*}[!htb]

\begin{center}
\end{center}
	\includegraphics[width = 1.0\linewidth]{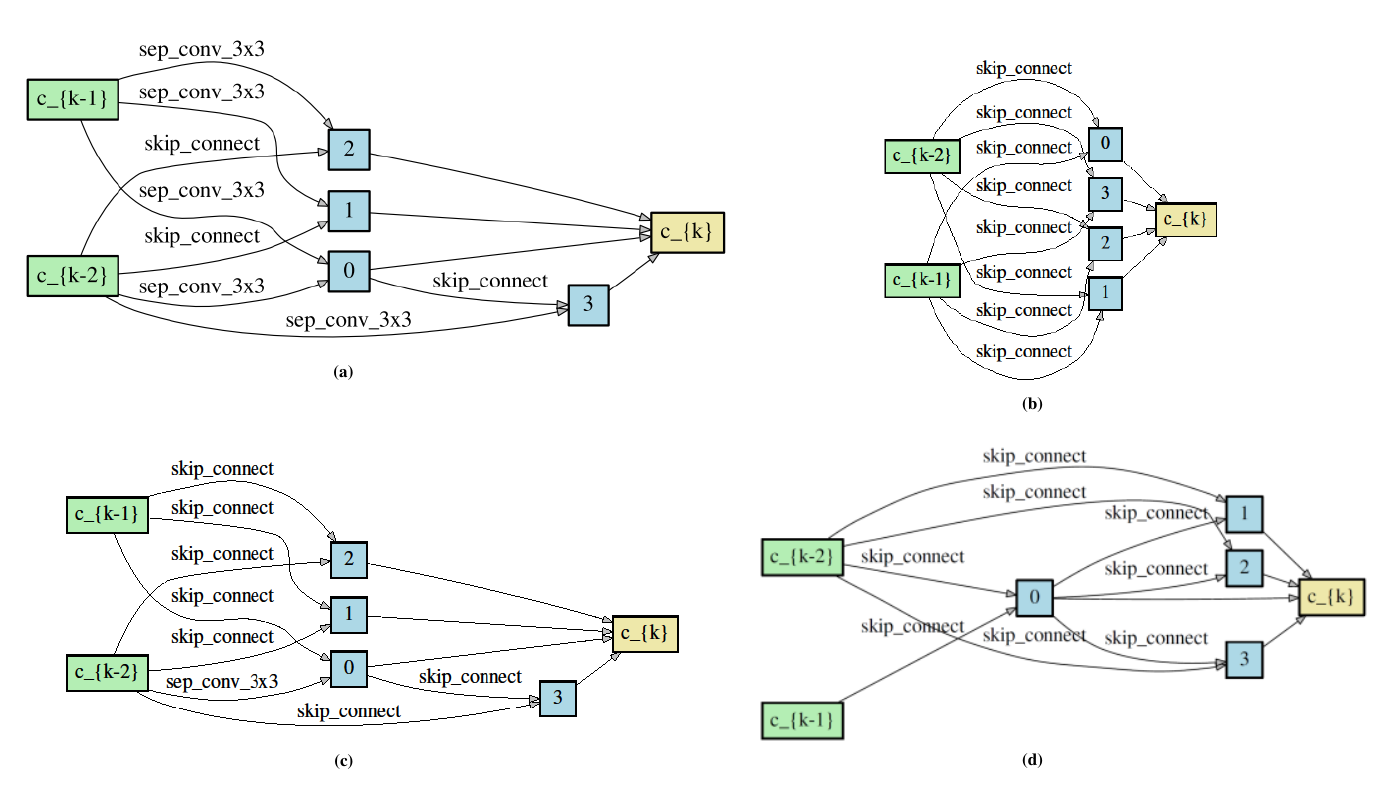}
   \caption{Demonstration of the degradation issue in DARTS. (a) is the selected architecture by DARTS in the original search space when the epoch number is set at 50; (b) and (c) are the ones selected by with first-order (DARTS-1st) and second-order (DARTS-2nd) gradient estimation respectively in a small search space; and (d) is the result when the epoch number is set at 200.}
\label{fig:fig0}
\end{figure*}

Recently, one-shot methods \cite{DBLP:conf/iclr/BrockLRW18, DBLP:conf/iclr/XieZLL19, DBLP:conf/iclr/ShinPS18, DBLP:conf/eccv/AhmedT18} have been investigated for fast NAS, where a single neural network is trained during the searching process, and the final architecture is derived as the solution to the specific optimization task. Differential ARchitecture Search (DARTS) \cite{DBLP:conf/iclr/LiuSY19} is a particularly popular instance and it relaxes the search space to be continuous by introducing architecture parameters into the gradient descent procedure. DARTS has two main phases: (1) in training, it simultaneously optimizes the architecture parameters and  model weights, and (2) in discretization, it approximates the neural architecture according to the values of the architecture parameters. Benefiting from the relaxation on the discrete search space, DARTS reports comparable performance at a much higher speed. Unfortunately, it suffers severe degradation, \emph{i.e.}, when the number of searching epochs becomes larger or the search space changes \cite{DBLP:conf/iclr/ZelaESMBH20}, the \emph{skip-connect} operation tends to dominate the final architecture. As shown in Fig. \ref{fig:fig0}, DARTS finds a sound architecture by the first 50 epochs in the original search space, but converges to a poor one (only \emph{skip-connects}) with a dramatic error increase as the epoch number reaches 200 or in different search spaces.

To address the issue aforementioned, \cite{DBLP:journals/corr/abs-1909-06035,DBLP:journals/corr/abs-1909-09656} adopt early stop to terminate searching before degradation occurs. \cite{DBLP:journals/corr/abs-1910-11831} amends the approximation in updating architecture parameters by gradient descent. \cite{DBLP:journals/corr/abs-1904-12760} manually intervenes the architecture based on a set of pre-defined rules. Such strategies do alleviate the degradation issue to some extent, but they are criticized for obvious drawbacks. Early stop and manual intervention bring in more hyper-parameters, and it still remains an open question to achieve appropriate parameters, making them not easy to be generalized to other scenarios (\emph{e.g.} different search spaces or tasks). In addition, due to the estimation of second-order gradients, approximation amendment incurs high computational cost and unexpected model instability, which impedes its further application.

In this paper, we claim that the degradation of DARTS is caused by the mechanisms in training and discretization. For the former, the norms of different nodes are usually not of the same scale, and this imbalance tends to highlight the importance of the nodes linked to the input.  DARTS thus builds more \textit{skip-connects} and converges to a shallow architecture. For the latter, as connections in the auxiliary one-shot model are highly redundant, the outputs of these operations are usually  correlated. In this case, it is probable to select the operations that generate redundant feature maps but with larger architecture parameters. Besides, some operations with fewer parameters than convolutions, \emph{e.g.,} \textit{skip-connects} and \textit{poolings}, are prone to gain larger architecture weights in the gradient backward process \cite{DBLP:journals/corr/abs-1904-12760}. Both the facts lead to shallow architectures full of \textit{skip-connects}.

Then, we propose an improved version of DARTS, namely iDARTS, which introduces node normalization to maintain the balance between norms of different nodes in training and discretizes the continuous architecture based on the similarity between the outputs of the cell and the decorrelated operations rather than the values of the architecture parameters. We conduct extensive evaluation on two popular benchmarks and achieve the competitive top-1 test errors of 2.25\% and 24.7\% with the searching time at 0.2 and 1.9 GPU-day on CIFAR-10 and ImageNet, respectively. Furthermore, we show the advantage of iDARTS in robustness and generalization for various search spaces and large training epochs by its superior performance in comparison with the major DARTS-based counterparts. 

The remainder of the paper is organized as follows.  Section \ref{S.RW} introduces the methods related to NAS, in particular DARTS-based ones. In Section \ref{S.Method}, we discuss the limitations of the original DARTS and present the solutions to them in detail. The experimental results are described and analyzed in Section \ref{S.EXP}. Finally, we conclude the paper in Section \ref{S.CON}.

\section{Related Work}
\label{S.RW}
In this section, we briefly review the major NAS methods in the literature.

RL based methods represent the network architecture as a variable-length string or a direct acyclic graph. NAS is then regarded as a process of sequential decision where an agent is made to learn how to design the neural architecture. Various agent policies and reward functions are employed to achieve performance gains. NAS-RL \cite{DBLP:conf/iclr/ZophL17} attempts to model an RNN-based controller  whose predictions are considered as actions used to construct the neural architecture. NASNet \cite{DBLP:conf/cvpr/ZophVSL18} adopts the proxy dataset and a carefully designed search space to reduce the computational cost. MetaQNN \cite{DBLP:conf/iclr/BakerGNR17} and Block-QNN \cite{DBLP:conf/cvpr/ZhongYWSL18} both apply the Q-learning paradigm with the epsilon-greedy exploration strategy while the latter chooses to construct the network in a block-wise manner. ENAS \cite{DBLP:conf/icml/PhamGZLD18} proposes a parameter sharing strategy among candidate architectures to accelerate the searching process. 

% MetaQNN \cite{DBLP:conf/iclr/BakerGNR17} uses Q-learning to record the agent rewards and select the operator based on Markov decision process. Block-QNN \cite{DBLP:conf/cvpr/ZhongYWSL18} also employ Q-learning paradigm with epsilon-greedy exploration strategy but it constructs the network in a block-wise manner, successfully narrowing down the search spaces.

On the other hand, some studies try to build neural networks based on EA to avoid human intervention as much as possible. Large-scale Evolution \cite{DBLP:conf/icml/RealMSSSTLK17} initializes a large population based on the simplest network structure, and the operations such as reproduction, mutation and selection are designed to obtain the best architecture. GeNet \cite{DBLP:conf/iccv/XieY17} encodes the network architecture to a fixed-length binary string rather than the graph-based forms. During iteration, the modified evolutionary operations are conducted on those binary strings to generate new individuals, where the most competitive one is taken as the final result. Hierarchical-EAS \cite{DBLP:conf/iclr/LiuSVFK18} presents a hierarchical genetic representation to imitate the human design procedure. Separable convolutions are involved to reduce the spatial resolution, keeping the structure consistency between normal and reduction cells. AmoebaNet \cite{DBLP:conf/aaai/RealAHL19} introduces an age property in EA to favor younger architectures and simplifies mutations in the search space of NASNet, reaching comparable performance to RL-based methods with lower computational cost.

Other optimization techniques are also exploited for this issue. PNAS \cite{DBLP:conf/eccv/LiuZNSHLFYHM18} progressively searches for the complex block architecture based on the SMBO algorithm, while DPP-Net \cite{DBLP:conf/eccv/DongCJWS18} considers the QoS (Quality of Service) and hardware requirments in the objective function. NASBOT \cite{DBLP:conf/nips/KandasamyNSPX18} develops a distance metric via optimal transport and adopts Gaussian process based Bayesian Optimal (BO) for architecture search. DeepArchitect \cite{DBLP:journals/corr/NegrinhoG17} designs a tree-structured search space and traverses it by MCTS. Despite promising results, such solutions bear quite high computational cost (\emph{e.g.} thousands of GPU-day to learn a classification model on CIFAR-10). 

Recently, one-shot methods have received increasing attention, aiming to reduce calculation and accelerate searching. SNAS \cite{DBLP:conf/iclr/XieZLL19}, DSNAS \cite{DBLP:conf/cvpr/HuXZLSLL20} and PVLL-NAS \cite{DBLP:conf/icml/LiDWX20} apply re-parameterization tricks to train neural operation and architecture distribution in the same round of back propagation. SMASH \cite{DBLP:conf/iclr/BrockLRW18} introduces HyperNet \cite{DBLP:conf/iclr/HaDL17} to generate weights for candidate architectures rather than training them from scratch, reaching a faster evaluating speed. DAS \cite{DBLP:conf/iclr/ShinPS18} starts to relax the discrete neural network architecture to a contiguously differentiable form, searching for the best hyperparameters of convolution layers as well as the weights. MaskConnect \cite{DBLP:conf/eccv/AhmedT18} explores the possibility to directly optimize the connectivity of modules with a modified version of gradient descent.

By combining contiguously differentiable architecture parameters and flexible connections, DARTS \cite{DBLP:conf/iclr/LiuSY19} reduces the searching cost by a large margin. It introduces concrete architecture parameters to represent the dense connections within the blocks, and the architecture parameters and model weights can thus be optimized simultaneously. Following the gradient based paradigm of DARTS, a number of investigations advance it to better fit more practical situations. P-DARTS \cite{DBLP:journals/corr/abs-1904-12760} progressively narrows down the gap between the search space and the target space. PC-DARTS \cite{DBLP:journals/corr/abs-1907-05737} increases the efficiency of searching by sampling a small part of the super-net to reduce the redundancy. GDAS \cite{DBLP:conf/cvpr/DongY19} presents a differentiable architecture sampler to produce subgraphs in the direct acyclic graph during training and only the sampled subgraph is optimized  in iteration, thus alleviating the searching cost. Gold-NAS \cite{DBLP:journals/corr/abs-2007-03331} and Proxyless-NAS \cite{DBLP:conf/iclr/CaiZH19} take the resource constraints into consideration to deliver a better balance between computational cost and model accuracy. 

Even though those methods make large progress to optimize DARTS, a severe problem remains unsolved.  DARTS encounters a degradation issue that the model is not stable when searching epochs become larger or the search space changes, prone to converge to a shallow architecture full of \textit{skip-connect} operations. \cite{DBLP:conf/iclr/ZelaESMBH20,DBLP:journals/corr/abs-1909-06035} decide to terminate searching in advance based on the eigenvalues or the composition of the architecture. P-DARTS \cite{DBLP:journals/corr/abs-1904-12760} and Amended-DARTS \cite{DBLP:journals/corr/abs-1910-11831} intervene the gradients or the architectures according to pre-defined rules. However, the degradation issue is only alleviated rather than eliminated, leaving much room for improvement.

In this paper, we claim that the degradation issue comes from both the training and discretization phases, and the proposed iDARTS deals with it by node normalization and decorrelation discretization respectively. Thanks to these two strategies, we significantly ameliorate the accuracy and robustness of DARTS.

\section{Method}
\label{S.Method}
\subsection{Preliminaries}
\begin{figure}[!htb]
\begin{center}
\end{center}
	\includegraphics[width = 1.0\linewidth]{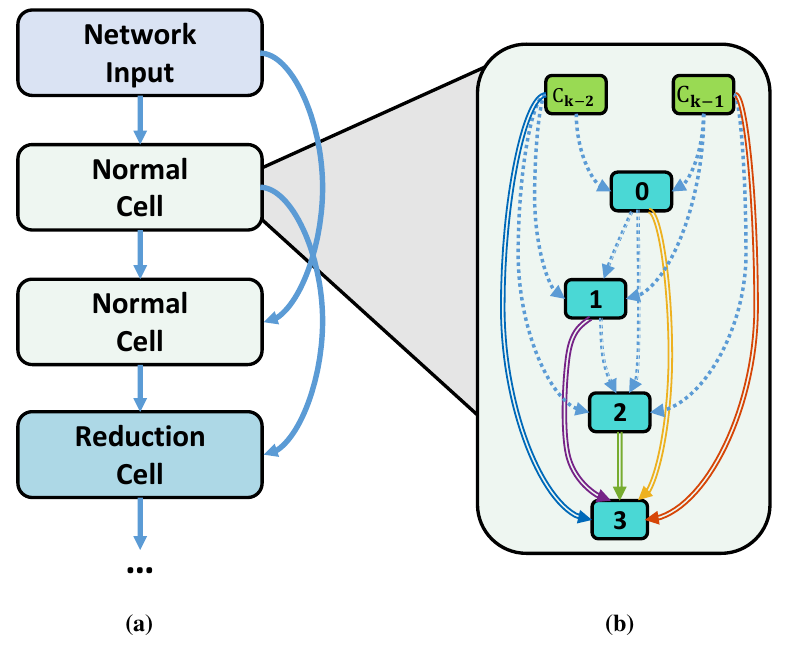}
   \caption{Visualization of DARTS: (a) the super-net stacked by several normal and reduction cells, and the inputs of each cell come from the previous two layers; and (b) the dense connections within a normal cell.}
\label{fig:darts}
\end{figure}

In DARTS, the architecture search is performed on a super-net. As shown in Fig.  \ref{fig:darts} (a), the super-net consists of a set of a pre-defined number of layers. Each layer has a normal cell for feature encoding or a reduction cell for feature down-sampling. There are several $nodes$ in a single cell, representing the mid-layer results. All these nodes are densely connected as shown in Fig. \ref{fig:darts} (b). Each cell outputs the concatenation of the nodes. The $edge$ between each pair of nodes is a mix of a set of candidate neural operators in the search space, \emph {i.e.,} \textit{skip-connects}, \textit{poolings}, or \textit{convolutions}. The output of the edge between node pair $(i,j)$ is treated as a softmax over all possible operations, which can be written as:

\begin{equation}
\begin{aligned}
\bar{o}^{(i,j)}(x)=\sum_{o\in \mathcal{O}}\frac{exp(\alpha_o^{(i,j)})}{\sum_{o'\in \mathcal{O}} exp(\alpha_{o'}^{(i,j)})}o(x)
\label{Formula.1}
\end{aligned}
\end{equation}
where $o$ denotes the operations in operation set $\mathcal{O}$ and $\alpha_o^{(i,j)}$ represents the architecture parameter for node pair $(i,j)$ corresponding to operation $o$. The output of node $j$ is the summarization of those of all its connected edges, written as:

\begin{equation}
\begin{aligned}
y^{(j)}(x)=\sum_{i<j}\bar{o}^{(i,j)}(x)
\label{Formula.1.1}
\end{aligned}
\end{equation}

The architecture parameter $\alpha$ and the model weight $w$ are simultaneously optimized by bi-level optimization. When the searching process finishes, $w$ is discarded and the architecture is discretized according to $\alpha$. This process is conducted based on the hypothesis that the operator with a larger value of $\alpha_o^{(i,j)}$ is more important to the mixed output. DARTS preserves the operator with the maximal value of the architecture parameters for each edge $(i,j)$. This strategy is widely adopted in the DARTS-based approaches.

\subsection{Problems in DARTS}
We claim that there are two major problems of DARTS in its training and discretization respectively, which trigger the degradation issue. In the former phase, DARTS directly optimizes the architecture parameter without considering the balance between norms of different nodes, incurring incorrect parameter updating. In the latter phase, DARTS neglects the correlation of candidate operations and  approximates the continuous architecture only based on the unstable architecture parameter, thus failing to achieve the ideal architecture.

\textbf{Imbalanced Norms.} Given mixed operation $\bar{o}^{(i,j)}(x)$ on edge $(i,j)$ with a set of operations $\{o|o\in \mathcal{O}\}$ and the corresponding architecture parameters $\alpha_o^{(i,j)}$, the updating process of $\alpha_o^{(i,j)}$ in the first-order term can be written as:

\begin{equation}
\begin{aligned}
\frac{\partial L_{valid}}{\partial \alpha_o^{(i,j)}} = &\frac{\partial L_{valid}}{\partial \bar{o}^{(i,j)}(x)}\cdot
\frac{\partial \bar{o}^{(i,j)}(x)}{\partial \alpha_o^{(i,j)}} \\= &  S_o \frac{\partial L_{valid}}{\partial \bar{o}^{(i,j)}(x)}\cdot (o(x)-\bar{o}^{(i,j)}(x))
\label{Formula.alpha_update}
\end{aligned}
\end{equation}
where $S_o$ is the weight of operation $o(x)$ after the softmax function. In this form, we can find when $o(x)$ approaches to 0, the gradient of its corresponding architecture parameter has a higher volatility. When $o(x)$ approaches to $\bar{o}^{(i,j)}(x)$, its corresponding update is close to 0, making it harder to tune $\alpha_o^{(i,j)}$.

\begin{figure*}[!htb]
\begin{center}
\end{center}
	\includegraphics[width = 1.0\linewidth]{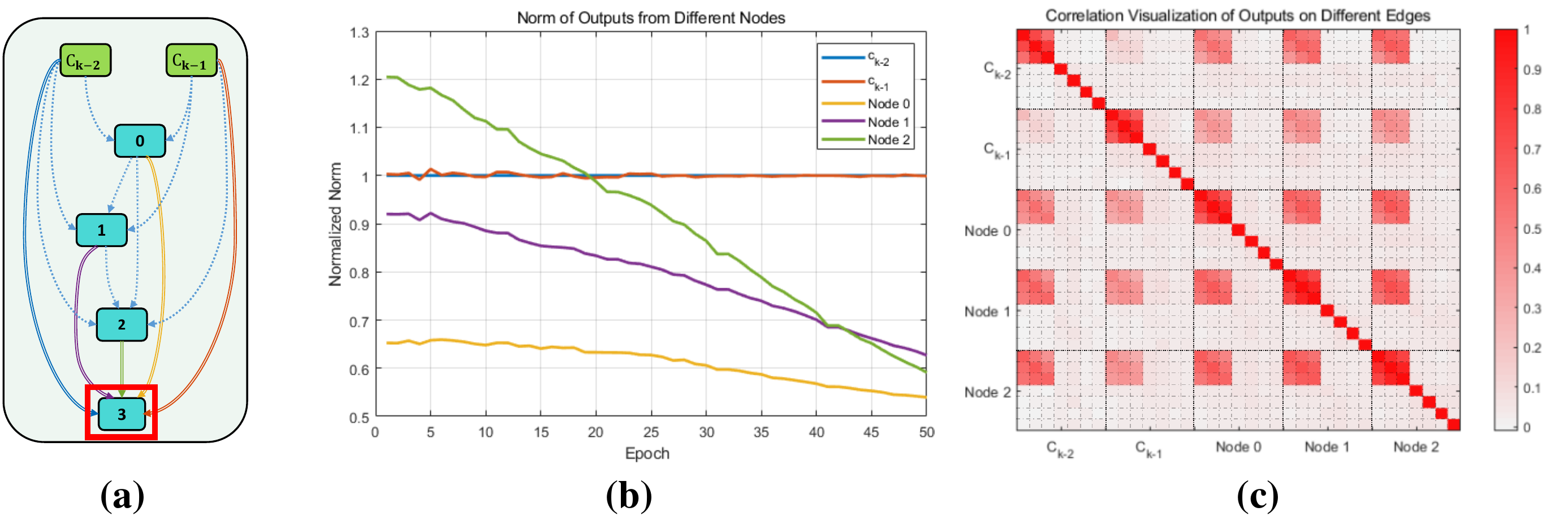}
   \caption{Visualization of the imbalance of nodes and the correlation between operations in a cell of DARTS: (a) the dense connection within a single cell; (b) the curves of norms of the nodes in training (normalized by the norm of Node $c_{k-2}$); and (c) the correlation matrix between operations of Node $3$ before discretization.}
\label{fig:problem}
\end{figure*} 

DARTS tries to fix the operation norms in the training phase by adding a static batch normalization, \emph{i.e.}, batch normalization without learnable affine parameters. However, the \textit{skip-connect} operation is neglected. For an arbitrary input $x$ with the shape of $(B,C,W,H)$, the norm of outputs from the operations excluding \textit{skip-connects} is a constant $\mathbf{C}$ which equals to $\sqrt{B \times C\times W\times H} $ since the static batch normalization rescales the outputs to a distribution with zero-mean and unit-variance. In the meantime, \textit{skip-connects} return the original norm of this arbitrary input. As shown in Fig. \ref{fig:problem} (b), in the search process of DARTS, the norms of inputs on different edges are usually not in the same scale, and this imbalance is further aggravated with more iterations, making the norm of the output from each node approach to 0. In this case, the mid-layer nodes become invalid and the ones linked to the input are highlighted. Therefore more \textit{skip-connects} are selected and a shallow architecture is established.

\label{Correlation}
\textbf{Correlated Operations.} In discretization, DARTS searches for a few operations whose synthesis well approximates to the super-net output. Recall that static batch normalization (denoted as $f$) is universally used in super-net. Each convolution-based operation is followed by a static batch normalization. The norms of their outputs are a constant $\mathbf{C}$ which only depends on the feature map size, batch size, and feature channels. We denote the output $z$ from mixed operation $\bar{o}$ after a static batch normalization as $z=f(\bar{o})$. Here, we take the single operation selection as an example. When the architecture is discretized, the selected operation $o_i$ and the edge output $z'$ is represented as $z'=f(o_i)$. The distance between the original output $z$ and the approximated output $z'$ is formulated as:
\begin{equation}
\begin{aligned}
||z-z'||_2=&\sqrt{ |z|^2+|z'|^2-2|z|\cdot|z'|\cdot cos\theta_i}\\=& \sqrt{2\mathbf{C}^2-2\mathbf{C}^2\cdot cos\theta_i}
\label{Formula.distance}
\end{aligned}
\end{equation}
where $\theta_i$ denotes the vectorial angle between $z$ and $z'$. In discretization, we need to minimize the gap between the super-net $z$ and the approximated one $z'$.

We then denote the normalization of $\bar{o}$ as $\hat{o}=\frac{\bar{o}}{\sigma_{\bar{o}}}$ where $\sigma_{\bar{o}}$ is the variance of $\bar{o}$. It is also a distribution with zero-mean and unit-variance as $o_i$. There is only the difference of constant times between $\hat{o}$ and $\bar{o}$, and due to the property of batch normalization, we have $z=f(\bar{o})=f(\hat{o})=\hat{o}$ and $z'=f(o_i)=o_i$. $\theta_i$ can thus be written as

\begin{equation}
\begin{aligned}
\theta_i = &arccos(\frac{z'\cdot z}{|z'||z|}) = arccos(\frac{o_i \cdot \hat{o}}{|o_i||\hat{o}|})=arccos (\frac{o_i\cdot \bar{o}}{\sigma_{\bar{o}} C^2})\\
= &arccos(\frac{o_i\cdot \sum_{j=1}^n\alpha_j o_j}{\sigma_{\bar{o}} C^2})\\
= & arccos(\frac{\alpha_i\cdot C^2+\sum_{j=0,j\neq i}^n{\alpha_j\cdot<o_i,o_j>}}{\sigma_{\bar{o}} C^2}) \\
= & arccos(\frac{\alpha_i+\sum_{j=0,j\neq i}^n{\alpha_j\cdot cos(\theta^{(i,j)})}}{\sigma_{\bar{o}}})
\label{Formula.distance}
\end{aligned}
\end{equation}
where $<\cdot>$ represents inner product and $\theta^{(i,j)}$ denotes the vectorial angle between $o_i$ and $o_j$. Notice $\sigma_{\bar{o}}$ is a constant if training ends. We can find that $\theta_i$ is not consistent with $\alpha_i$ if the correlation exists among operations ($cos(\theta^{(i,j)})$ is a non-zero value). 
When the similarity of the selected operation and the mixed ones increases, the discrepancy between the outputs before and after discretization decreases. 

Since the connections in the auxiliary one-shot model are highly redundant, the operation outputs are indeed correlated as in Fig. \ref{fig:problem} (c). It increases the risk in selecting the operations that produce redundant feature maps but with larger values of architecture parameters. Besides, as stated in the previous studies \cite{DBLP:journals/corr/abs-1904-12760}, some operations \emph{i.e.} \textit{skip-connects} and \textit{poolings} are prone to gain larger architecture values in the gradient backward process, because they have fewer parameters than convolutions. Therefore, DARTS usually delivers shallow architectures full of \textit{skip-connects}.

\begin{algorithm}[htb]
\caption{Decorrelation Discretization for iDARTS}
\label{Alg:1}
\begin{algorithmic}[1]
\REQUIRE ~~\\
$I_{val}$: validation set; $C$: cell number; $N$: node number; $K$: number of predecessors for each node. \\
\ENSURE ~~\\
$S$: selected operator set.
\STATE Initialize: $S = []$\\
\FOR {$k=1$ to $N \times K$}
{
\FOR {$c=1$ to $C$}
{\FOR {$n=1$ to $N$}
\STATE Calculate node output $y_c^{(n)}=\sum_{i<n}\bar{o}^{(i,n)}$ in $I_{val}$
\STATE Orthogonalize $y_c^{(n)}$  by all selected operators in $S$
\STATE Calculate the cosine similarity $\theta_{c,j}^{(n)}$ between  $y_c^{(n)}$ and corresponding operators $o_j$.
\ENDFOR
}
\ENDFOR
}
\STATE Calculate the mean cosine similarity over all the cells as ${\bar{\theta}}_{k,j}^{(n)}=\frac{1}{C}\sum_{c=1}^C\theta_{c,j}^{(n)}$
\STATE Select the $j_k^{(n)}$-th operator in node $n$ according to $j_k^{(n)}=argmax\{j:{\bar{\theta}_{k,j}^{(n)}}\}$ and add it into $S$
\ENDFOR\\
\STATE Return $S$.

\end{algorithmic}
\end{algorithm}

\subsection{Solutions}
According to the analysis above, we propose two effective solutions to deal with the problems of imbalanced norms and correlated outputs in training and discretization respectively.

\textbf{Node Normalization.} As stated in Section \ref{Correlation}, we know that imbalanced norms exist in DARTS, leading to unstable updating of architecture parameters. Since the norms of the outputs from the operators excluding \textit{skip-connects} are a constant $\mathbf{C}$, we can introduce additional normalization for \textit{skip-connects} to ensure the norm consistency among operators.  

\begin{figure}[!hbt]
\begin{center}
\end{center}
	\includegraphics[width = 1.0\linewidth]{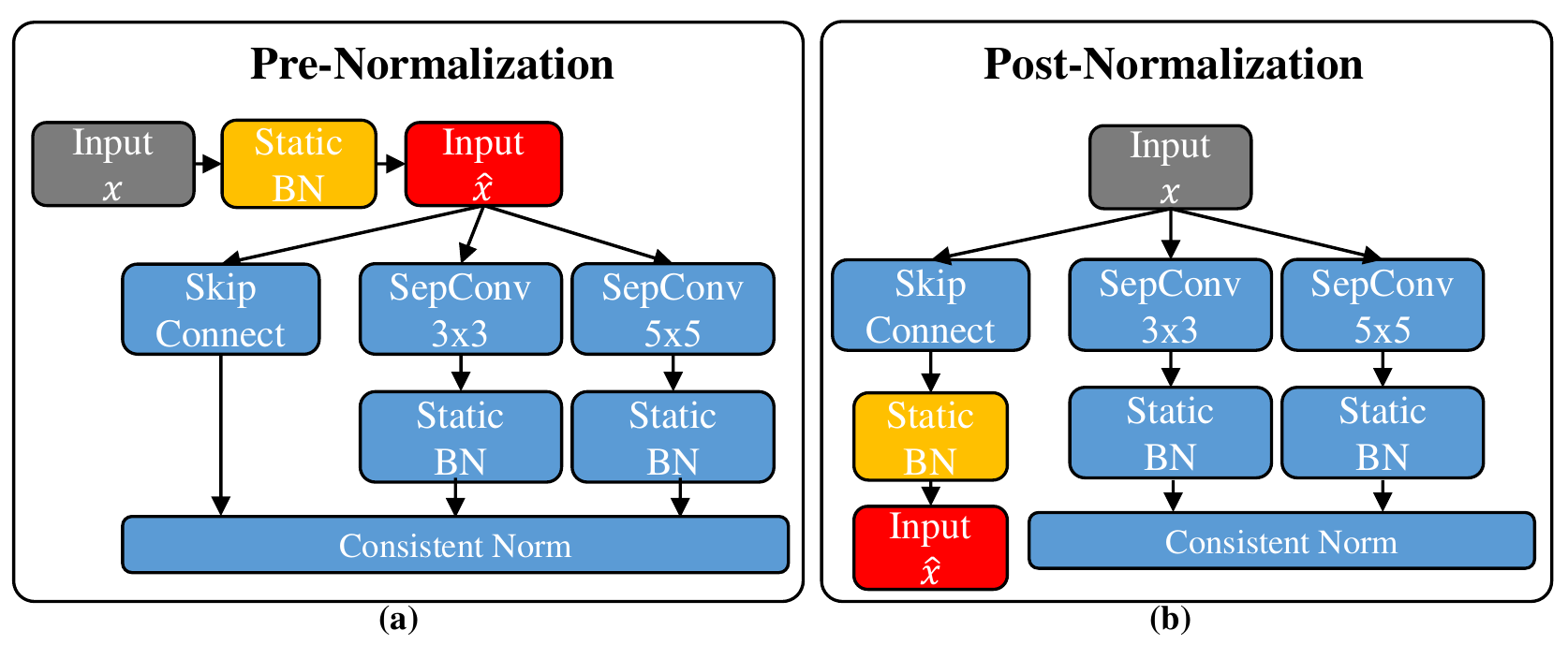}
   \caption{Visualization of candidate solutions to imbalanced norms. (a) and (b) are pre-normalization and post-normalization to \textit{skip-connect} operations. The gray blocks denote the original input for each cell, the orange blocks denote additional normalization, the blue blocks denote the operations in DARTS, and the red blocks denote the normalized results.}
\label{fig:norm}
\end{figure}

As Fig. \ref{fig:norm} illustrates, we can apply pre-normalization or  post-normalization on \textit{skip-connect}. Post-normalization is an intuitive solution, and it directly adds a static batch normalization after each \textit{skip-connect} operation. For input $x$, its output is normalized to $\hat{x}$ with a constant norm $\mathbf{C}$.  However, it breaks the consistency between the output of \textit{skip-connects} and the input of other operations on the same edge, which contravenes the design of \textit{skip-connects} blocks. 

The alternative is pre-normalization. When the summarization is finished and $x$ is reached in each node, we add a static batch normalization to $x$. After that, the subsequent nodes take the normalized $\hat{x}$ as input and the outputs of those operations including \textit{skip-connects} are constrained to the same scale without extra computation burden. Finally, we employ pre-normalization as our node normalization strategy in practice. It significantly improves the stability of the parameter updating process.

\textbf{Decorrelation Discretization.} As in Section \ref{Correlation}, we clarify that there exist correlations among the operators. The value of $\alpha$ cannot fully reflect the importance in such a situation. Therefore, the correlation among those operators should be taken into consideration for architecture discretization. Recall that we aim to preserve the operations whose synthesis is most similar to the original output of the supernet. 

To this end, an intuitive solution is to project the outputs of the operators onto an orthogonal set using the Gram-Schmidt process, and the importance of the operators is decided by the projection weights on those orthogonal biases. But it is not deterministic since there exist multiple orthogonal bias sets in the same operation space. An alternative is to find which set of the operators can synthesize the output of the super-net to the most extent. Nevertheless, there are $C_M^K$ sets for the edge with $M$ candidate operators to test and it is unaffordable to go through all the possible combinations. 

Instead, we propose a novel discretization strategy to approximate the optimal combination. We directly search for the operator $o^{(i,j)}$ with the maximum projection length on the synthesized output $y^{(j)}$ and then remove its projection component from $y^{(j)}$. In this case, the remainder  $\hat{y}^{(j)}$ is orthogonal to the selected operation $o^{(i,j)}$. The orthonormalizing step can be illustrated as follows:

\begin{equation}
\begin{aligned}
\hat{y}^{(j)} = y^{(j)}-\frac{<o^{(i,j)},y^{(j)}>}{<o^{(i,j)},o^{(i,j)}>}o^{(i,j)}
\label{Formula.2}
\end{aligned}
\end{equation}

Since the output of each cell concatenates all the nodes, such dense connections also incur the redundancy in the cell output. We conduct the decorrelation step at both the node-level and cell-level. Firstly, we recursively select the operation with the highest similarity to the node outputs and decorrelate it from the node outputs until the number of predecessors for each node is satisfied. Then, we decorrelate each selected operation from the outputs of all the nodes rather than the single node it belongs to. The discretization details are illustrated in Algorithm \ref{Alg:1}. Benefiting from decorrelation discretization, the proper operators synthesize the supernet to the most extent.

\section{Experiments}
\label{S.EXP}
\subsection{Datasets and Settings}
\label{S.exp.ds}
We launch extensive experiments on CIFAR-10  \cite{krizhevsky2009learning} and ImageNet \cite{DBLP:conf/cvpr/DengDSLL009}. CIFAR-10 has 60K images of the resolution at $32\times 32$, equally distributed over 10 classes. We adopt the standard split where 50K images are used for training and the rest for testing. ImageNet contains around 1.3M images belonging to 1,000 classes, with 1.2M images for training and 50K images for validation. We follow the general setting to resize all the images to 224$\times$224 for both training and testing.

We employ the same search space (denoted as $S1$) as in DARTS, \emph{i.e.} 8 different candidate operations, including 3$\times $3 and 5$\times $5 separable convolutions, 3$\times $3 and 5$\times $5 dilated separable convolutions, 3$\times $3 max pooling, 3$\times $3 average pooling, identity, and \emph{zero}. For searching on CIFAR-10, we use the same one-shot model as DARTS where 8 cells with 16 initial channels are trained. We take SGD for optimization with initial learning rate at 0.025 and cosine annealing to 0.001. The momentum is 0.9 and the weight decay is 3$\times10^{-4}$. The epoch number is set at 50 and the batch size at 64. Following \cite{DBLP:journals/corr/abs-1907-05737}, we freeze the architecture parameters in the first 15 epochs. We randomly select 5,000 images from the training set for discretization. 

For directly searching on ImageNet, following previous studies \cite{DBLP:journals/corr/abs-1909-06035, DBLP:journals/corr/abs-1907-05737}, we reduce the resolution of the input image from 224$\times$224 to 28$\times$28 with three stacked convolution layers of stride 2. The one-shot model also consists of 8 cells with 16 initial channels as on CIFAR-10. We select 5\% data from the training set to update model weights and another 5\% to update architecture parameters. The epoch number is set at 60 for further convergence. Batch size is set at 128 for both training and validation. SGD is applied for optimization with the initial learning rate at 0.2. We freeze the architecture parameters in the first 35 epochs. For architecture parameters, the weight decay is set at 0.001 and the learning rate is 0.006. A set with 10,000 images are randomly sampled from the training set for discretization.

For both the datasets, the weights of the one-shot model and the architecture parameters are updated alternatively. We determine the final structure based on the proposed discretization strategy. 

\begin{figure*}[!htbp]
	\subfigure[Normal cell searched on CIFAR-10]{
    \begin{minipage}[h]{0.5\linewidth}
    \centering
      \includegraphics[width = 1\linewidth]{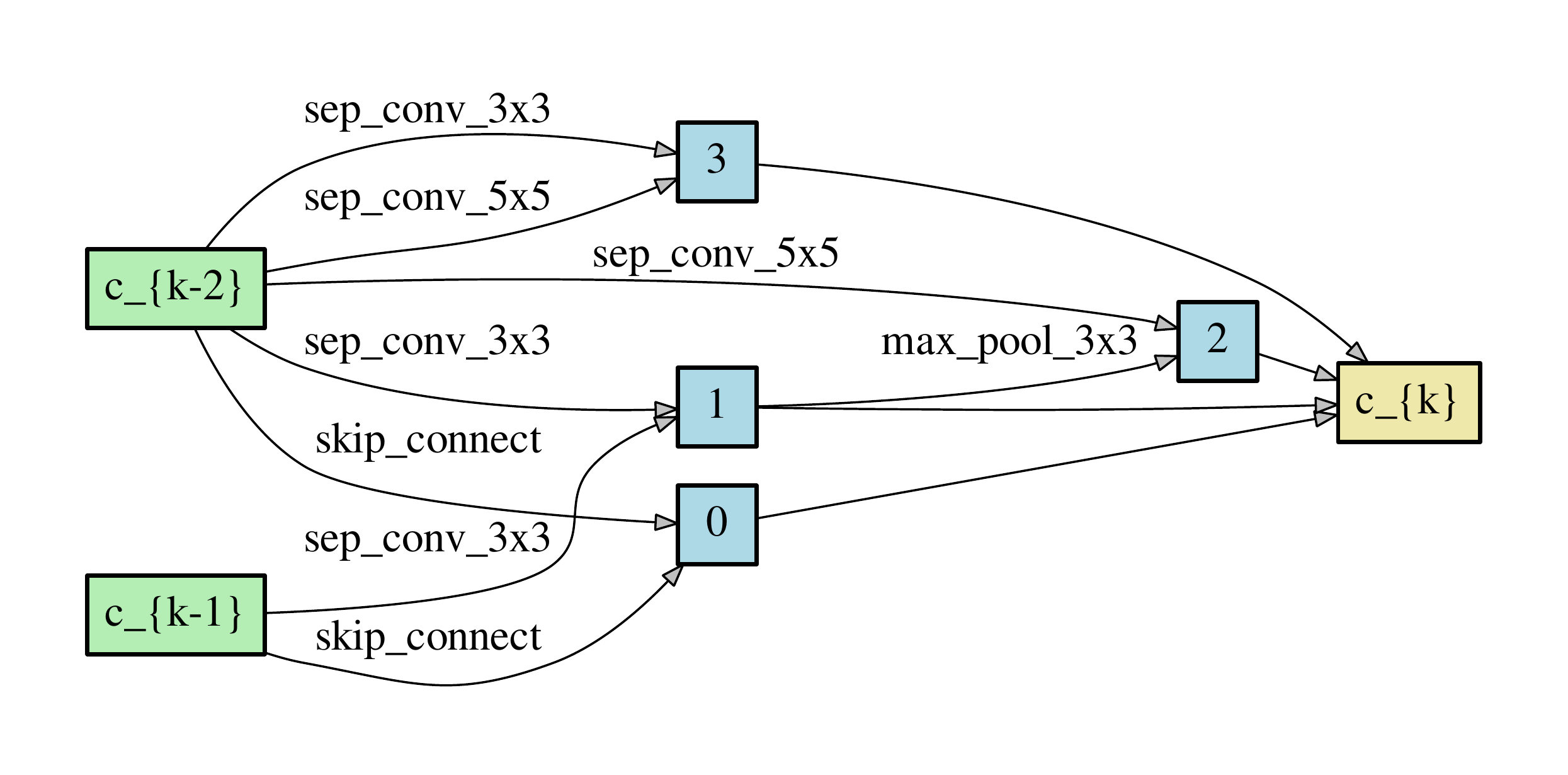}
    \end{minipage}
    }
    \subfigure[Reduction cell searched on CIFAR-10]{
    \begin{minipage}[h]{0.5\linewidth}
    \centering
      \includegraphics[width= 1\linewidth]{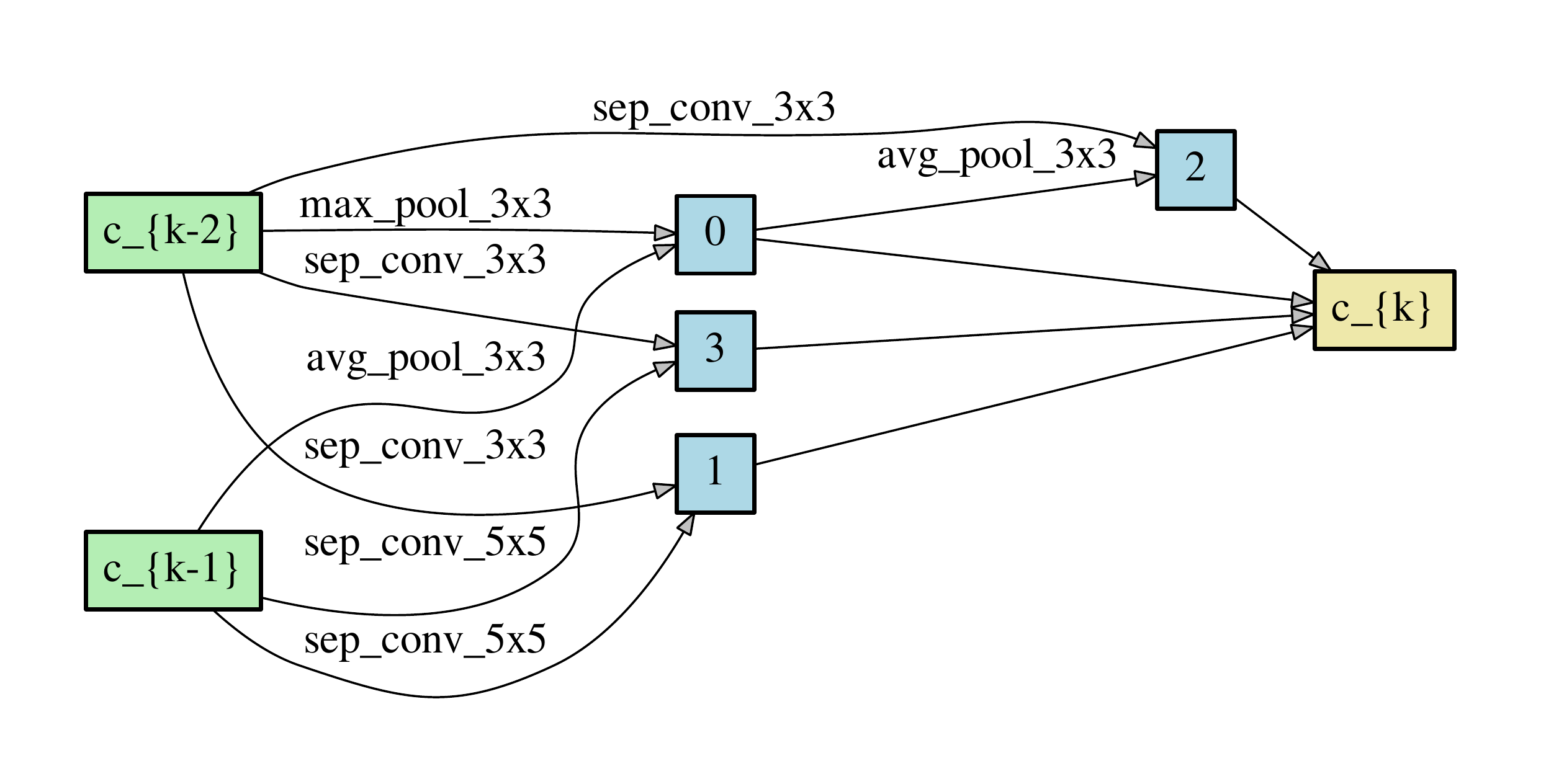}
    \end{minipage}
    }
    \subfigure[Normal cell searched on ImageNet]{
    \begin{minipage}[h]{0.5\linewidth}
    \centering
      \includegraphics[width = 1\linewidth]{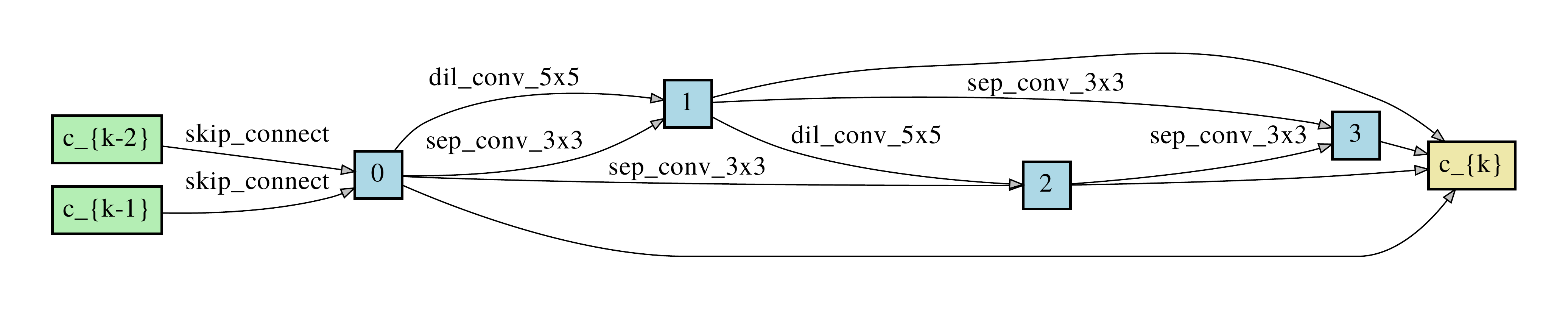}
    \end{minipage}
    }
    \subfigure[Reduction cell searched on ImageNet]{
    \begin{minipage}[h]{0.5\linewidth}
    \centering
      \includegraphics[width= 1\linewidth]{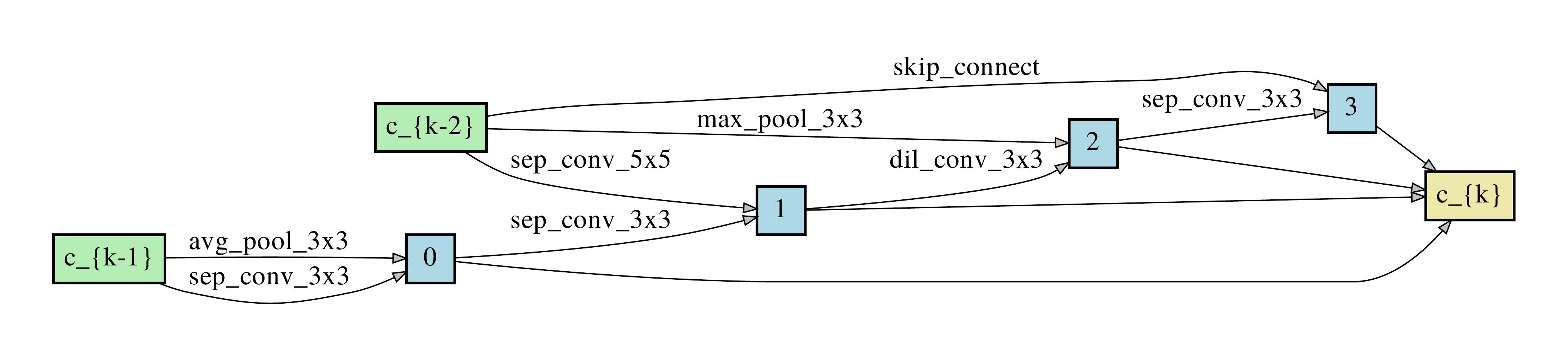}
    \end{minipage}
    }
   \caption{Visualization of the normal cell (a) (c) and the reduction cell (b) (d) learned on CIFAR-10 and ImageNet.}
\label{fig:S0}
\end{figure*}

The searching process takes 0.2 GPU-day on CIFAR-10 and 1.9 GPU-day on ImageNet with a single Nvidia Tesla V100. The architectures achieved on CIFAR-10 and ImageNet are shown in Fig. \ref{fig:S0}.

For training on CIFAR-10, we use the network consisting of 20 cells with 36 initial channels. We adopt two training protocols for comprehensive comparison with the state-of-the-arts. The first protocol is the same as DARTS. The training lasts 600 epochs and the hyper-parameters are the same as those in DARTS. The second protocol follows DARTS+ \cite {DBLP:journals/corr/abs-1909-06035}, where the number of the training epochs is set at 2,000 for better convergence and the weight decay is set at 5$\times 10^{-4}$. We also conduct data enhancement including cutout, path dropout and auxiliary towers as in \cite{DBLP:journals/corr/abs-1909-06035,DBLP:journals/corr/abs-1910-11831,DBLP:journals/corr/abs-1909-09656,DBLP:journals/corr/abs-1904-12760,DBLP:conf/iclr/LiuSY19}. The training time lasts 0.9 day for 600 epochs and 3 days for 2,000 epochs with a single Nvidia Tesla V100.

For training on ImageNet, we consider the mobile setting where the input image size is $224\times 224$ and the number of multiply-add operations is less than 600M. We follow DARTS and set the number of cells at 14 with 48 initial channels. The model is trained for 250 epochs with the batch size of 1,024. We also take SGD as the optimizer with initial learning rate at 0.4 (cosine annealing to 0), momentum at 0.9 and weight decay at $3\times 10^{-5}$. We exploit learning rate warmup, label smoothing, and auxiliary loss tower as in \cite{DBLP:journals/corr/abs-1909-06035,DBLP:journals/corr/abs-1910-11831,DBLP:journals/corr/abs-1909-09656,DBLP:journals/corr/abs-1904-12760,DBLP:conf/iclr/LiuSY19}. The training phase lasts for 3.8 days with 3 Nvidia Tesla V100 GPUs.

\subsection{Results on CIFAR-10}
\label{EXP.Res}
\textbf{Comparison to State-of-the-arts.} A number of state-of-the-art methods are considered for comparison, including DARTS-based methods and classical handcrafted architectures. To reduce the randomness in the initialization and optimization procedure, we repeat the training process 4 times. The best, the mean and the variance of the 4 runs are reported as recent studies do. The search cost is also reported as in most DARTS-based methods \cite{DBLP:conf/iclr/LiuSY19, DBLP:journals/corr/abs-1907-05737, DBLP:journals/corr/abs-1904-12760, DBLP:journals/corr/abs-1909-06035, DBLP:journals/corr/abs-1909-09656, DBLP:conf/iclr/Wang21, DBLP:journals/corr/abs-1910-11831}, which denotes the training processof the one-shot model.

\begin{table*}[t]
\begin{center}
\caption{Comparison to the state-of-the-art methods on the CIFAR-10 dataset (ME: training with 2,000 epochs; \dag: including searching and training; \ddag: the re-implemented results; and *: evaluation on Nvidia Tesla V100).} 
{\begin{tabular}{c|c|c|c|c}
\hline
\hline
\multirow{2}{*}{Architecture} & Test Err. & Params & Search Cost & \multirow{2}{*}{Search Method}\\
\cline{2-4}
& (\%) & (M) &  (GPU-days) \\
\hline
DenseNet \cite{DBLP:conf/cvpr/HuangLMW17} & 3.46 & 25.6 & - &Manual\\
NASNet-A + cutout \cite{DBLP:conf/cvpr/ZophVSL18} & 2.65 & 3.3 & 1,800 & RL \\
AmoebaNet-A + cutout \cite{DBLP:conf/aaai/RealAHL19} & 3.34 $\pm$ 0.06 & 3.2 & 3,150  & Evolution \\
AmoebaNet-B + cutout \cite{DBLP:conf/aaai/RealAHL19} & 2.55 $\pm$ 0.05 & 2.8 & 3,150 & Evolution \\
Hierarchical Evolution \cite{DBLP:conf/iclr/LiuSVFK18} & 3.75 $\pm$ 0.12 & 15.7 & 300 & Evolution \\
PNAS \cite{DBLP:conf/eccv/LiuZNSHLFYHM18} & 3.41 $\pm$ 0.09 & 3.2 & 225 & SMBO\\
ENAS + cutout \cite{DBLP:conf/icml/PhamGZLD18} & 2.89 & 4.6 & $0.45^{\dag}$ & RL\\
NAONet-WS \cite{DBLP:conf/nips/LuoTQCL18} & 3.53 & 2.5 & $0.3^{\dag}$ & NAO \\
\hline
DARTS (1st order) + cutout \cite{DBLP:conf/iclr/LiuSY19} & 3.00 $\pm$ 0.14 & 3.3 & $0.4^{\ddag}$ & Gradient-based\\
DARTS (2nd order) + cutout \cite{DBLP:conf/iclr/LiuSY19} & 2.76 $\pm$ 0.09 & 3.3 & $1^{\ddag}$ & Gradient-based\\
SNAS (mild)+ cutout \cite{DBLP:conf/iclr/XieZLL19} & 2.98 & 2.9 & 1.5 & Gradient-based\\
ProxylessNAS + cutout \cite{DBLP:conf/iclr/CaiZH19} & 2.08 & 5.7 & $4^{\ddag}$ & Gradient-based \\
P-DARTS + cutout \cite{DBLP:journals/corr/abs-1904-12760} & 2.50 & 3.4 & 0.3 & Gradient-based \\
BayesNAS + cutout \cite{DBLP:conf/icml/ZhouYWP19} & 2.81 $\pm$ 0.04 & 3.4 & 0.2 & Gradient-based\\
PC-DARTS + cutout \cite{DBLP:journals/corr/abs-1907-05737} & 2.57 $\pm$ 0.07 & 3.6 & 0.1 &Gradient-based \\
DARTS+ (Rule 1) + cutout + ME \cite{DBLP:journals/corr/abs-1909-06035} & 2.50 $\pm$ 0.11 & 3.7 & $0.4^{*}$ &Gradient-based \\
DARTS+ (Rule 2) + cutout + ME \cite{DBLP:journals/corr/abs-1909-06035} & 2.37 $\pm$ 0.13 & 4.3 & $0.6^{*}$ &Gradient-based \\
Amended-DARTS + cutout \cite{DBLP:journals/corr/abs-1910-11831} & 2.60 $\pm$ 0.15 & 3.6 & 1.1 &Gradient-based \\
SDARTS-RS + cutout \cite{DBLP:conf/icml/ChenH20} & 2.67 $\pm$ 0.03 & 3.4 & 0.4 & Gradient-based \\
DARTS+PT + cutout \cite{DBLP:conf/iclr/Wang21} & 2.61 $\pm$ 0.08 & 3.0 & 0.8 & Gradient-based\\
R-DARTS (L2) + cutout \cite{DBLP:journals/corr/abs-1909-09656} & 2.95 $\pm$ 0.21 & - & 1.6 & Gradient-based\\
\hline
iDARTS + cutout & 2.35 (2.45 $\pm$ 0.05) & 3.6 & 0.4/$0.2^{*}$ & Gradient-based \\
iDARTS + cutout + ME & 2.25 (2.38 $\pm$ 0.10) & 3.6 & 0.4/$0.2^{*}$ &Gradient-based\\
\hline
\hline
\end{tabular}}
\label{T.CIFAR10}
\end{center}
\end{table*}

From Table \ref{T.CIFAR10}, we can see that iDARTS delivers a very competitive accuracy with the top-1 error of 2.45\% in 600 epochs and 2.38\% in 2,000 epochs, and the model is built in 0.2 GPU-day for architecture search. We notice that DARTS (1st-order) and iDARTS adopt the same bi-level optimization algorithm, but due to node normalization and decorrelation discretization, iDARTS significantly reduces the top-1 error of DARTS from 3.00\% to 2.45\% without introducing any hyper-parameters or amending the gradient-descent direction. iDARTS outperforms all the counterparts except \cite{DBLP:conf/iclr/CaiZH19} and \cite{DBLP:journals/corr/abs-1909-06035}. It should be noted that Proxyless-NAS \cite{DBLP:conf/iclr/CaiZH19} makes use of much more parameters and consumes much higher computational cost (4 GPU-day).  Although DARTS+ \cite{DBLP:journals/corr/abs-1909-06035} reaches a comparable performance with iDARTS, its early-stop criterion needs to be empirically decided, which is not always guaranteed, in particular when applied to different search spaces. Meanwhile, both DARTS+PT \cite{DBLP:conf/iclr/Wang21} and iDARTS modify the value-based discretization strategy. iDARTS employs more efficient discretization and additional norm constraints, and this solution reaches higher performance with much lower searching cost.

\label{EXP.Ablation}
\textbf{Ablation Study.} We investigate the impact of the proposed node normalization and decorrelation discretization solutions on CIFAR-10. We take DARTS-1st as the baseline for its efficiency. We re-implement DARTS and discretize the architecture by the value-based strategy and the decorrelation discretization in the same run for fair comparison. The training epoch is extended to 2,000 for better convergence. Results are shown in Table \ref{T.Ablation}.

As in Table \ref{T.Ablation}, when only decorrelation discretization is used in DARTS, the error is largely reduced to 2.57$\pm$0.05\%, which highlights its effectiveness. When node normalization is then added to build iDARTS, the result is optimized to 2.38$\pm$0.10\%, proving its necessity. It should be noted that it does not make much sense to only use node normalization in DARTS (the error is 2.90$\pm$0.10\%), as we cannot deliver a sound architecture without correctly approximating the continuous one. To sum up, node normalization and decorrelation discretization improve DARTS from different aspects and their combination reaches the best performance. Besides, node normalization merely introduces a static batch normalization to each cell and decorrelation discretization is applied only once in the last epoch. The overall search cost is still 0.2 GPU-day on a single Tesla V100.

\begin{figure*}[!htbp]
%\begin{center}
%\end{center}
	\subfigure[Normal cell searched in $S2$ on CIFAR-10]{
    \begin{minipage}[h]{0.5\linewidth}
    \centering
      \includegraphics[width = 1\linewidth]{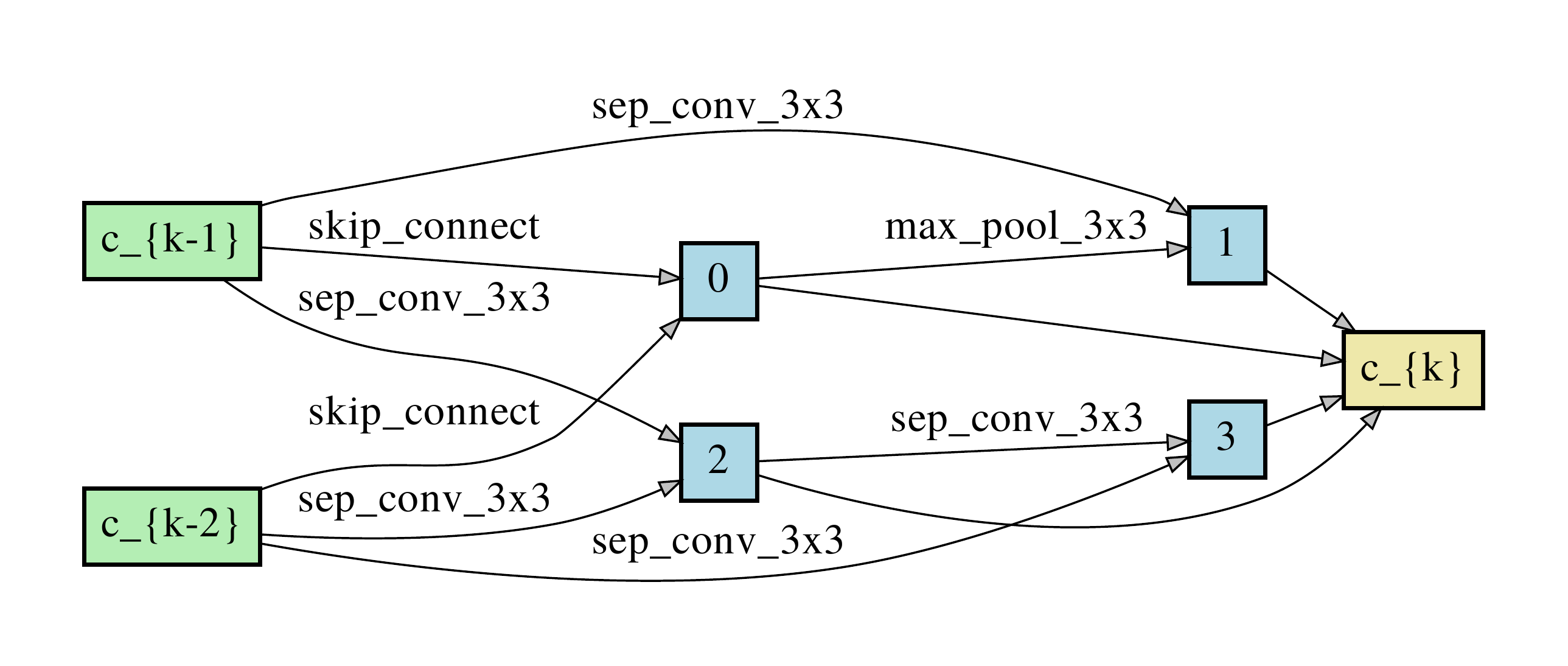}
    \end{minipage}
    }
    \subfigure[Reduction cell searched in $S2$ on CIFAR-10]{
    \begin{minipage}[h]{0.5\linewidth}
    \centering
      \includegraphics[width= 1\linewidth]{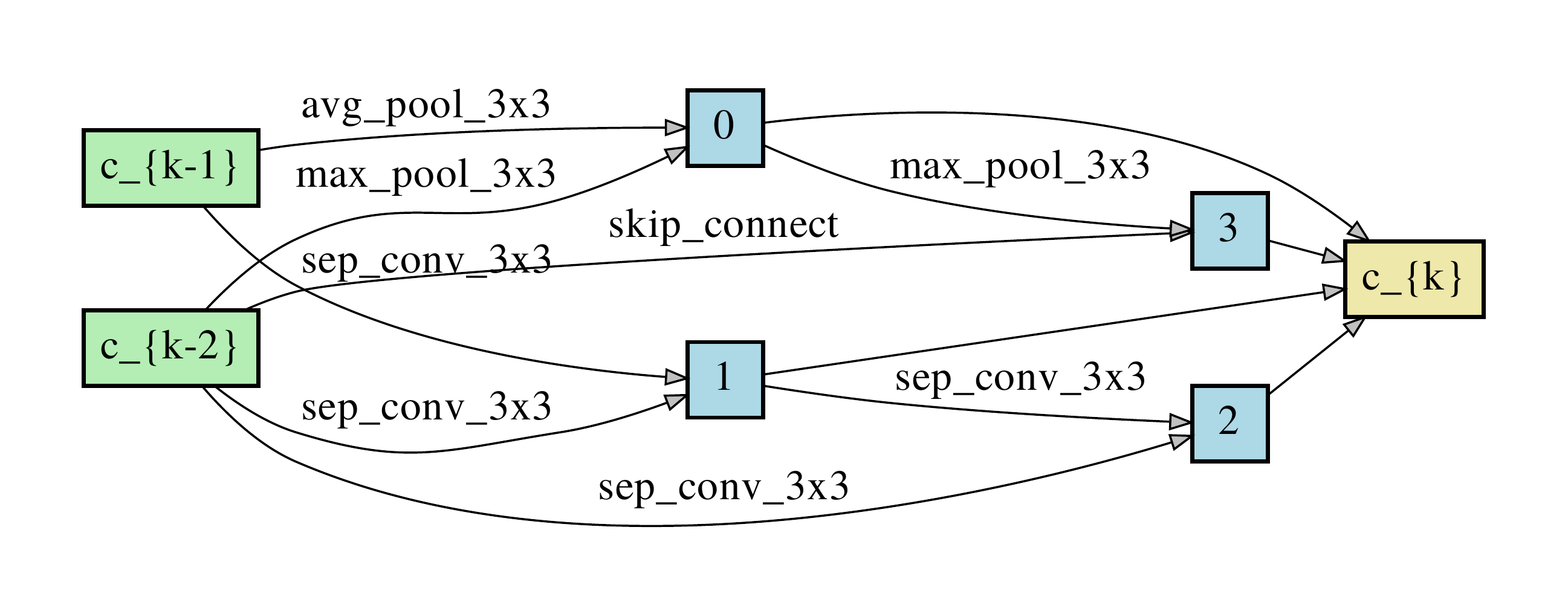}
    \end{minipage}
    }
    \subfigure[Normal cell searched in $S3$ on CIFAR-10]{
    \begin{minipage}[h]{0.5\linewidth}
    \centering
      \includegraphics[width = 1\linewidth]{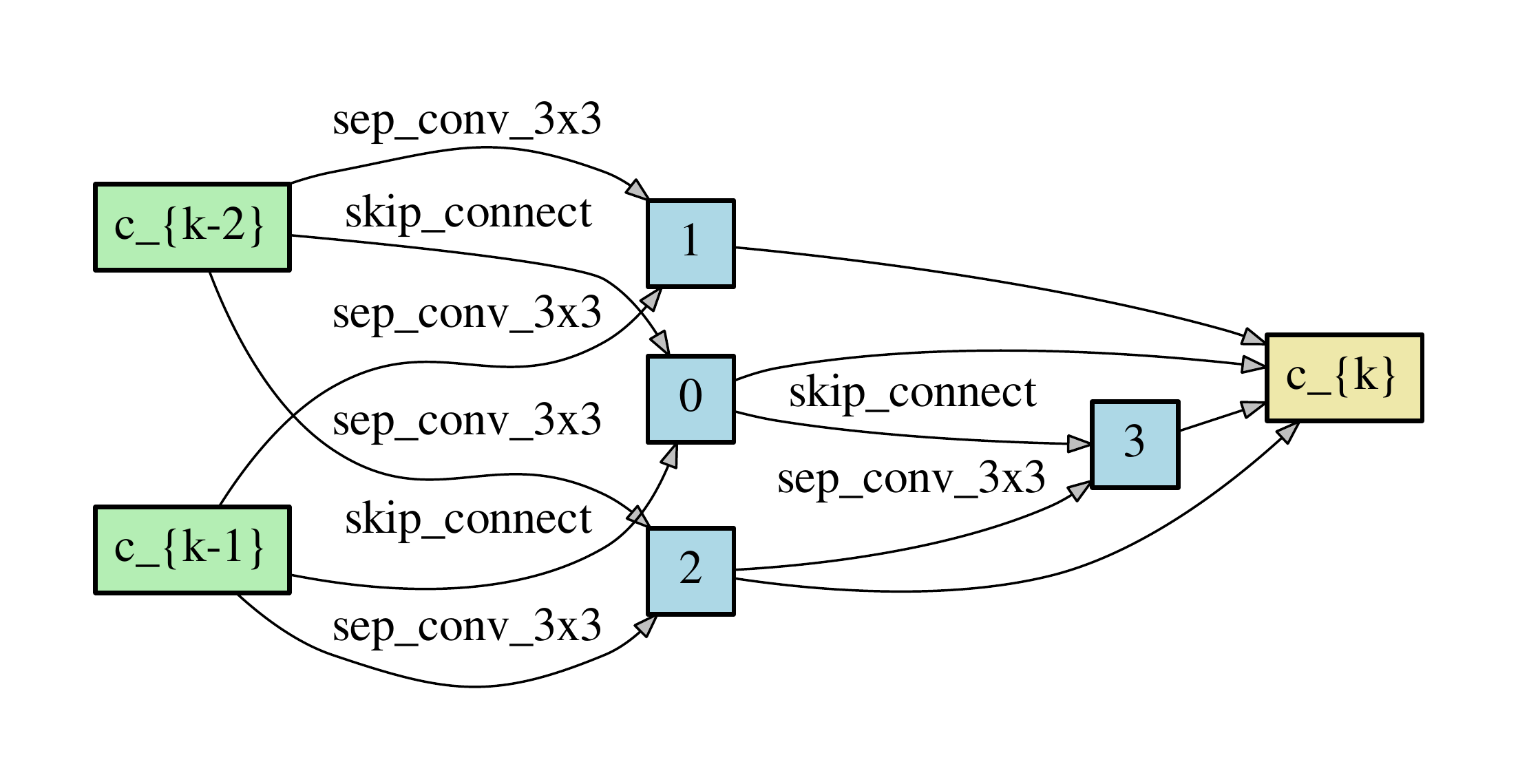}
    \end{minipage}
    }
    \subfigure[Reduction cell searched in $S3$ on CIFAR-10]{
    \begin{minipage}[h]{0.5\linewidth}
    \centering
      \includegraphics[width= 1\linewidth]{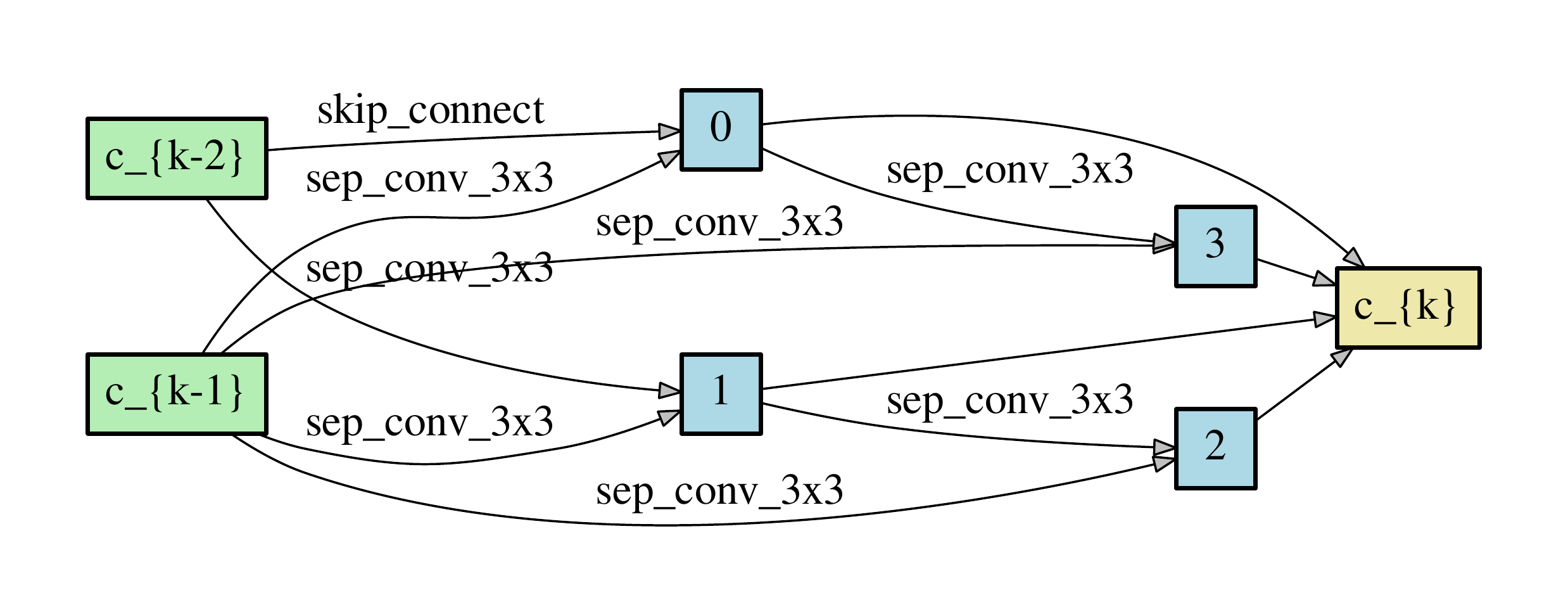}
    \end{minipage}
    }
   \caption{Normal and reduction cells learned on CIFAR-10 in different settings. (a) and (b) are the normal and reduction cells using the original model in $S2$. (c) and (d) are the normal and reduction cells with the small model in $S3$}
\label{fig:S1}
\end{figure*}

\label{EXP.Gen}
\textbf{Evaluation on Generalization.} To evaluate the generalization ability of iDARTS, additional experiments are conducted in two more settings.

%Two different search spaces are considered. One is $S1$ as introduced in Sec. \ref{S.exp.ds} and the other (denoted as $S2$) is smaller, which only contains three operations, namely 3$\times$3 separable convolution, identity and $zero$. We search for a shallower model only with 2 nodes in a single cell on $S1$ as well as a full model on $S2$. The learned architectures are shown in Fig. \ref{fig:S1}. We take DARTS, DARTS+, and PC-DARTS as counterparts in comparison. 

Two additional search spaces are considered. One is $S2$ which contains none, max pooling 3$\times$3, average pooling 3$\times$3, skip connection and separate convolution 3$\times$3. $S3$ is smaller, which only contains three operations, namely 3$\times$3 separable convolution, identity and $zero$. We search for full models on all the search spaces. The learned architectures are shown in Fig. \ref{fig:S1}. We take DARTS, DARTS+, and PC-DARTS as counterparts in comparison.

%\begin{figure*}[!htb]
%\begin{center}
%\end{center}
%	\includegraphics[width = 1.0\linewidth]{fig/archi2.pdf}
%   \caption{Normal and reduction cells learned on CIFAR-10 in different settings. (a) and (b) are the normal and reduction cells using the original model in $S2$. (c) and (d) are the normal and reduction cells with the small model in $S3$}
%\label{fig:S1}
%\end{figure*} 

\begin{table}[!htpb]
\begin{center}
\caption{Ablation studies on the CIFAR-10 dataset (NN: Node Normalization, DD: Decorrelation Discretization).} 
\begin{tabular}{c|c|c|c|c}
\hline
\hline
\multicolumn{2}{c|}{Improvements} & Test Err & Params & Multiply-Add \\
\cline{1-5}
 NN & DD & (\%) & (M) & (M)\\
\hline

$\times$ & $\times$ & 2.81 $\pm$ 0.07 &  2.8 & 464.8\\
$\times$ & $\checkmark$ & 2.57 $\pm$ 0.10 & 2.9 & 471.9\\
$\checkmark$ & $\times$ & 2.90 $\pm$ 0.10 & 2.0 & 331.1\\
$\checkmark$ & $\checkmark$ & \textbf{2.38 $\pm$ 0.10} & 3.6 & 575.8\\
\hline
\hline
\end{tabular}
\label{T.Ablation}
\end{center}
\end{table}

\begin{table*}[!htpb]
\begin{center}
\caption{Comparison of DARTS-based methods in different settings on the CIFAR-10 dataset. R1 and R2 denote Criterion 1 and Criterion 1* in DARTS+ respectively.} 
{
\begin{tabular}{c|c|c|c|c|c|c}
\hline
\hline
\multirow{2}{*}{Architecture} & \multicolumn{2}{c|}{$S1$} & \multicolumn{2}{c}{$S2$} & \multicolumn{2}{c}{$S3$}\\
\cline{2-7}
& Err (\%) & Para (M) &  Err (\%) & Params (M) &  Err (\%) & Para (M)\\
\hline
Baseline & 3.68$\pm$0.09 & 1.8 & 3.09$\pm$0.09 &2.3&  2.81$\pm$0.11 & 3.1 \\
\hline
DARTS (1st) & 3.00$\pm$0.14 & 2.9 & 2.83 $\pm$ 0.14 & 4.4 & 5.52 $\pm$ 0.25 & 1.6\\
DARTS (2nd)& 2.76$\pm$0.09 & 3.3 & 2.82 $\pm$ 0.04 & 4.5 & 3.28 $\pm$ 0.11 & 1.9\\
DARTS+ (R1)& 2.58$\pm$0.05\% & 3.3 & 2.94 $\pm$ 0.04& 4.4 & 5.52 $\pm$ 0.25 & 1.6\\
DARTS+ (R2)& 2.46$\pm$0.02\% & 3.6 & 2.96 $\pm$ 0.10 & 2.2 & 3.04 $\pm$ 0.05 & 3.9\\
PC-DARTS & 2.57$\pm$0.07 & 3.6 & 3.22 $\pm$ 0.07 & 3.9 & 2.60 $\pm$ 0.16 & 3.6\\
\hline
iDARTS & \textbf{2.45 $\pm$0.05} & 3.6 & \textbf{2.50 $\pm$ 0.02} & 3.4 & \textbf{2.53 $\pm$ 0.03} & 3.6\\
\hline
\hline
\end{tabular}}
\label{T.Search}
\end{center}
\end{table*}

\begin{table*}[!htpb]
\begin{center}
\caption{Comparison to the state-of-the-art methods on the ImageNet dataset (ME: training with 800 epochs; and *: evaluation on Nvidia Tesla V100).} 
{
\begin{tabular}{c|c|c|c|c|c|c}
\hline
\hline
\multirow{2}{*}{Architecture} & \multicolumn{2}{c|}{Test Error} & Params & Multiply-Add & Search Cost & \multirow{2}{*} {Search Method} \\
\cline{2-6}
& Top-1 & Top-5 & (M) & (M) & (GPU-day)\\
\hline
Inception-v1 \cite{DBLP:conf/cvpr/SzegedyLJSRAEVR15} & 30.2 & 10.1 & 6.6 & 1,448 &- & Manual \\
MobileNet-v2 (1.4$\times$) \cite{DBLP:conf/cvpr/SandlerHZZC18} & 25.3 & - & 6.9 & 585 & - &Manual\\
ShuffleNet-v2 (1$\times$) \cite{DBLP:conf/eccv/MaZZS18} & 26.4 & 10.2 & ~5 & 524 & - & Manual\\
\hline
NASNet-A \cite{DBLP:conf/cvpr/ZophVSL18} & 26.0 & 8.4 & 5.3 & 564 & 1,800 & RL\\
AmoebaNet-A \cite{DBLP:conf/aaai/RealAHL19} & 25.5 & 8.0 & 5.1 & 555 & 3,150 & Evolution\\
PNAS \cite{DBLP:conf/eccv/LiuZNSHLFYHM18} & 25.8 & 8.1 & 5.1 & 588 & 225 & SMBO\\
MnasNet-92 \cite{DBLP:conf/cvpr/TanCPVSHL19} & 25.2 & 8.0 & 4.4 & 388 & - & RL\\
EfficientNet-B0 \cite{DBLP:conf/icml/TanL19} & 23.7 & 6.8 & 5.4 & 390 & - & RL\\
\hline
DARTS (2nd order) \cite{DBLP:conf/iclr/LiuSY19} & 26.7 & 8.7 & 4.7 & 574 & 4.0 & Gradient-based\\
SNAS (mild) \cite{DBLP:conf/iclr/XieZLL19} & 27.3 & 9.2 & 4.3 & 522 & 1.5 & Gradient-based \\
ProxylessNAS \cite{DBLP:conf/iclr/CaiZH19} & 24.9 & 7.5 & 7.1 & 465 & 8.3 & Gradient-based \\
P-DARTS \cite{DBLP:journals/corr/abs-1904-12760} & 24.4 & 7.4 & 4.9 & 557 & 0.3 &Gradient-based \\
BayesNAS \cite{DBLP:conf/icml/ZhouYWP19} & 26.5 & 8.9 & 3.9 & - & 0.2 & Gradient-based \\
ASAP \cite{DBLP:journals/corr/abs-1904-04123} & 24.4 & - & - & -& 0.2 &Gradient-based \\
XNAS \cite{DBLP:conf/nips/NaymanNRFJZ19} & 24.0 & - & 5.2 & 600 & 0.3 &Gradient-based \\
PC-DARTS \cite{DBLP:journals/corr/abs-1907-05737} & 25.1 & 7.8 & 5.3 & 586 & 0.1 & Gradient-based \\
DARTS+ (ImageNet) + ME \cite{DBLP:journals/corr/abs-1909-06035} & 23.9 & 7.4 & 5.1 & 582 & 6.8* &Gradient-based \\
Amended-DARTS \cite{DBLP:journals/corr/abs-1910-11831} & 24.3 & 7.4 & 5.5 & 590 & 1.1 & Gradient-based \\
\hline 
iDARTS (CIFAR-10) ($S1$) & 25.2 & 7.9 & 5.1 & 578 & 0.2* &Gradient-based\\
iDARTS (ImageNet) ($S1$) & 24.7 & 7.7 & 5.1 & 568 & 1.9* &Gradient-based \\
\hline
\hline
\end{tabular}}
\label{T.ImageNet}
\end{center}
\end{table*}

We use the official implementations of DARTS and PC-DARTS. DARTS+ does not release the code and we apply the two early-stop strategies in the official DARTS implementation, generating the architectures according to the paper. To be specific, the search process of DARTS+ is terminated when there exist more than 2 \textit{skip-connect} operations (R1) or the architecture becomes stable for more than 10 epochs (R2). In the searching and retraining part, all the hyper-parameters are strictly set following the original settings for fair comparison. We also re-implement DARTS+ in the original search space on CIFAR-10, and the results are 2.54$\pm$0.01\% and 2.40$\pm$0.02\% with 2,000 epochs respectively, which are comparable to their results reported in the original paper (2.50$\pm$0.01\% and 2.37$\pm$0.13\%). All the architectures are re-trained with 600 epochs as in DARTS rather than 2,000 epochs. The results are shown in Table \ref{T.Search}.

As illustrated in Table \ref{T.Search}, we can see that iDARTS achieves the most stable performance on all the spaces, demonstrating its strong generalization ability due to node normalization and decorrelation discretization. On the other side, it is worth noting that the early stop strategies in DARTS+ are not well generalized to $S2$ and $S3$, since the criteria need to be carefully adjusted when the search space is changed. 
PC-DARTS reaches comparable results on $S1$ and $S3$, but fails on $S2$. Most operators in $S2$ are highly correlated and partial connections in PC-DARTS tend to be disturbed by such correlations, thus resulting in degraded performance.

%we can see that iDARTS reaches most stable performance on all the search spaces, demonstrating the strong generalize ability of NN and DD. In contrast, the hand-crafted stop criteria in DARTS+ fail in $S2$ and $S3$, which also encounter degradation as DARTS does. It confirms that the hand-crafted criteria need to be carefully designed by human experts according to extensive experiments and are thus hard to generalize to different scenarios. We can also see that the 2nd-order approximation in DARTS relieves degradation on $S3$, but it does not deliver the score as good as that of iDARTS. PC-DARTS reaches comparable results on $S1$ and $S3$, but it fails in $S2$. Since most operators in $S2$ are highly correlated, the partial connection in PC-DARTS may be disturbed by those correlations, thus result in the degradation of performance.

%The results are displayed in Tab. \ref{T.Search}, and we find that iDARTS reaches the best accuracy on both the settings. In contrast, the hand-crafted stop criteria in DARTS+ fail in $S2$, which also encounter degradation as DARTS does. It confirms that the hand-crafted criteria need to be carefully designed by human experts according to extensive experiments and are thus hard to generalize to different scenarios. We can also see that the 2nd-order approximation in DARTS relieves degradation on $S2$, but it does not deliver the score as good as that of iDARTS. 

%
\textbf{Validation of Robustness.} To validate the robustness of iDARTS, we take DARTS (1st-order) as the counterpart since they share the same optimization procedure. Recall that their major differences lie in training and discretization. In training, iDARTS uses node normalization to keep the balance between the norms of different nodes, and in discretization, it approximates the architecture based on the similarity between operations and outputs rather than the values of architecture parameters.

\begin{figure}[!htbp]
\begin{center}
\end{center}
	\includegraphics[width = 1.0\linewidth]{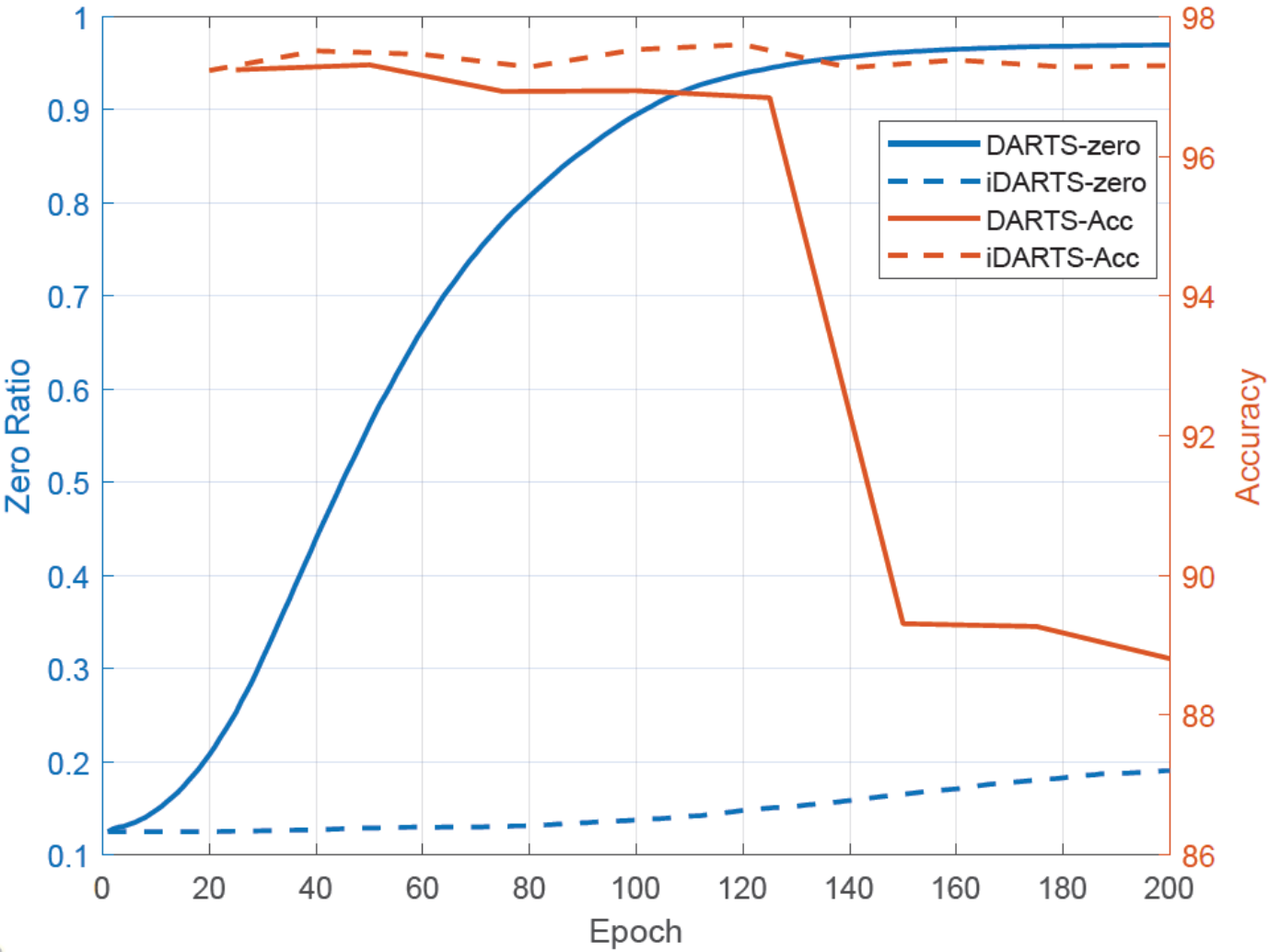}
	\caption{Curves in terms of \textit{zero}-ratio and accuracy of DARTS and iDARTS in the searching process when the epoch number increases to 200.}
\label{fig:epoch200}
\end{figure}

We extend the number of search epochs to 200 and re-train the discreted architecture in the searching process. The architectures approximated by iDARTS are illustrated in Fig. \ref{fig:idarts200}. The averaged \textit{zero}-ratio in the architecture parameters and the accuracy are displayed in Fig. \ref{fig:epoch200}.

From Fig. \ref{fig:epoch200}, we can see that the average weight of \textit{zero} operations over 14 edges increases to an abnormal value (0.98) at 200 epochs. This phenomenon also appears in \cite{DBLP:journals/corr/abs-1910-11831}. The retraining accuracy drops consistently when epochs become larger.

When node normalization is adopted, the average weight of \textit{zero} operations slightly increases in terms of epoch, and it stops in 0.18, which is a much more reasonable value. Besides, benefiting from decorrelation discretization, the retraining accuracy of iDARTS is more stable. The highest accuracy of 97.6\% is achieved at 120 epochs and a slight drop occurs when the searching process goes to more epochs, finally reaching an accuracy of 97.3\% at 200 epochs. This slight accuracy drop is caused by the difference in data distribution between the training and validation sets.  Ideally, the alternate updating on both the datasets drives ($w$, $\alpha$) to converge towards the global optimal ($w^*$, $\alpha^*$). However, if the number of epochs is set at a large value (\emph{e.g.} 200), $w$ tends to overfit to the sub-optimal values $w_{train}^\dag$ on the training set, which then affects the updating process of $\alpha$, leading to unstable results (\emph{i.e.} \textit{zero}-ratio starts to increase). When comparing DARTS and iDARTS, DARTS degrades at the beginning while iDARTS delivers a much more stable performance during iteration due to node normalization and decorrelation discretization. We visualize those architectures given by DARTS and iDARTS in Fig. \ref{fig:darts200} and Fig. \ref{fig:idarts200} in Appendix.

\subsection{Results on ImageNet}

We further evaluate the architectures reached by iDARTS on ImageNet. For comparison, two architectures are taken, which are searched with 4 nodes in a single cell on CIFAR-10 and ImageNet respectively. The architectures are demonstrated in Fig. \ref{fig:S0}. Results are shown in Table \ref{T.ImageNet}.

As in Table \ref{T.ImageNet}, we can find that both the architectures learned on CIFAR-10 and ImageNet achieve comparable performance with the state-of-the-art manual or RL methods. The architecture learned on CIFAR-10 reaches a top-1 error of 25.2\%. When directly searching on ImageNet, the top-1 error furtherly decreases to 24.7\%, demonstrating its effectiveness.

\section{Conclusion}
\label{S.CON}
In this paper, we propose an improved version of DARTS, namely iDARTS, to address the architecture degradation issue. Our motivation lies in that DARTS-based approaches neglect the imbalanced norms between different nodes and the high correlation between operations. We then introduce the node normalization and decorrelation discretization strategies to solve such problems. Our approach delivers better performance with stronger generalization ability as well as stability.

%\bibliographystyle{IEEEtran}
%\bibliography{example_paper}

\begin{figure*}[!htbp]
%\begin{center}
%\end{center}
	\subfigure[Normal cell of original DARTS]{
    \begin{minipage}[h]{0.23\linewidth}
    \centering
      \includegraphics[width = 1\linewidth]{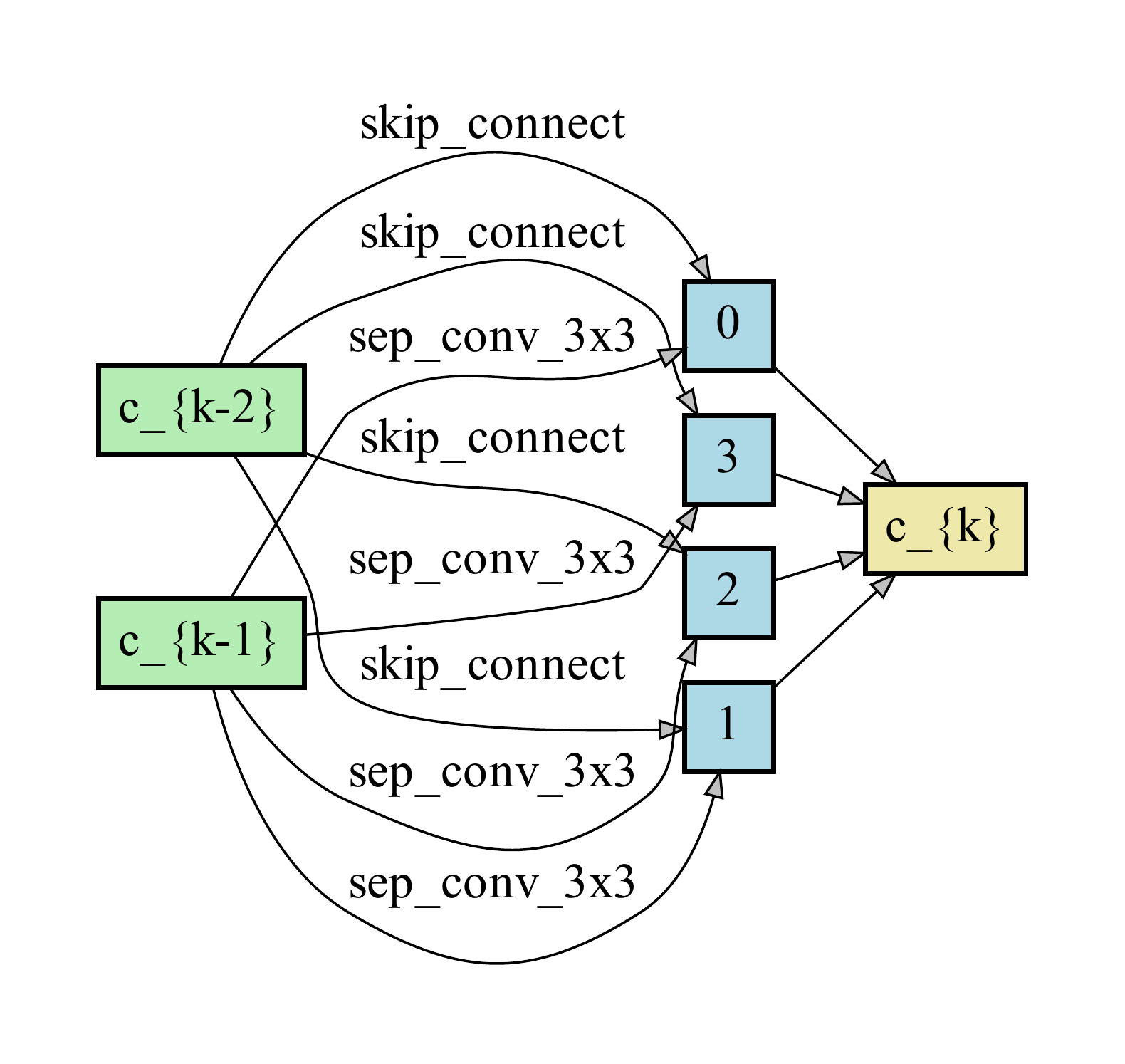}
    \end{minipage}
    }
    \subfigure[Reduction cell of original DARTS]{
    \begin{minipage}[h]{0.23\linewidth}
    \centering
      \includegraphics[width= 1\linewidth]{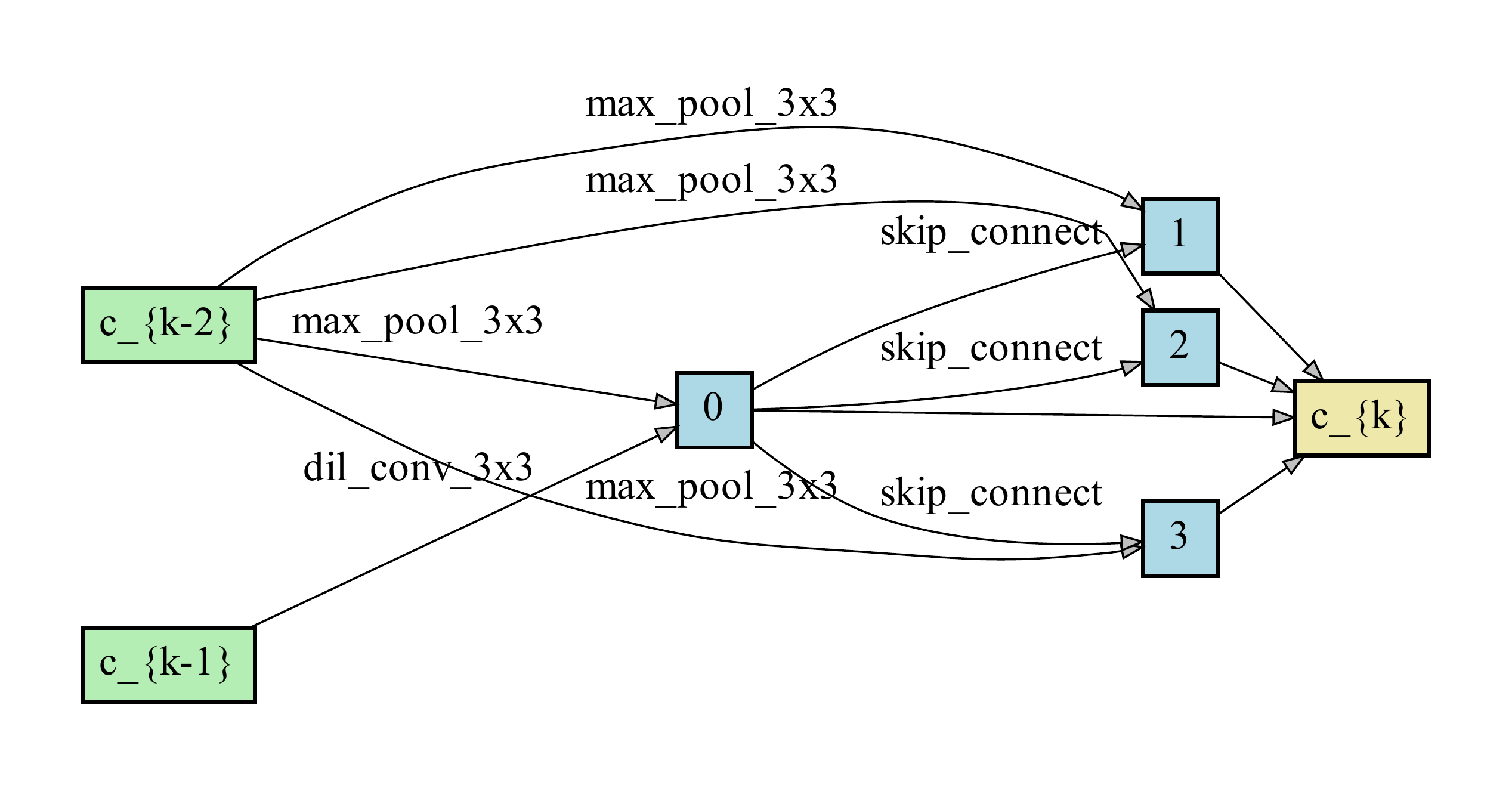}
    \end{minipage}
    }
    \subfigure[Normal cell of DARTS with NN]{
    \begin{minipage}[h]{0.23\linewidth}
    \centering
      \includegraphics[width = 1\linewidth]{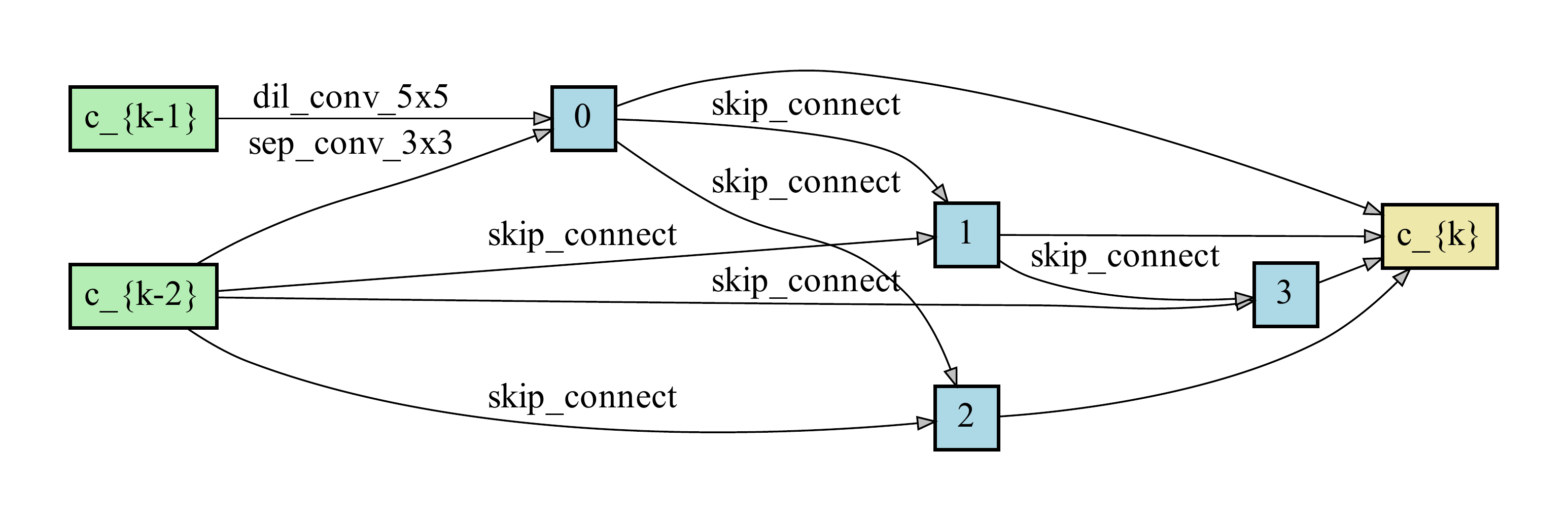}
    \end{minipage}
    }
    \subfigure[Reduction cell of DARTS with NN]{
    \begin{minipage}[h]{0.23\linewidth}
    \centering
      \includegraphics[width= 1\linewidth]{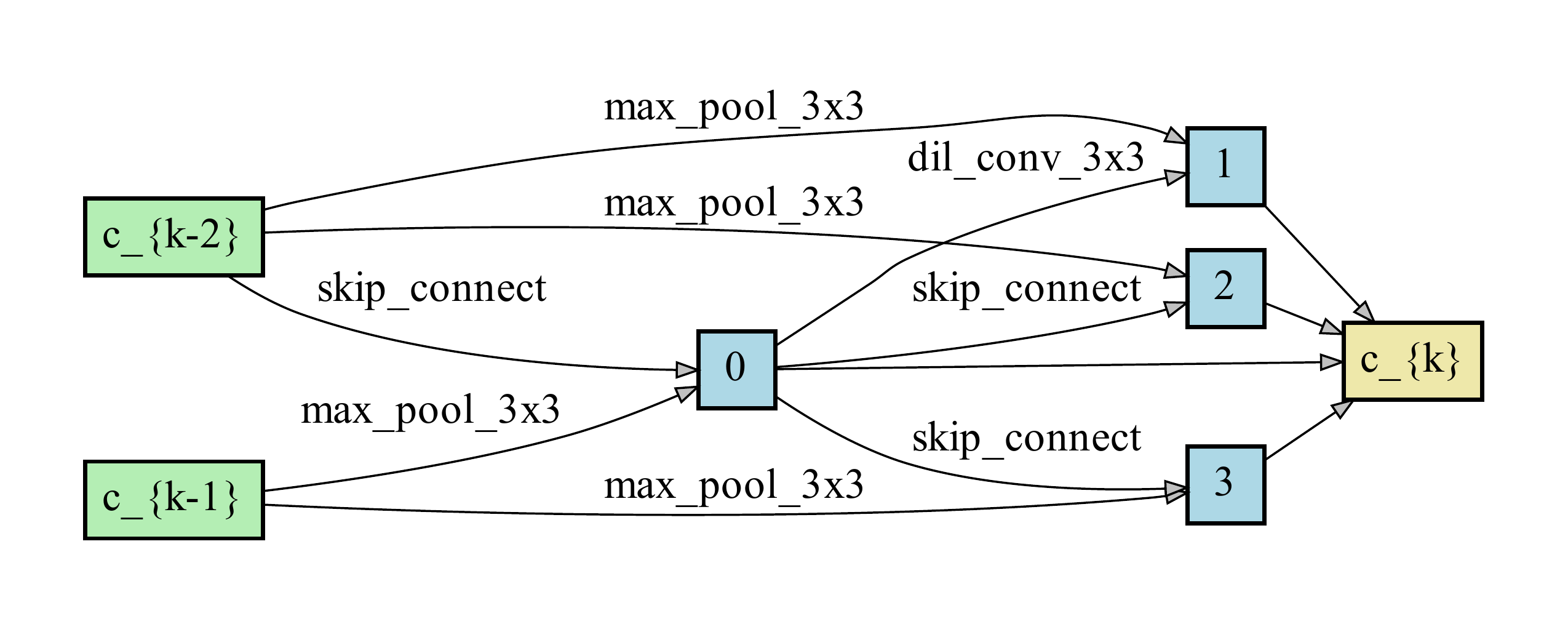}
    \end{minipage}
    }
    \subfigure[Normal cell of DARTS with DD]{
    \begin{minipage}[h]{0.23\linewidth}
    \centering
      \includegraphics[width = 1\linewidth]{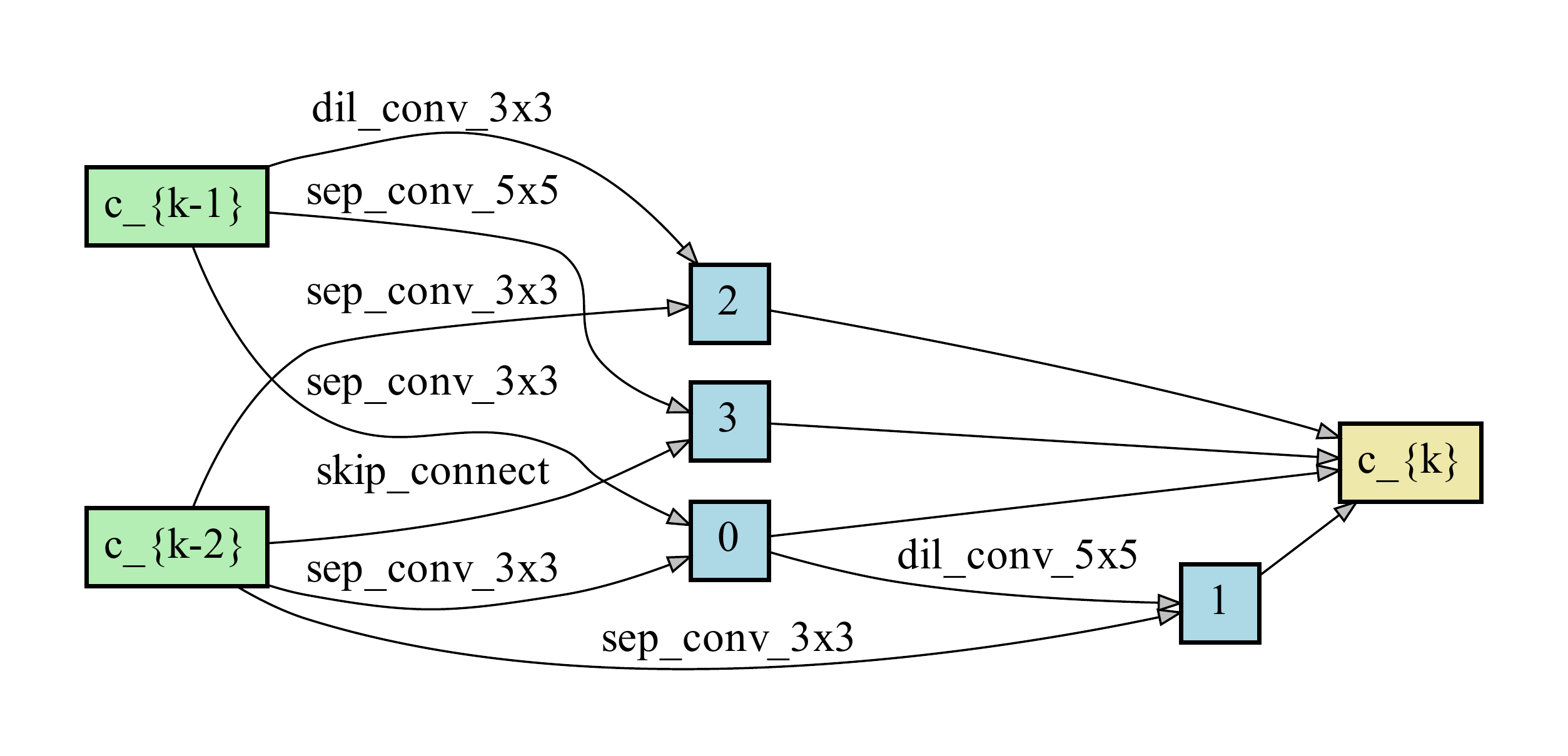}
    \end{minipage}
    }
    \subfigure[Reduction cell of DARTS with DD]{
    \begin{minipage}[h]{0.23\linewidth}
    \centering
      \includegraphics[width= 1\linewidth]{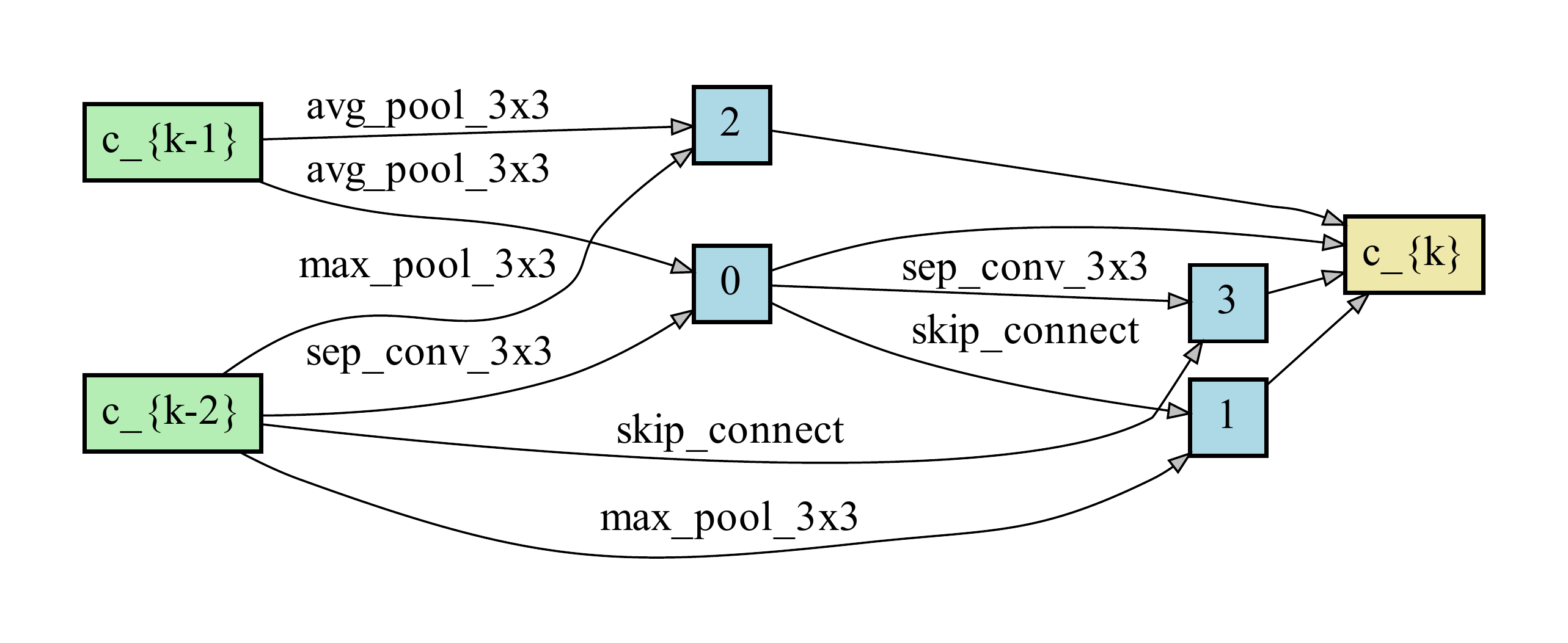}
    \end{minipage}
    }
    \subfigure[Normal cell of iDARTS]{
    \begin{minipage}[h]{0.23\linewidth}
    \centering
      \includegraphics[width = 1\linewidth]{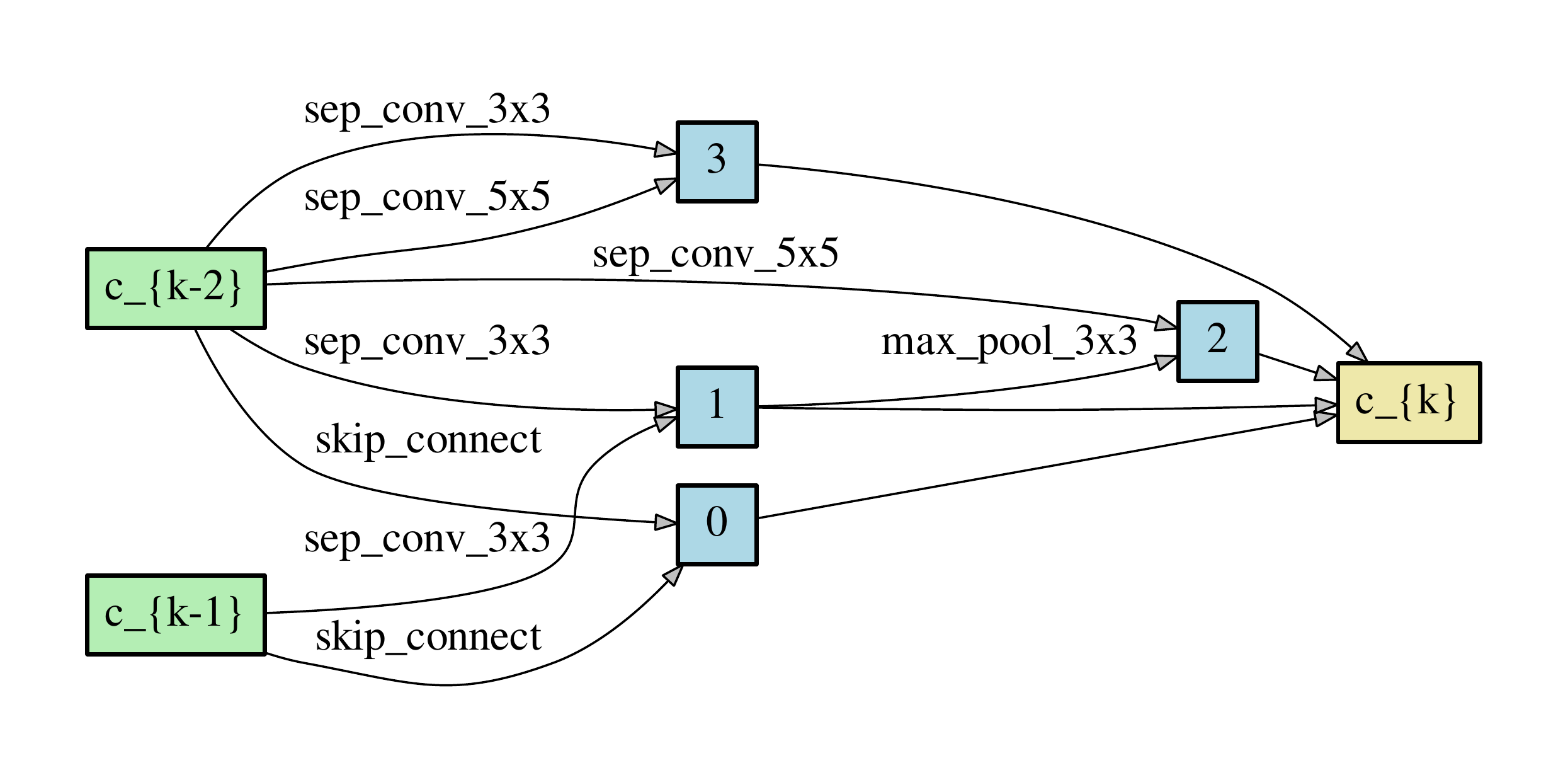}
    \end{minipage}
    }
    \subfigure[Reduction cell of iDARTS]{
    \begin{minipage}[h]{0.23\linewidth}
    \centering
      \includegraphics[width= 1\linewidth]{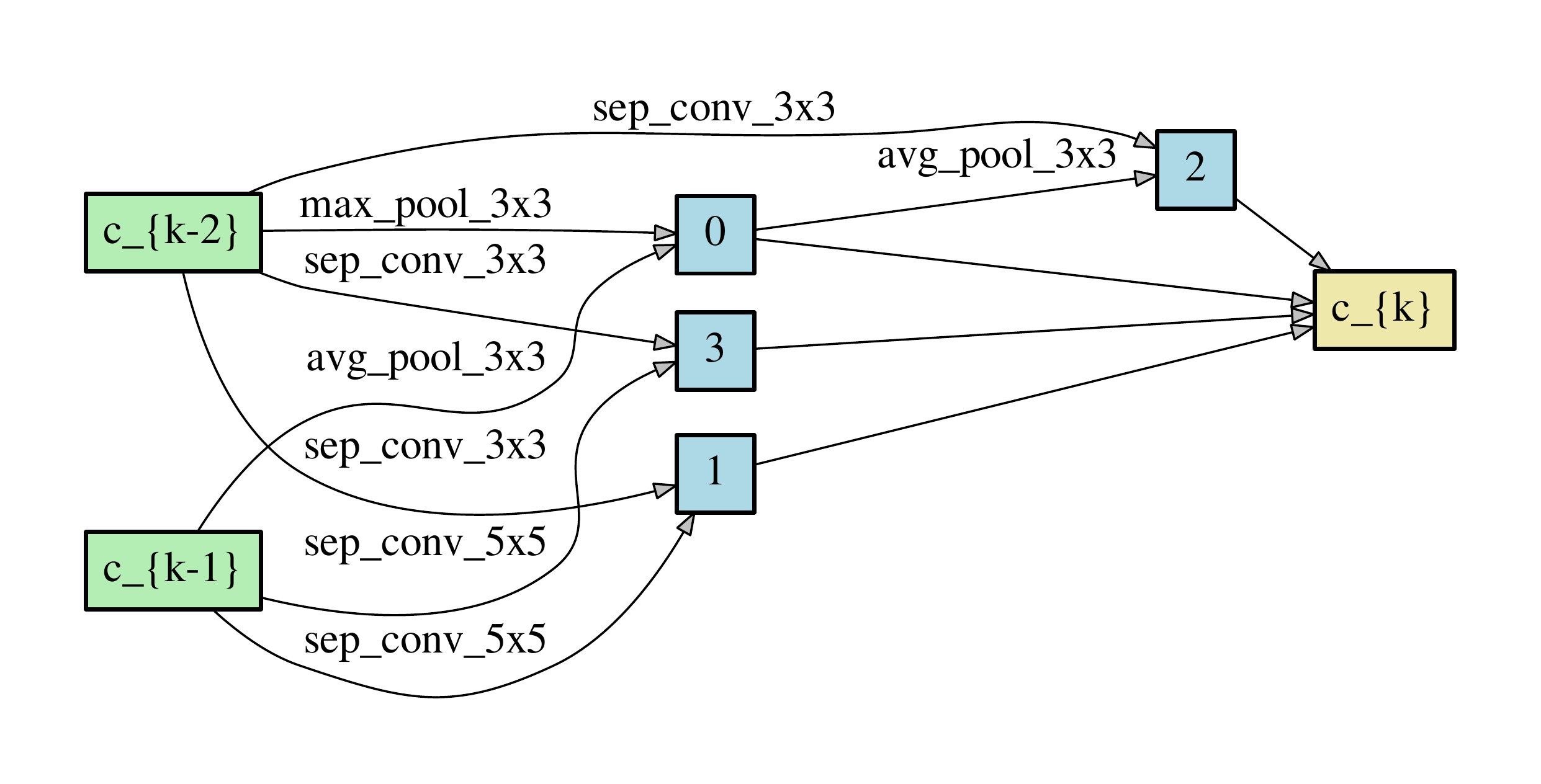}
    \end{minipage}
    }
   \caption{Normal and reduction cells given by iDARTS and DARTS on CIFAR-10 w/ or w/o NN and DD in $S1$.}
\label{fig:ablation}
\end{figure*}
\begin{figure*}[!htbp]
%\begin{center}
%\end{center}
	\subfigure[Normal cell at epoch 25]{
    \begin{minipage}[h]{0.23\linewidth}
    \centering
      \includegraphics[width = 1\linewidth]{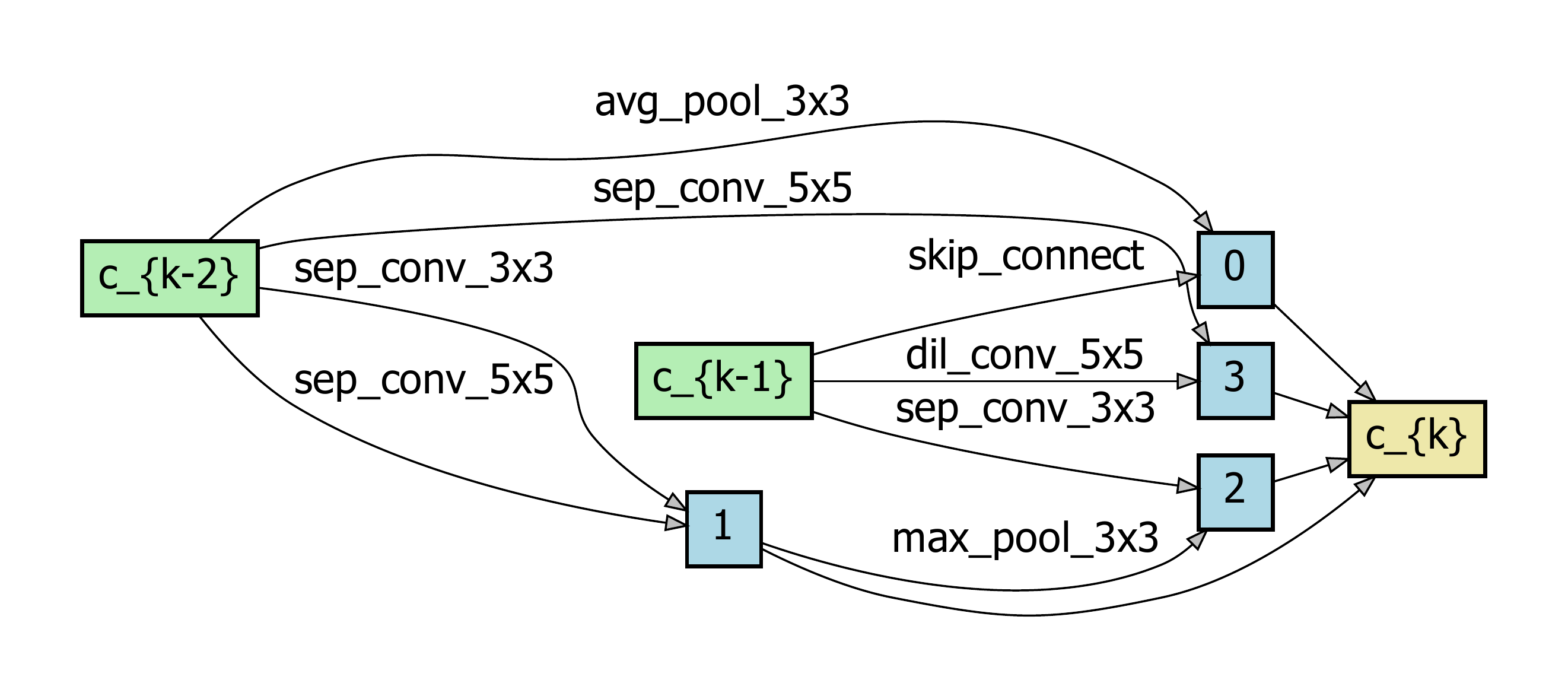}
    \end{minipage}
    }
    \subfigure[Reduction cell at epoch 25]{
    \begin{minipage}[h]{0.23\linewidth}
    \centering
      \includegraphics[width= 1\linewidth]{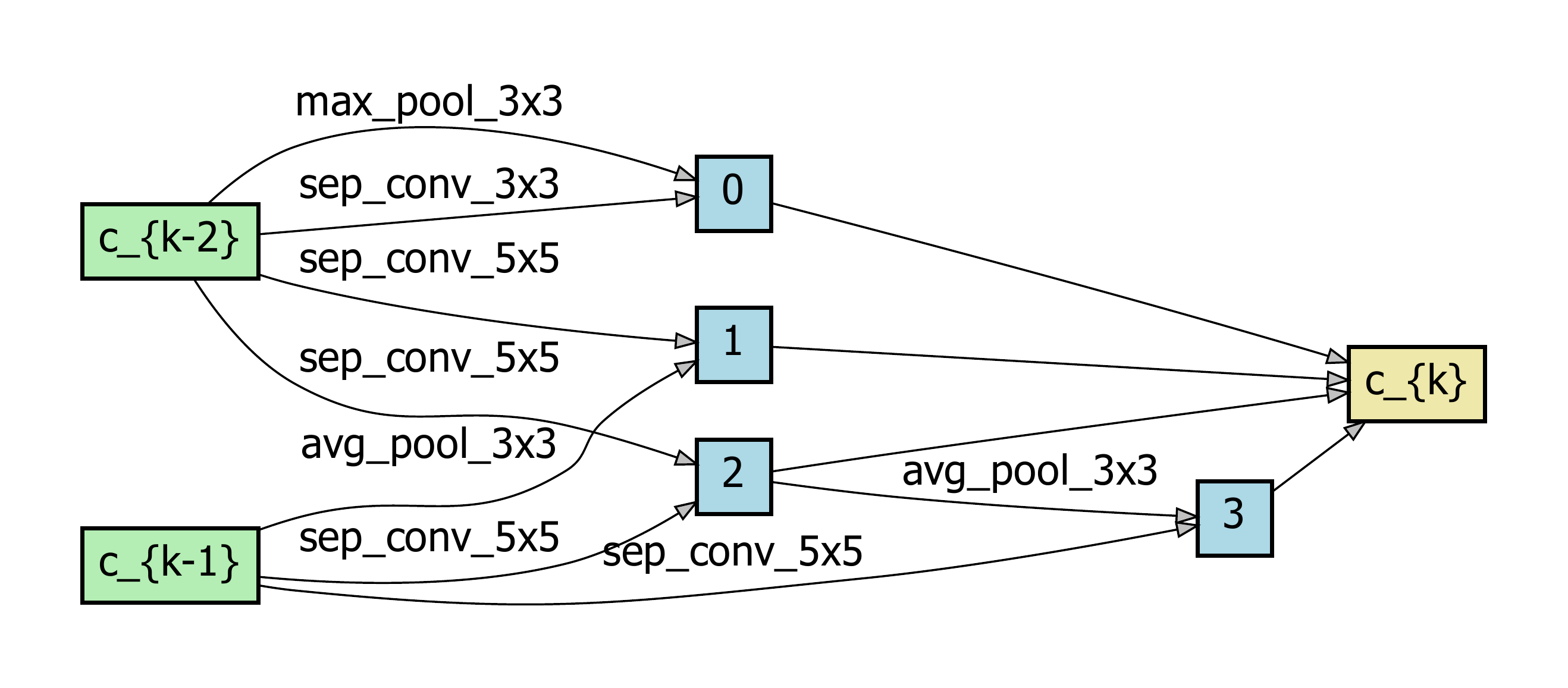}
    \end{minipage}
    }
    \subfigure[Normal cell at epoch 50]{
    \begin{minipage}[h]{0.23\linewidth}
    \centering
      \includegraphics[width = 1\linewidth]{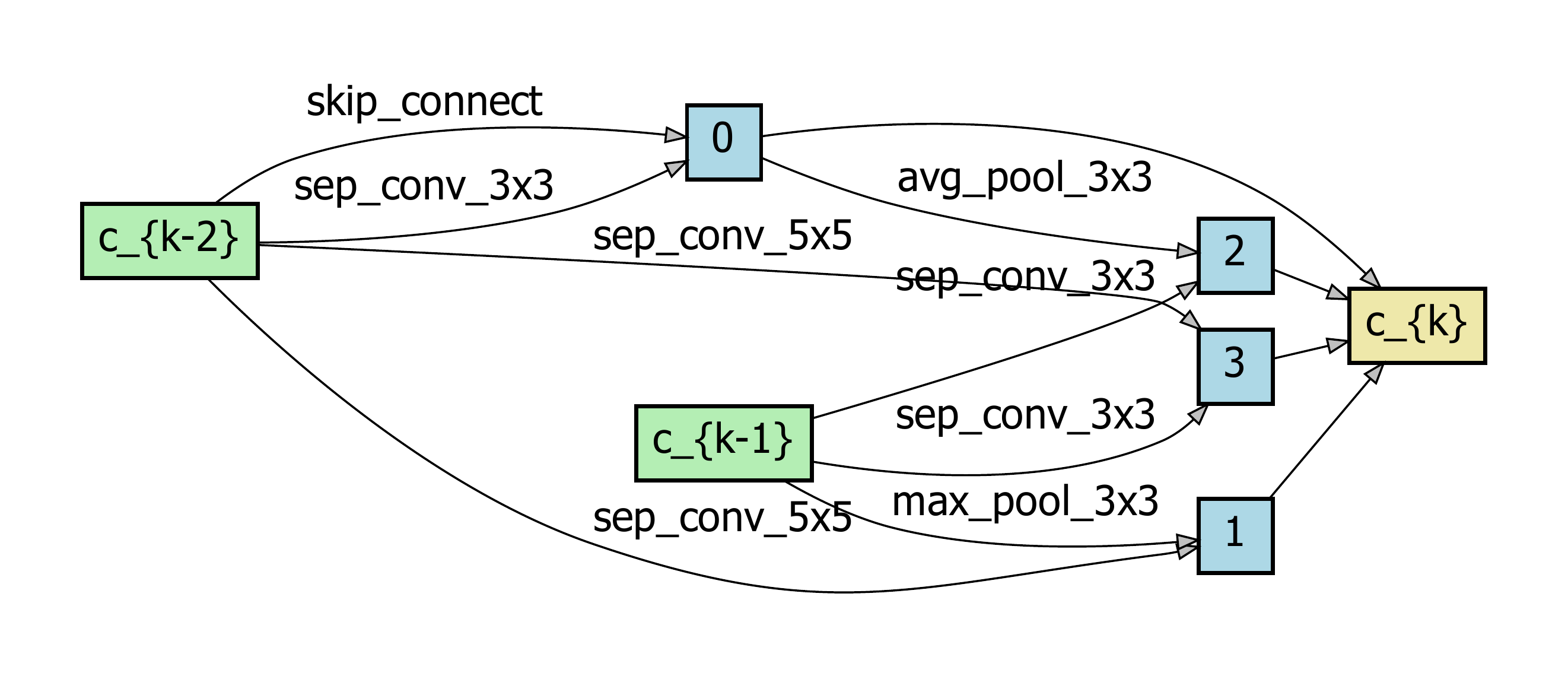}
    \end{minipage}
    }
    \subfigure[Reduction cell at epoch 50]{
    \begin{minipage}[h]{0.23\linewidth}
    \centering
      \includegraphics[width= 1\linewidth]{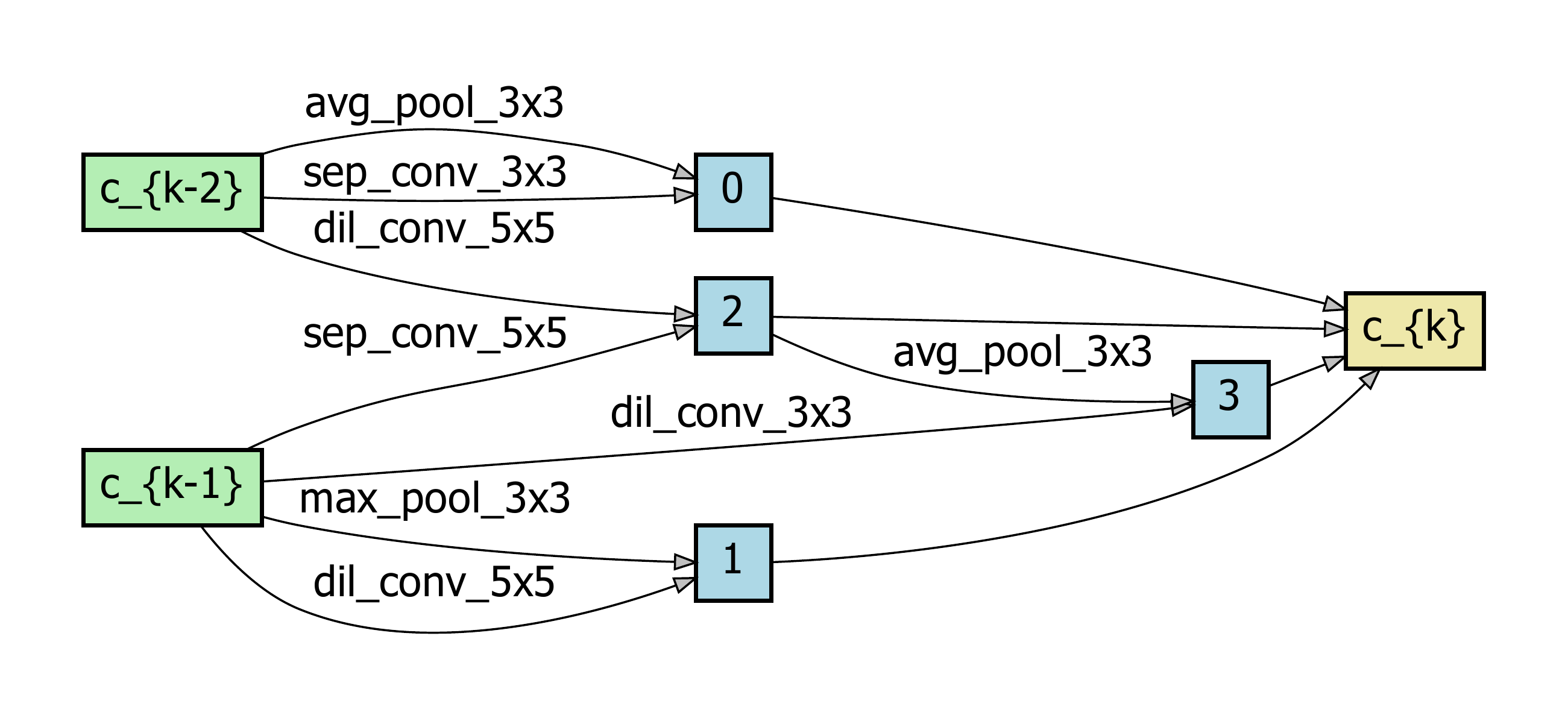}
    \end{minipage}
    }
    \subfigure[Normal cell at epoch 75]{
    \begin{minipage}[h]{0.23\linewidth}
    \centering
      \includegraphics[width = 1\linewidth]{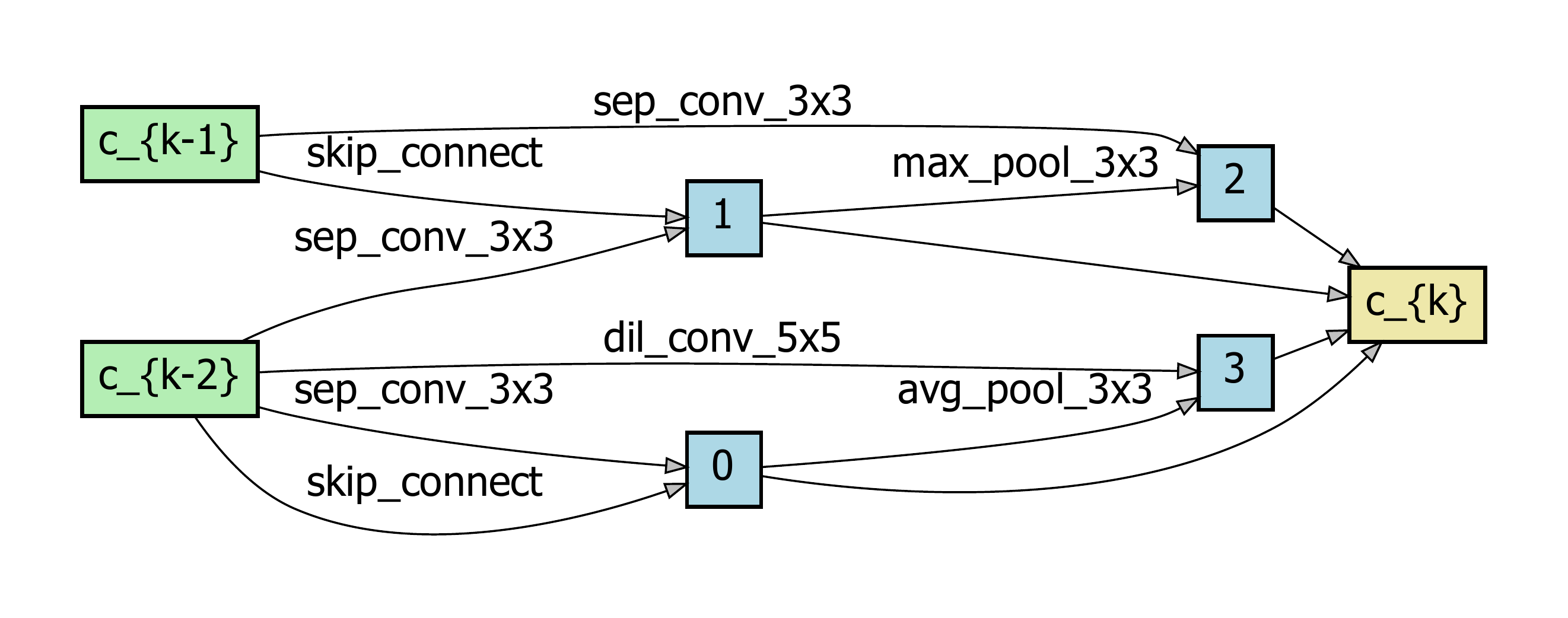}
    \end{minipage}
    }
    \subfigure[Reduction cell at epoch 75]{
    \begin{minipage}[h]{0.23\linewidth}
    \centering
      \includegraphics[width= 1\linewidth]{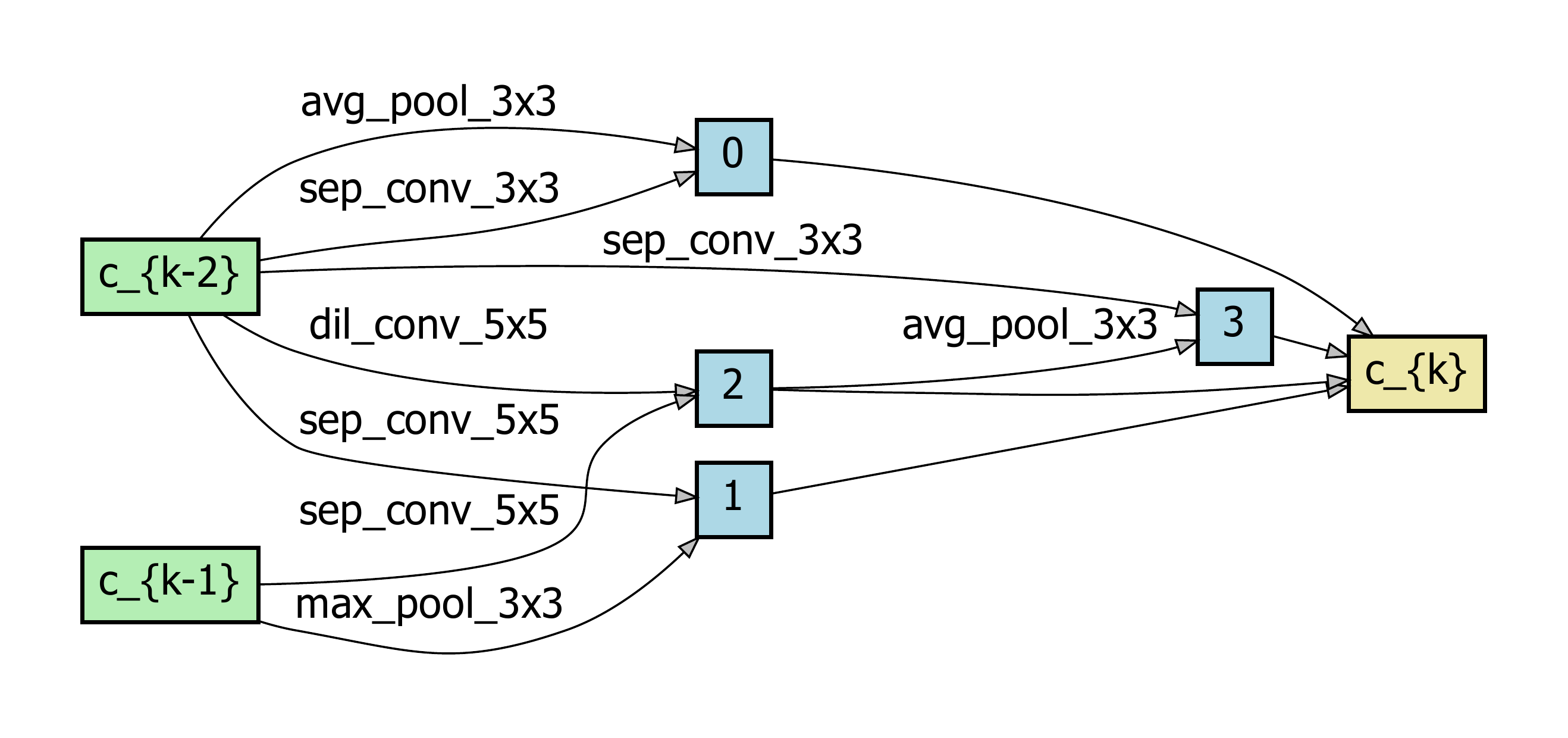}
    \end{minipage}
    }
    \subfigure[Normal cell at epoch 100]{
    \begin{minipage}[h]{0.23\linewidth}
    \centering
      \includegraphics[width = 1\linewidth]{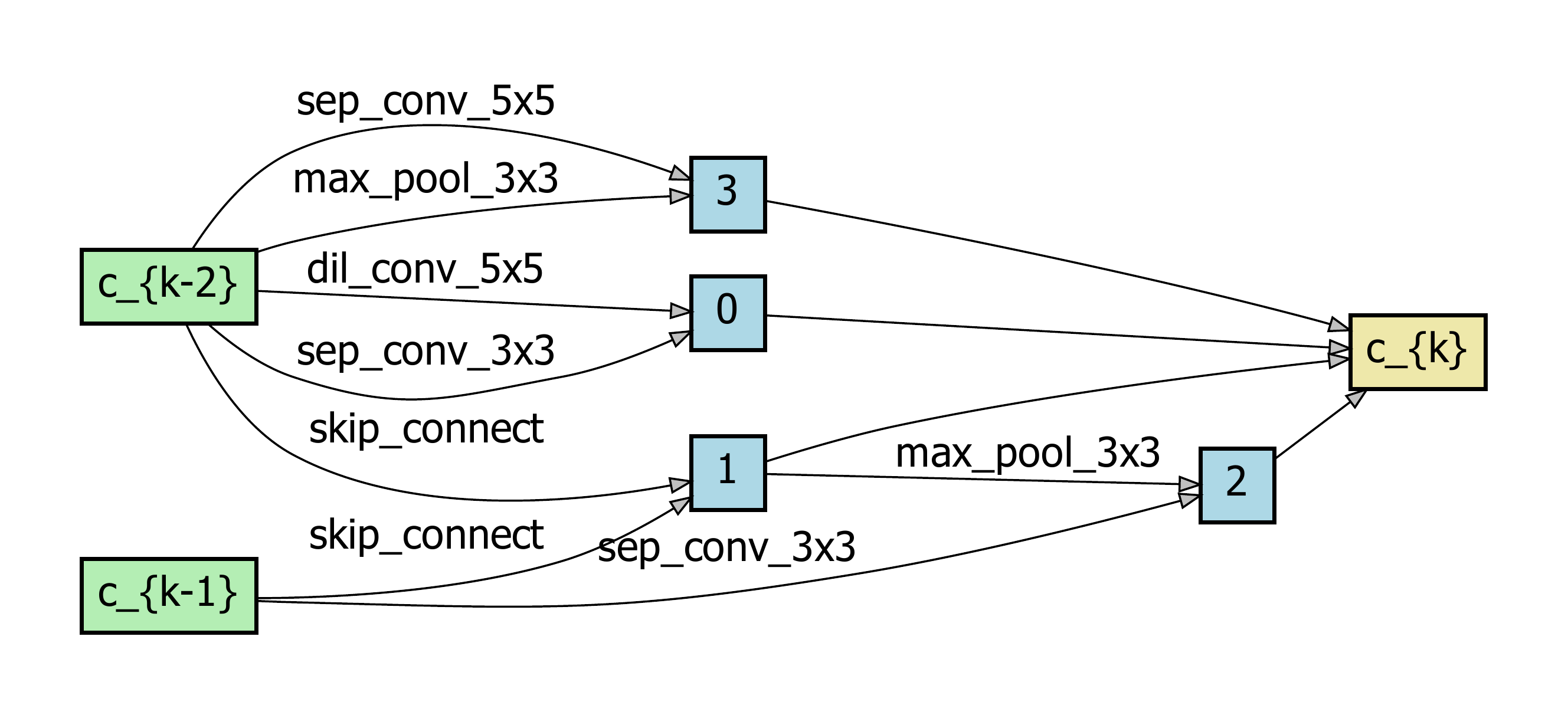}
    \end{minipage}
    }
    \subfigure[Reduction cell at epoch 100]{
    \begin{minipage}[h]{0.23\linewidth}
    \centering
      \includegraphics[width= 1\linewidth]{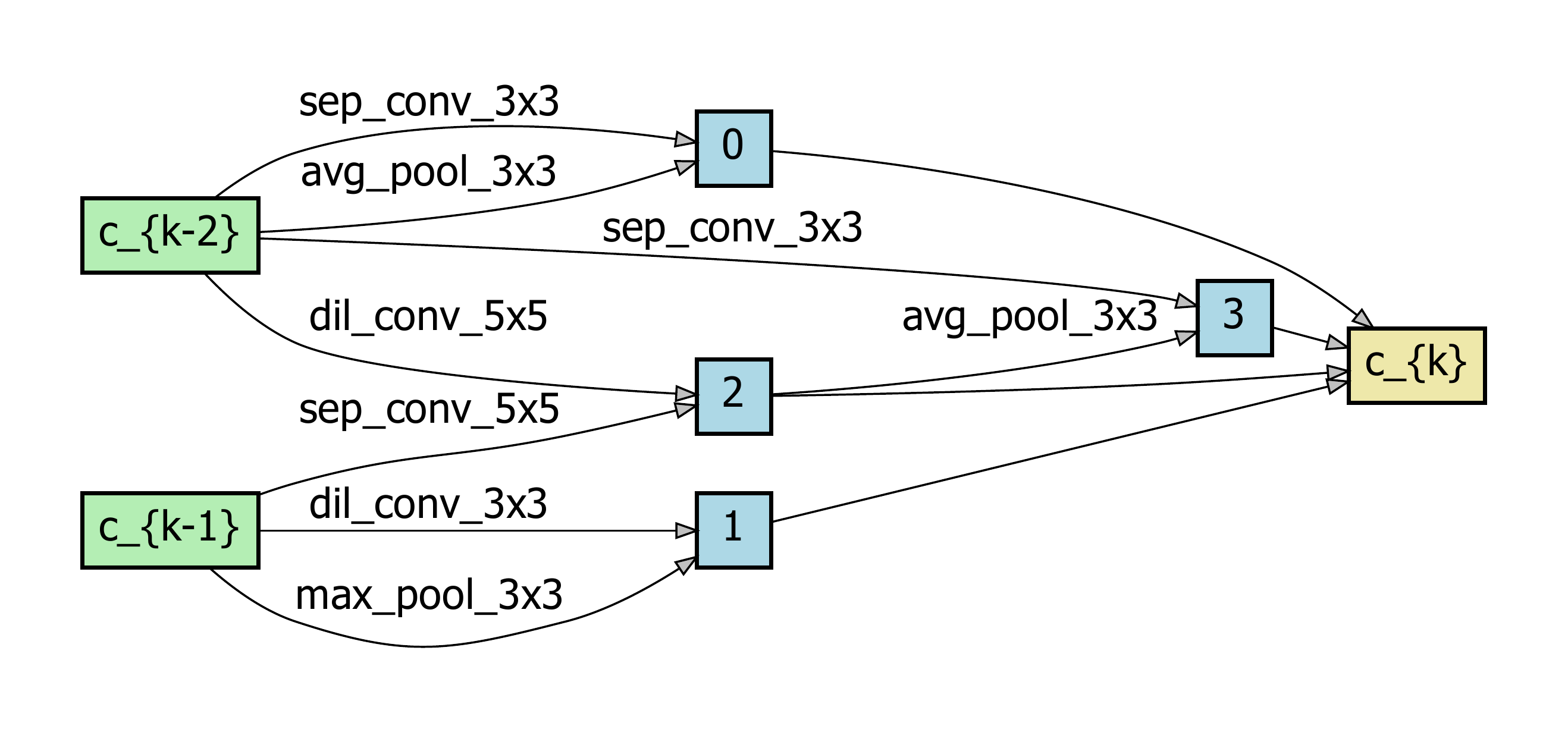}
    \end{minipage}
    }
    \subfigure[Normal cell at epoch 125]{
    \begin{minipage}[h]{0.23\linewidth}
    \centering
      \includegraphics[width = 1\linewidth]{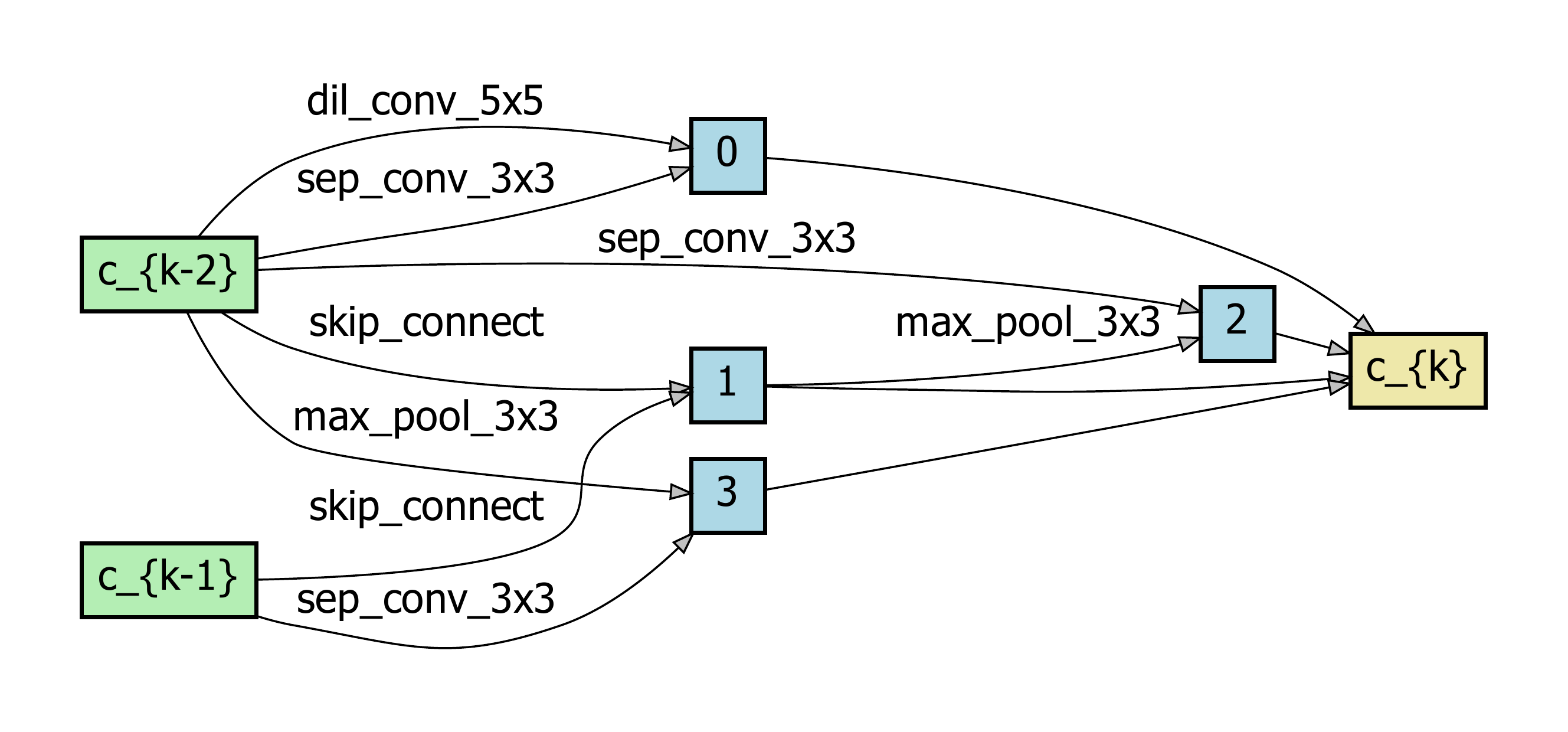}
    \end{minipage}
    }
    \subfigure[Reduction cell at epoch 125]{
    \begin{minipage}[h]{0.23\linewidth}
    \centering
      \includegraphics[width= 1\linewidth]{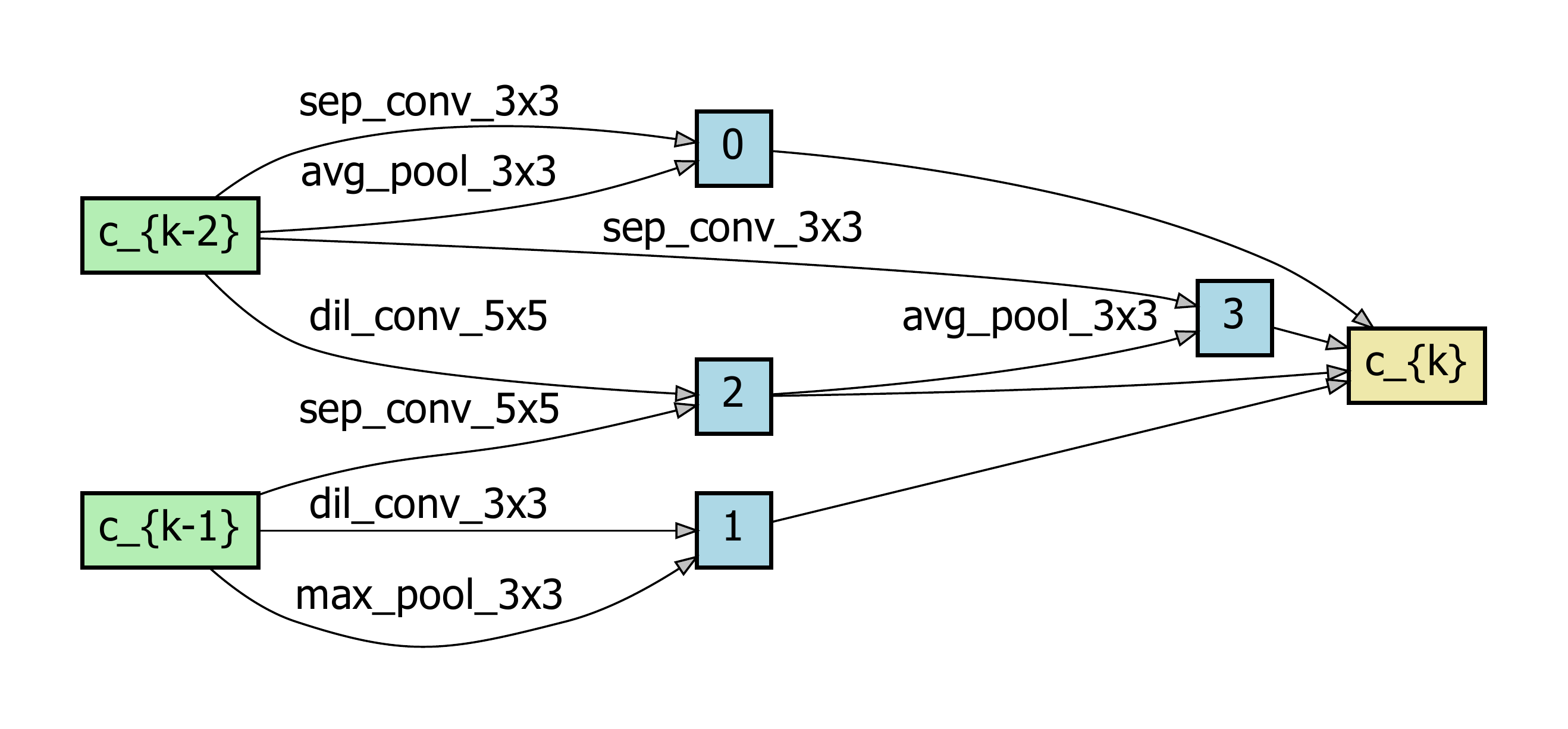}
    \end{minipage}
    }
    \subfigure[Normal cell at epoch 150]{
    \begin{minipage}[h]{0.23\linewidth}
    \centering
      \includegraphics[width = 1\linewidth]{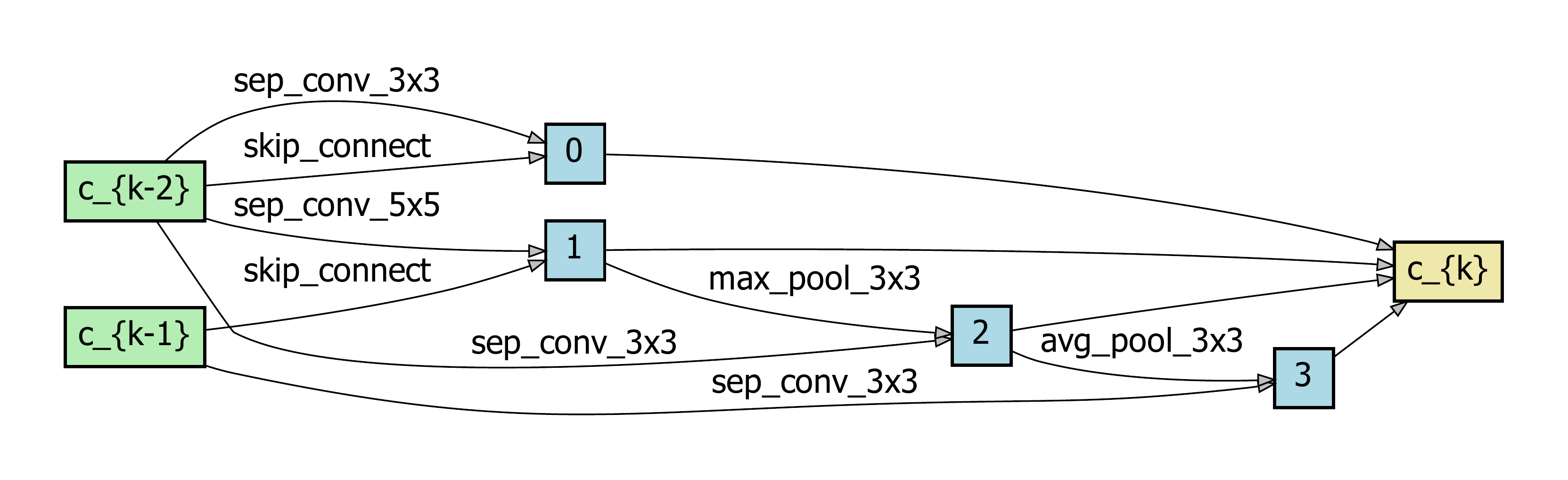}
    \end{minipage}
    }
    \subfigure[Reduction cell at epoch 150]{
    \begin{minipage}[h]{0.23\linewidth}
    \centering
      \includegraphics[width= 1\linewidth]{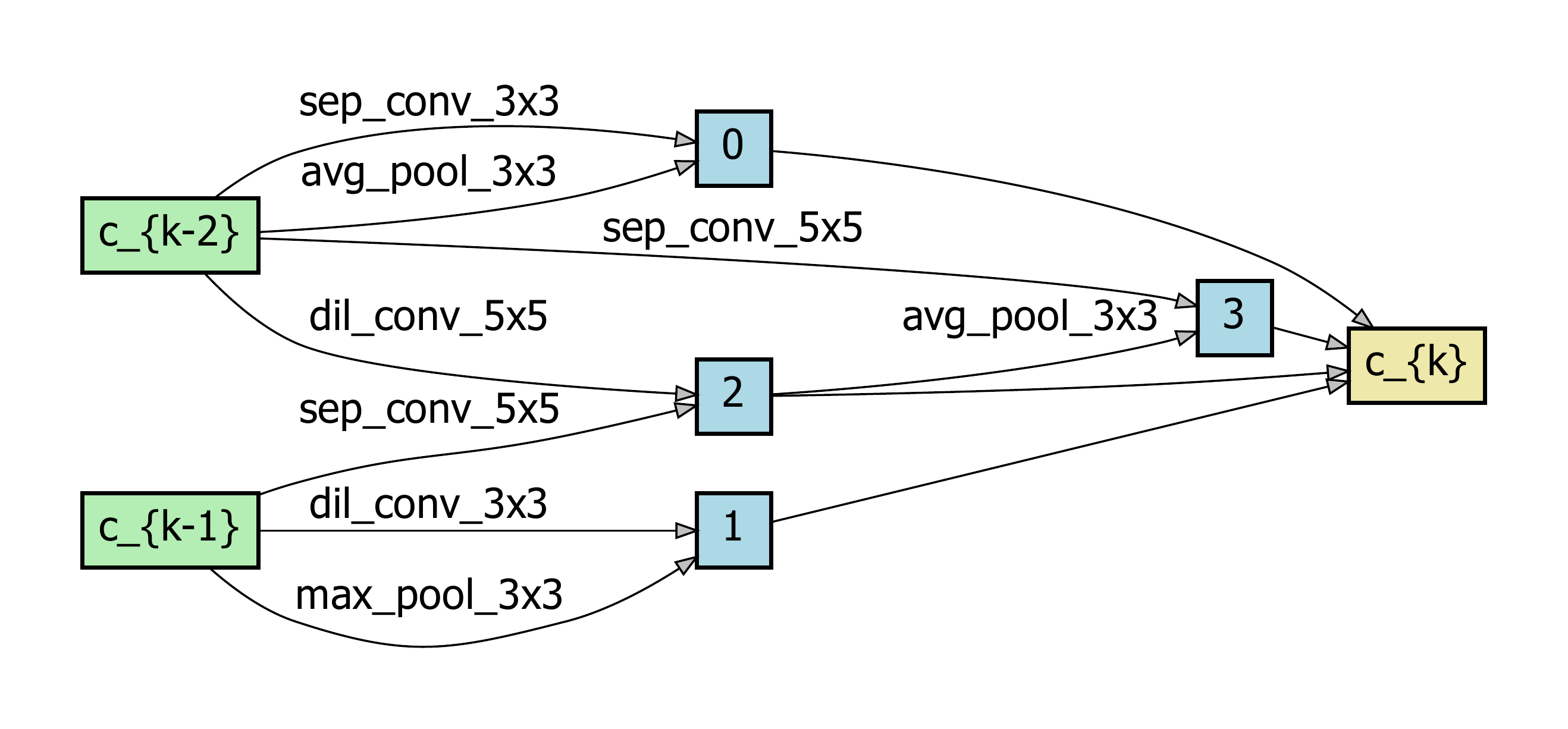}
    \end{minipage}
    }\\
    \subfigure[Normal cell at epoch 175]{
    \begin{minipage}[h]{0.23\linewidth}
    \centering
      \includegraphics[width = 1\linewidth]{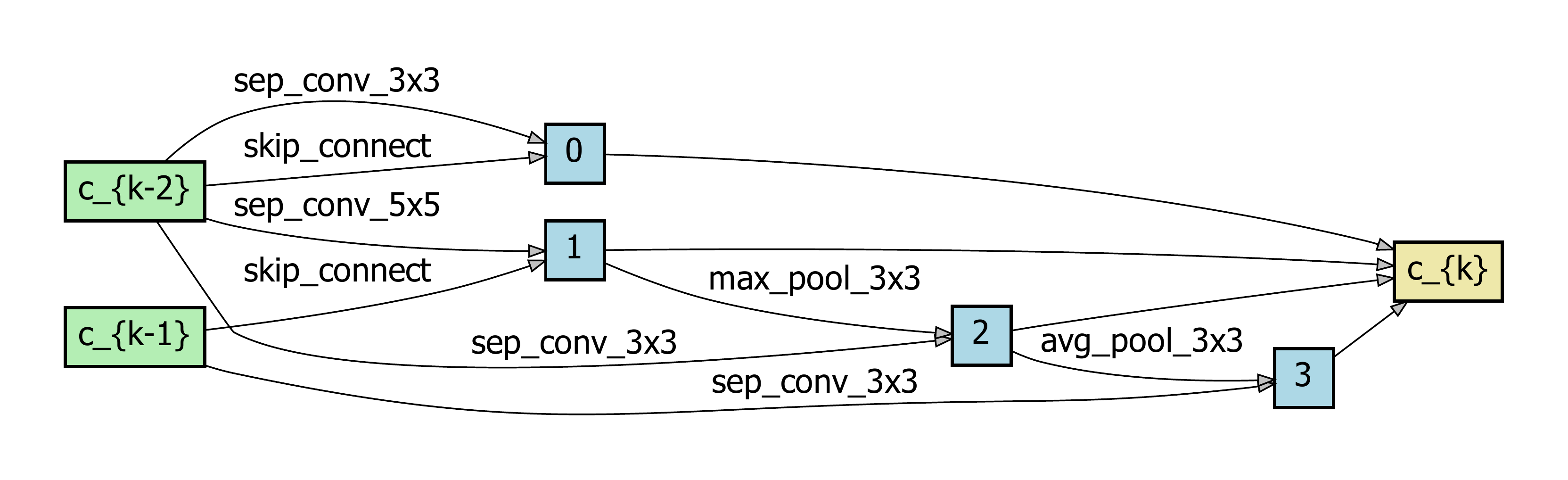}
    \end{minipage}
    }
    \subfigure[Reduction cell at epoch 175]{
    \begin{minipage}[h]{0.23\linewidth}
    \centering
      \includegraphics[width= 1\linewidth]{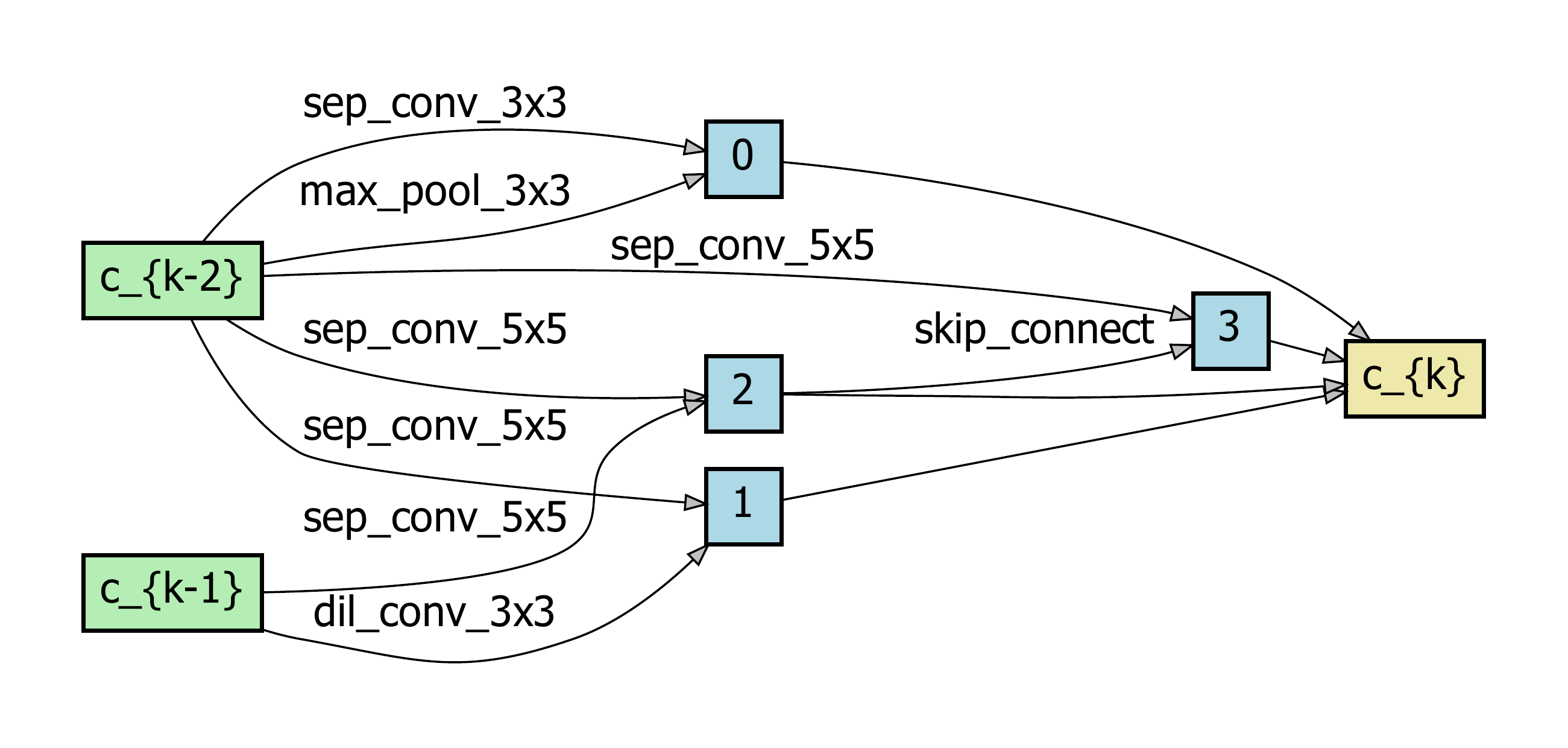}
    \end{minipage}
    }
    \subfigure[Normal cell at epoch 200]{
    \begin{minipage}[h]{0.23\linewidth}
    \centering
      \includegraphics[width = 1\linewidth]{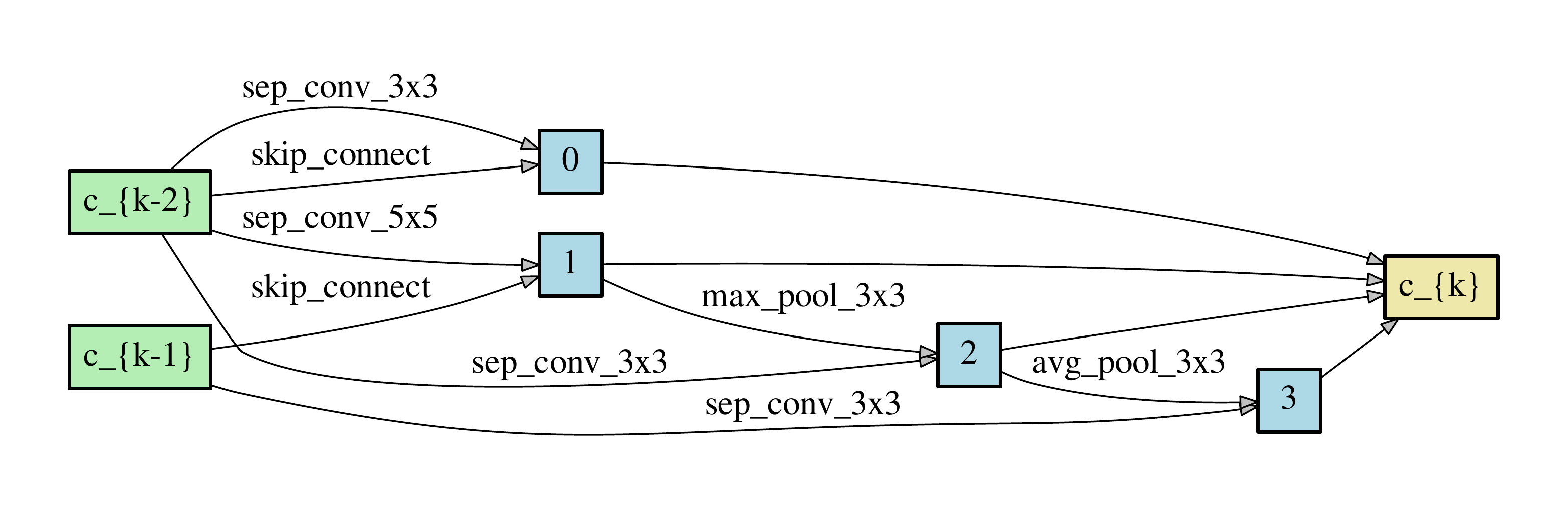}
    \end{minipage}
    }
    \subfigure[Reduction cell at epoch 200]{
    \begin{minipage}[h]{0.23\linewidth}
    \centering
      \includegraphics[width= 1\linewidth]{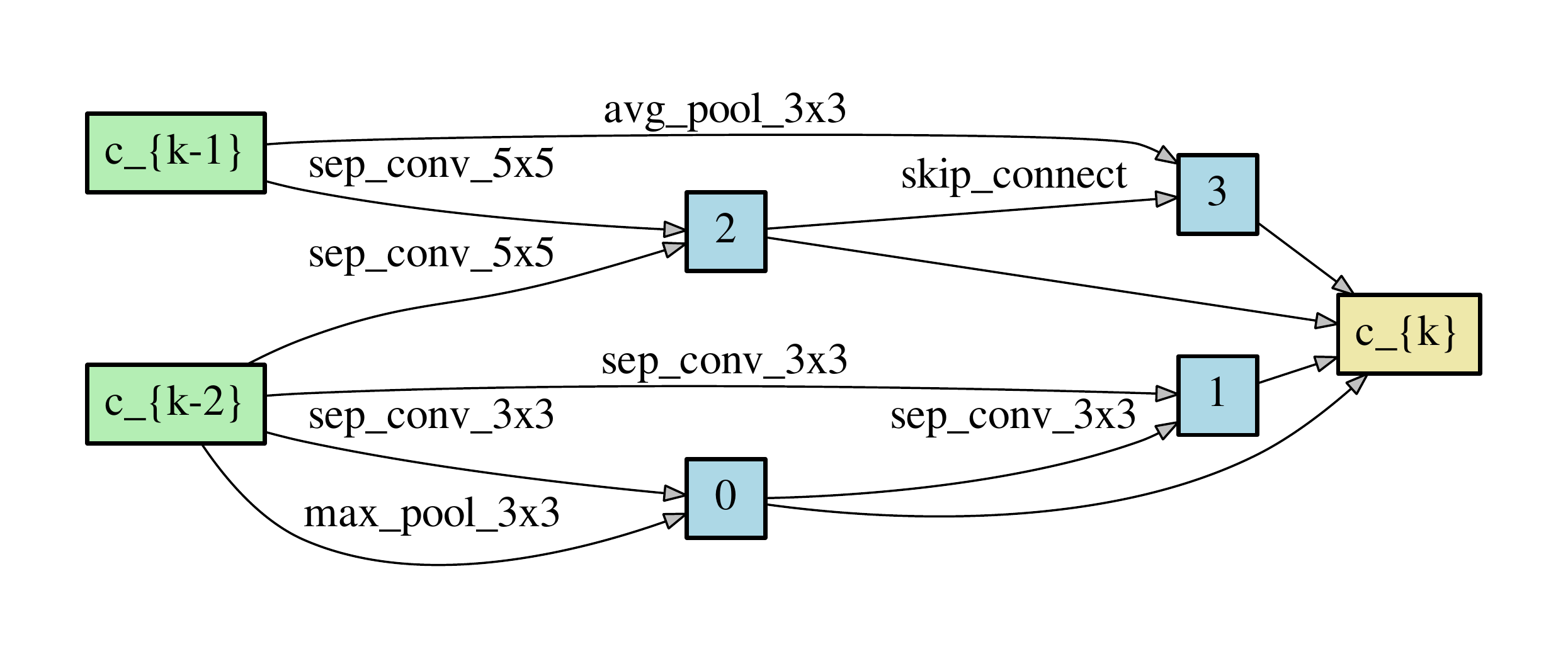}
    \end{minipage}
    }
   \caption{Normal and reduction cells given by iDARTS on CIFAR-10 in $S1$ when searching for 200 epochs.}
\label{fig:idarts200}
\end{figure*}
\begin{figure*}[!htbp]
%\begin{center}
%\end{center}
	\subfigure[Normal cell at epoch 25]{
    \begin{minipage}[h]{0.23\linewidth}
    \centering
      \includegraphics[width = 1\linewidth]{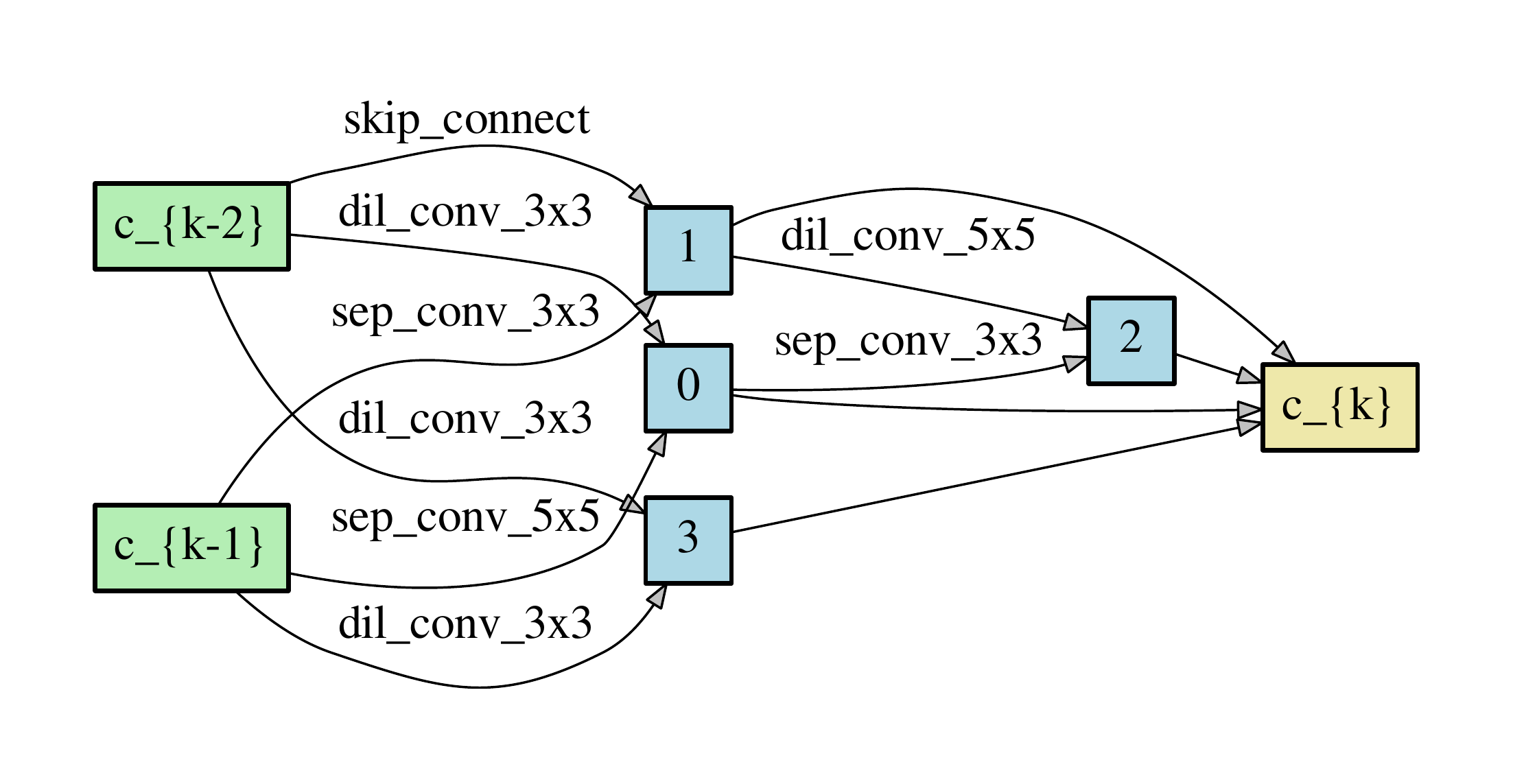}
    \end{minipage}
    }
    \subfigure[Reduction cell at epoch 25]{
    \begin{minipage}[h]{0.23\linewidth}
    \centering
      \includegraphics[width= 1\linewidth]{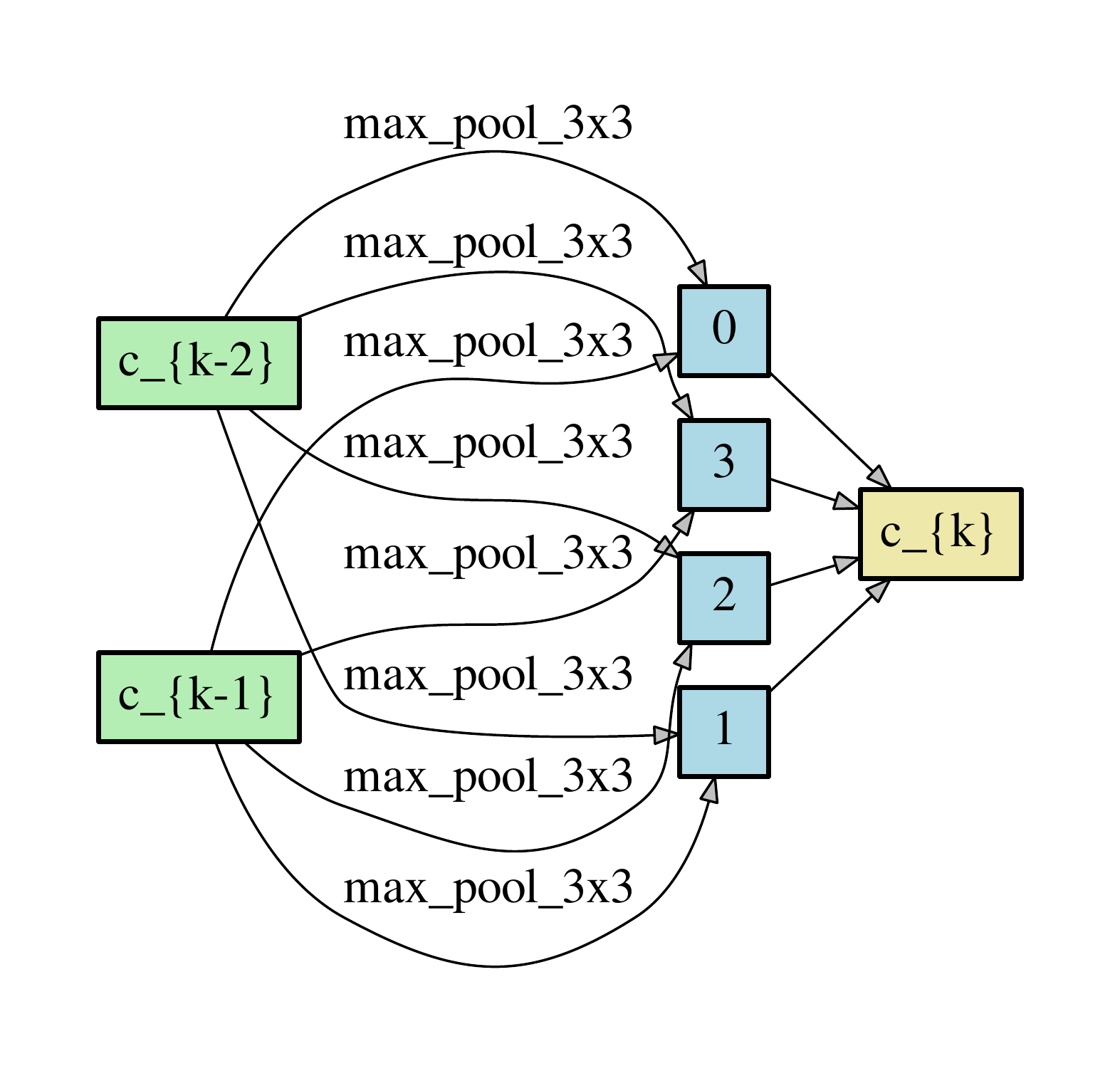}
    \end{minipage}
    }
    \subfigure[Normal cell at epoch 50]{
    \begin{minipage}[h]{0.23\linewidth}
    \centering
      \includegraphics[width = 1\linewidth]{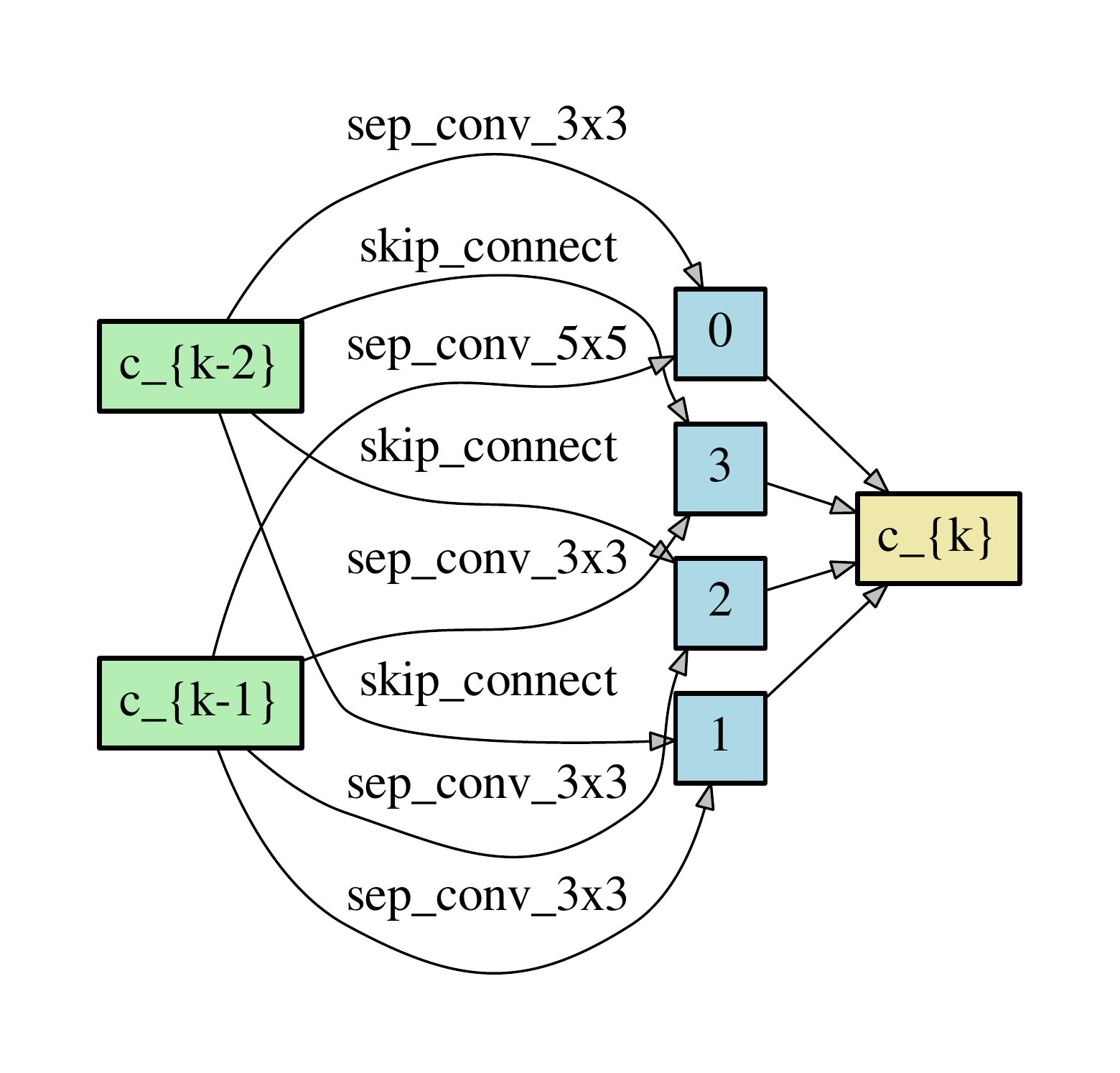}
    \end{minipage}
    }
    \subfigure[Reduction cell at epoch 50]{
    \begin{minipage}[h]{0.23\linewidth}
    \centering
      \includegraphics[width= 1\linewidth]{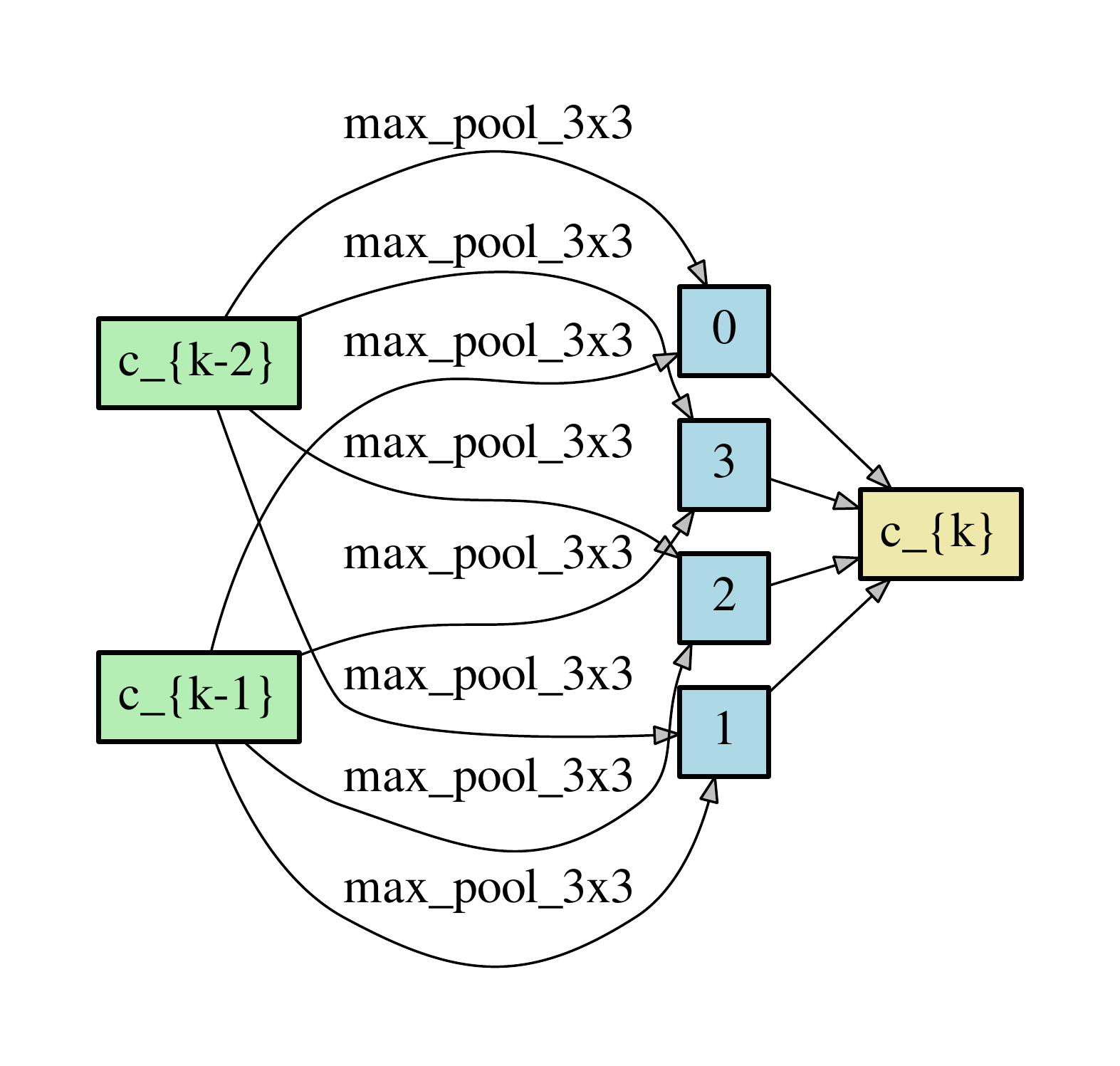}
    \end{minipage}
    }
    \subfigure[Normal cell at epoch 75]{
    \begin{minipage}[h]{0.23\linewidth}
    \centering
      \includegraphics[width = 1\linewidth]{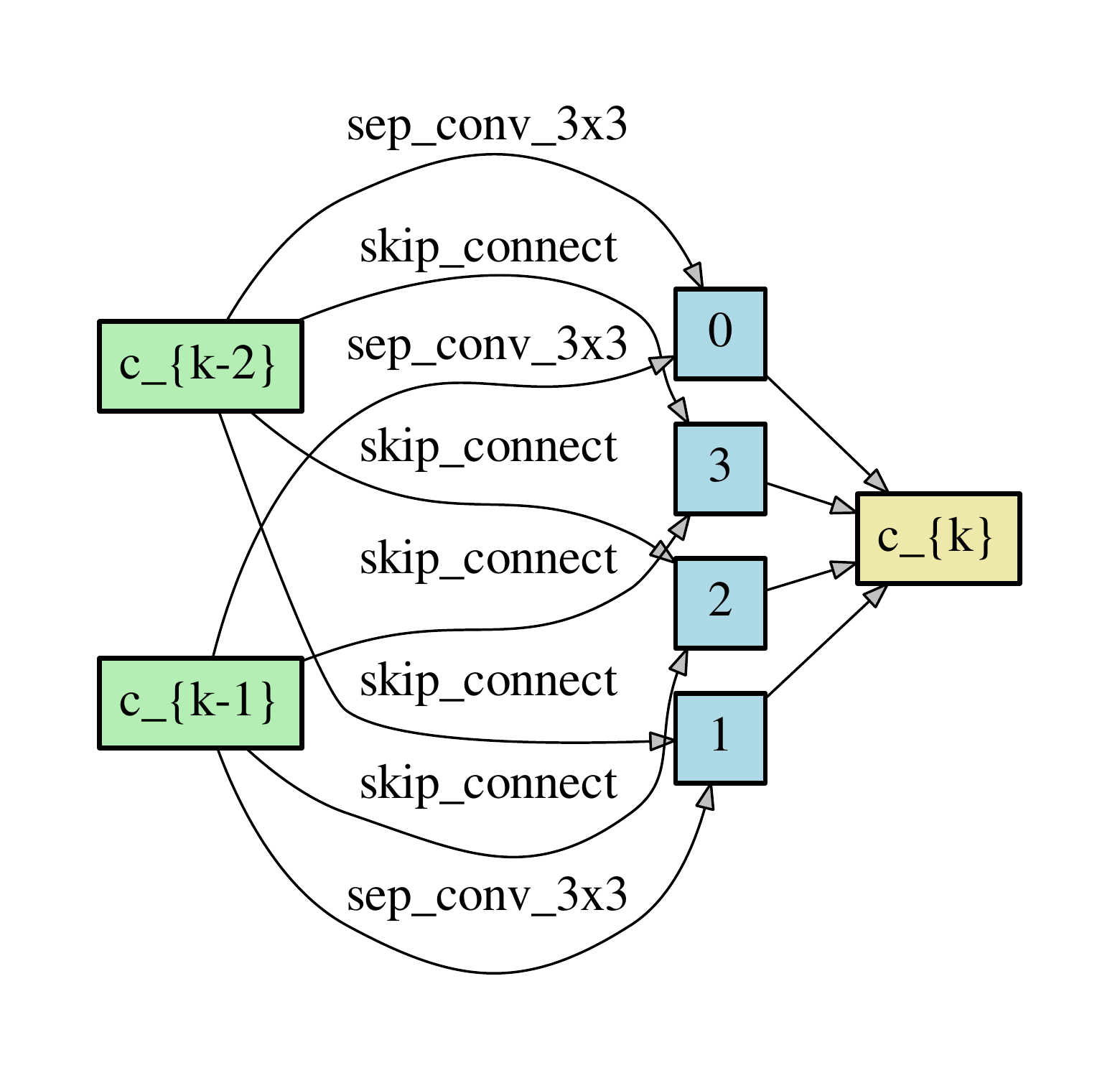}
    \end{minipage}
    }
    \subfigure[Reduction cell at epoch 75]{
    \begin{minipage}[h]{0.23\linewidth}
    \centering
      \includegraphics[width= 1\linewidth]{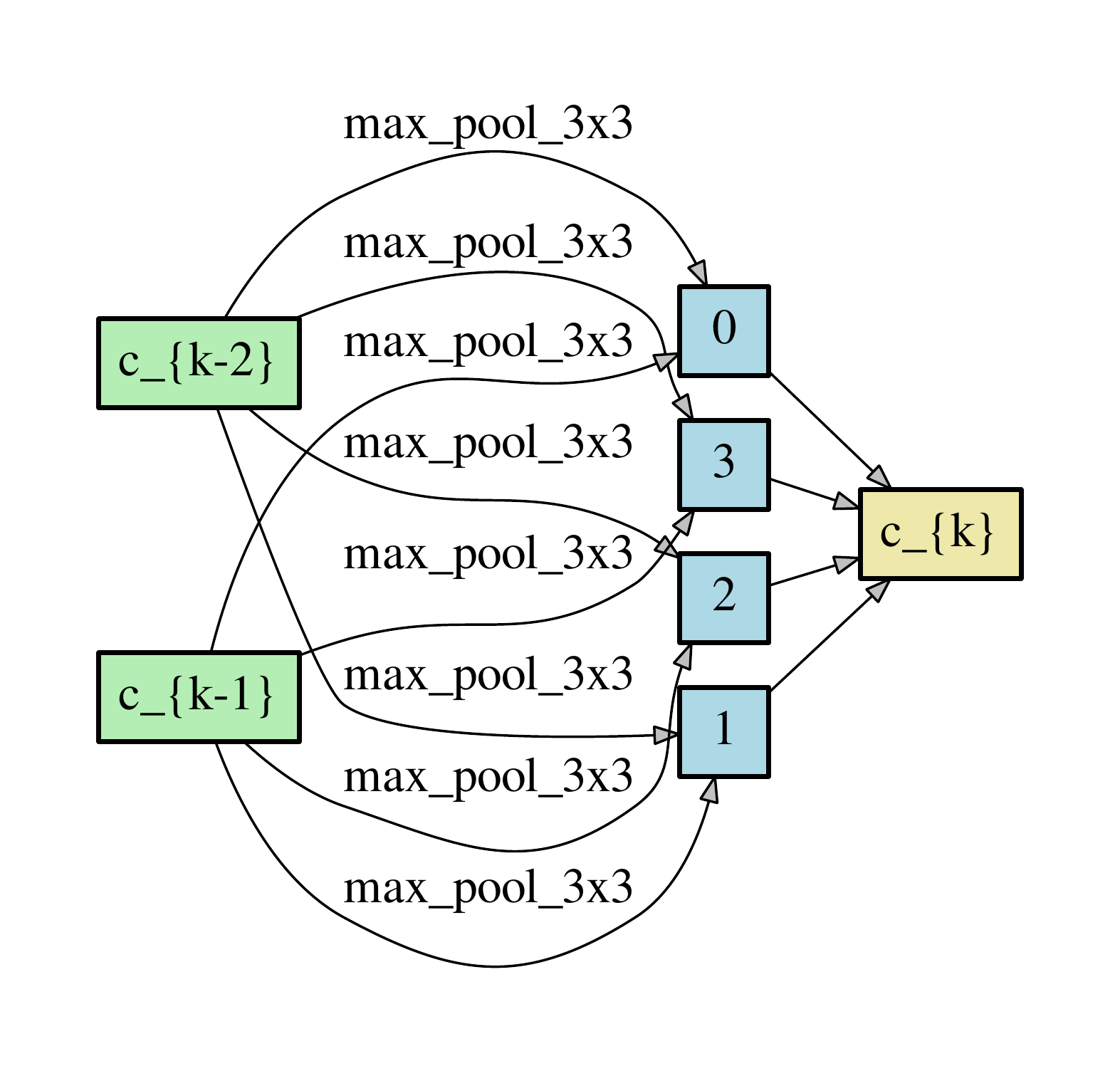}
    \end{minipage}
    }
    \subfigure[Normal cell at epoch 100]{
    \begin{minipage}[h]{0.23\linewidth}
    \centering
      \includegraphics[width = 1\linewidth]{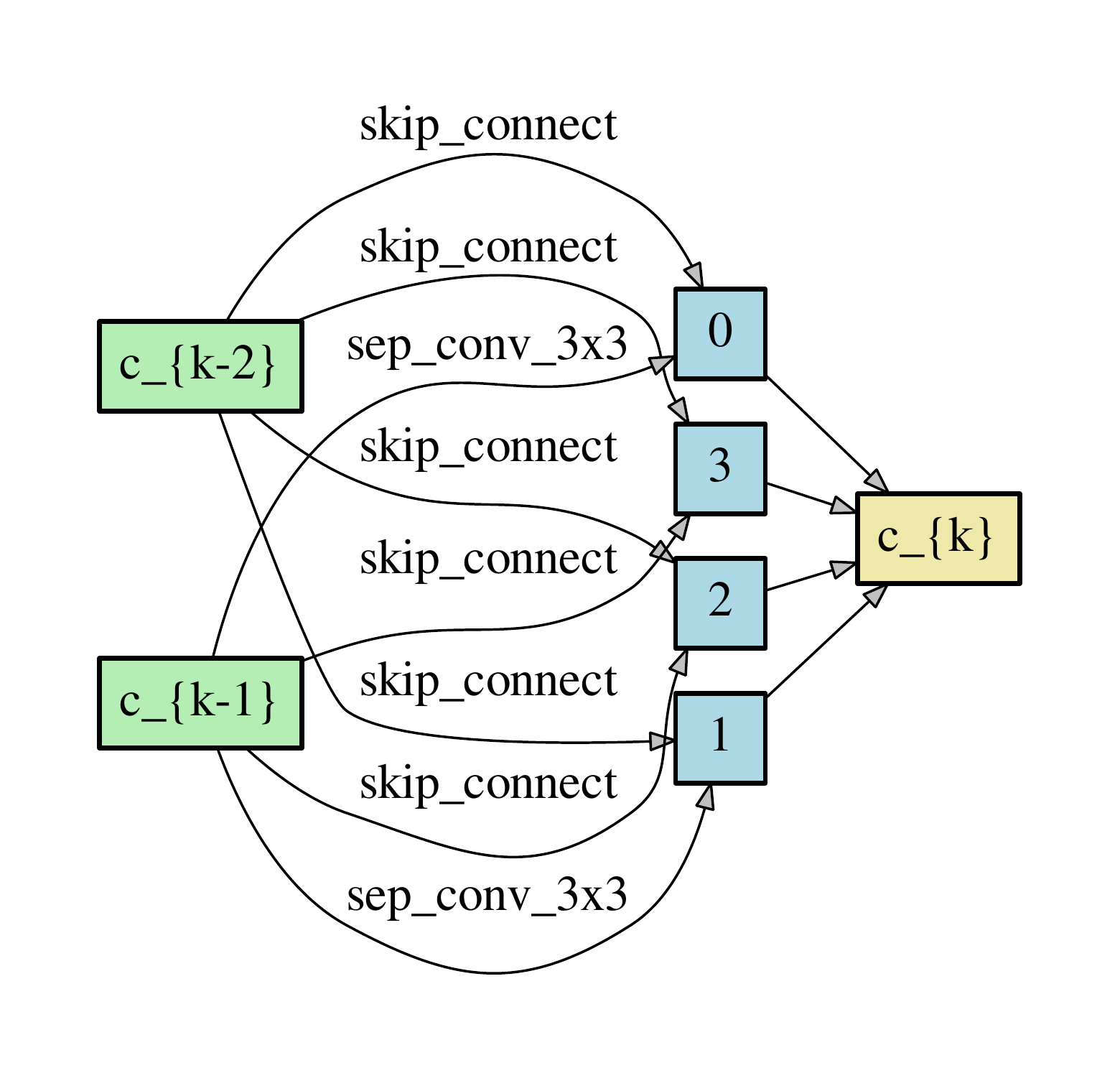}
    \end{minipage}
    }
    \subfigure[Reduction cell at epoch 100]{
    \begin{minipage}[h]{0.23\linewidth}
    \centering
      \includegraphics[width= 1\linewidth]{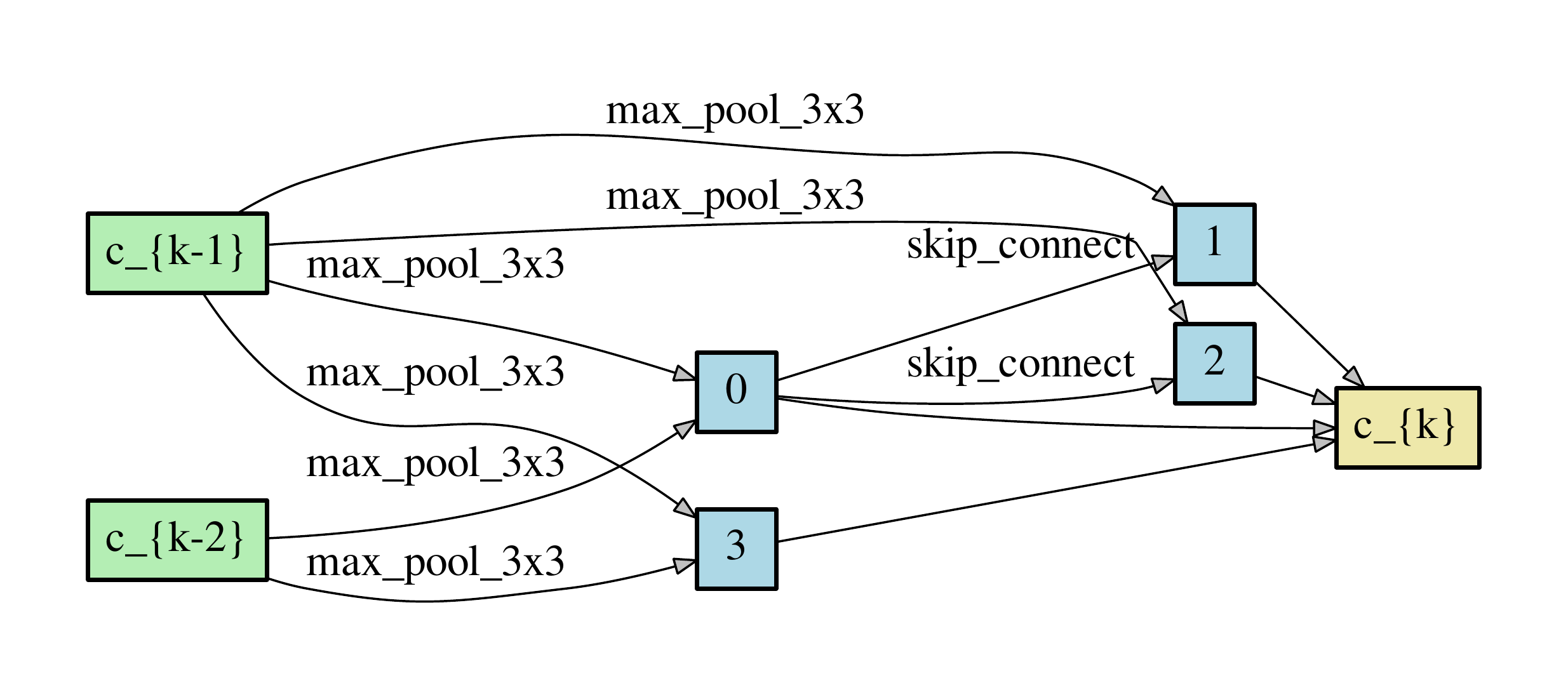}
    \end{minipage}
    }
    \subfigure[Normal cell at epoch 125]{
    \begin{minipage}[h]{0.23\linewidth}
    \centering
      \includegraphics[width = 1\linewidth]{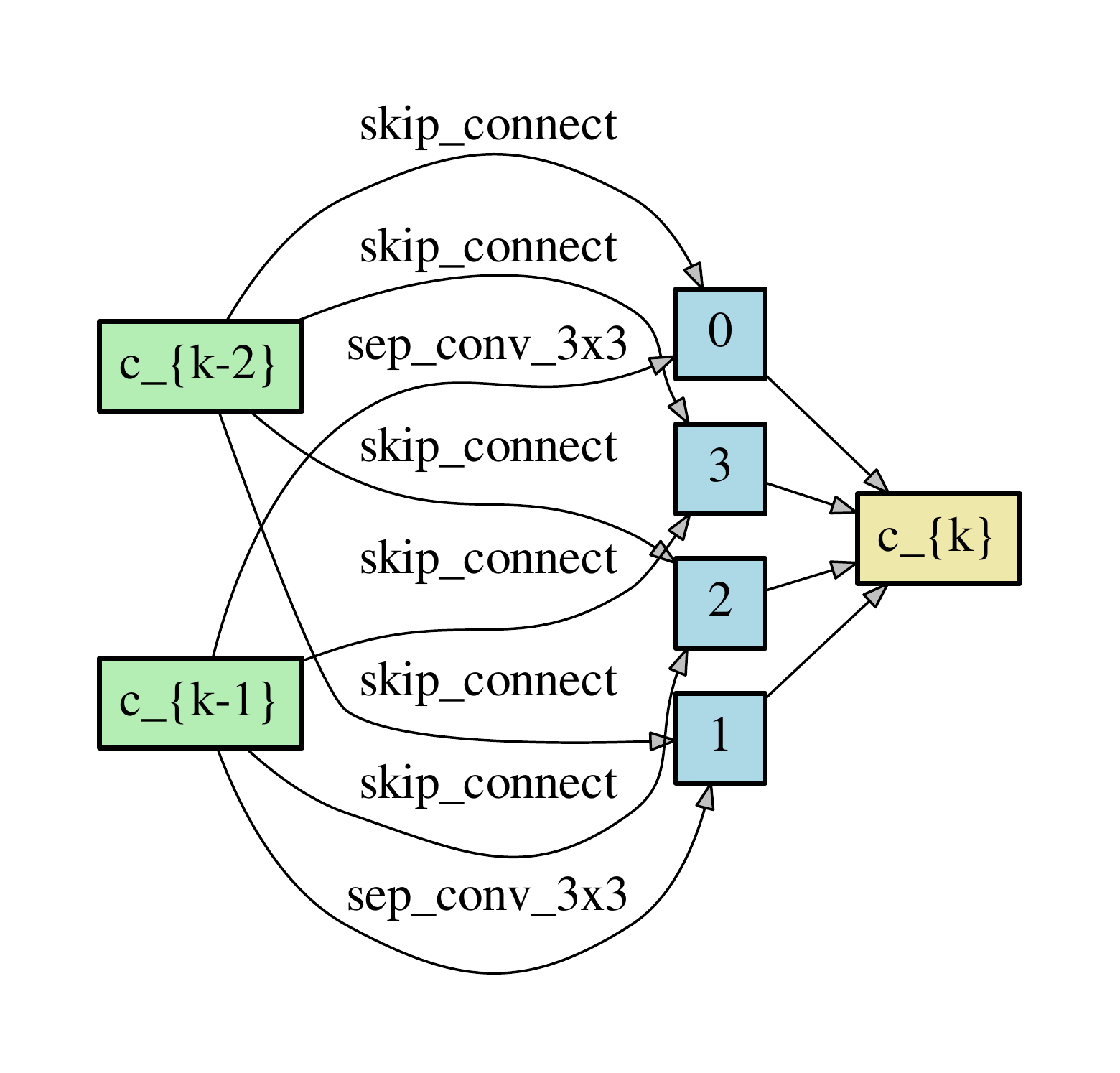}
    \end{minipage}
    }
    \subfigure[Reduction cell at epoch 125]{
    \begin{minipage}[h]{0.23\linewidth}
    \centering
      \includegraphics[width= 1\linewidth]{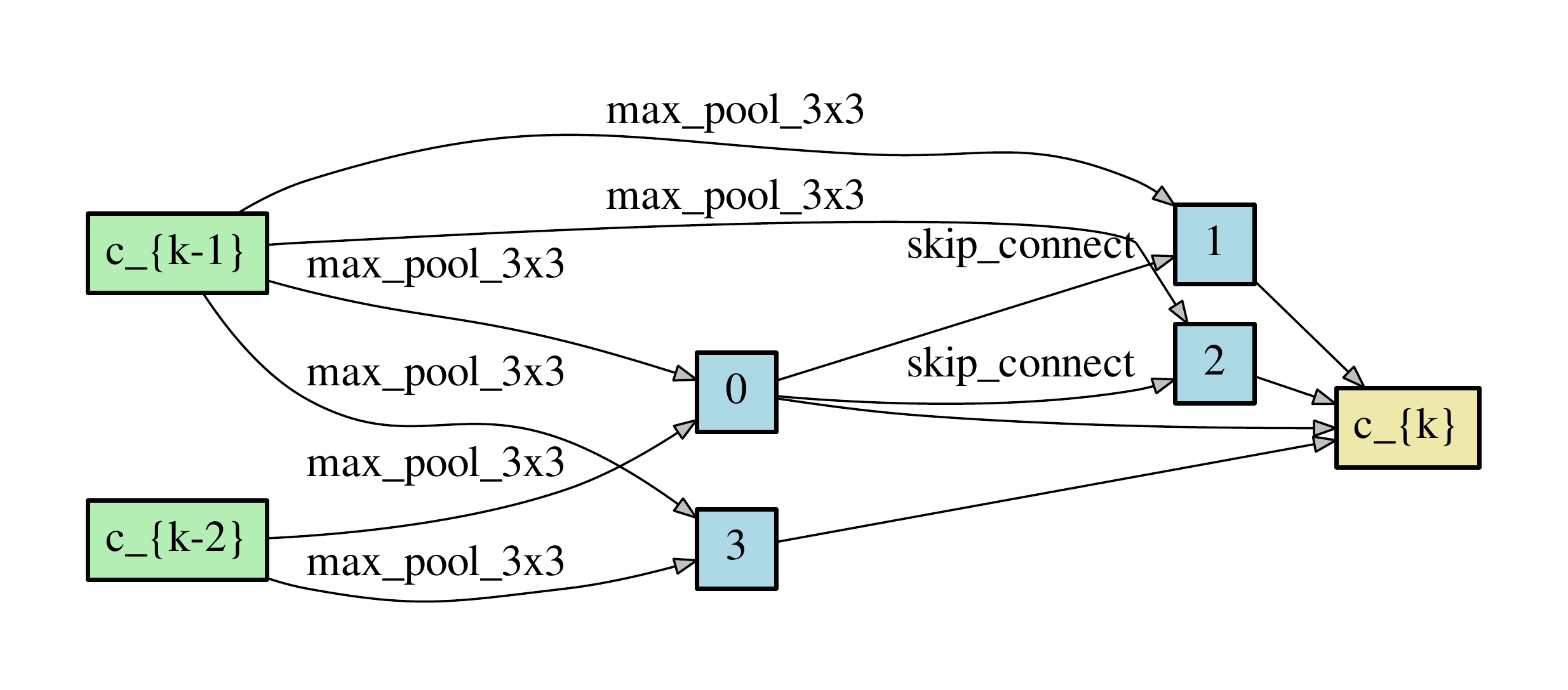}
    \end{minipage}
    }
    \subfigure[Normal cell at epoch 150]{
    \begin{minipage}[h]{0.23\linewidth}
    \centering
      \includegraphics[width = 1\linewidth]{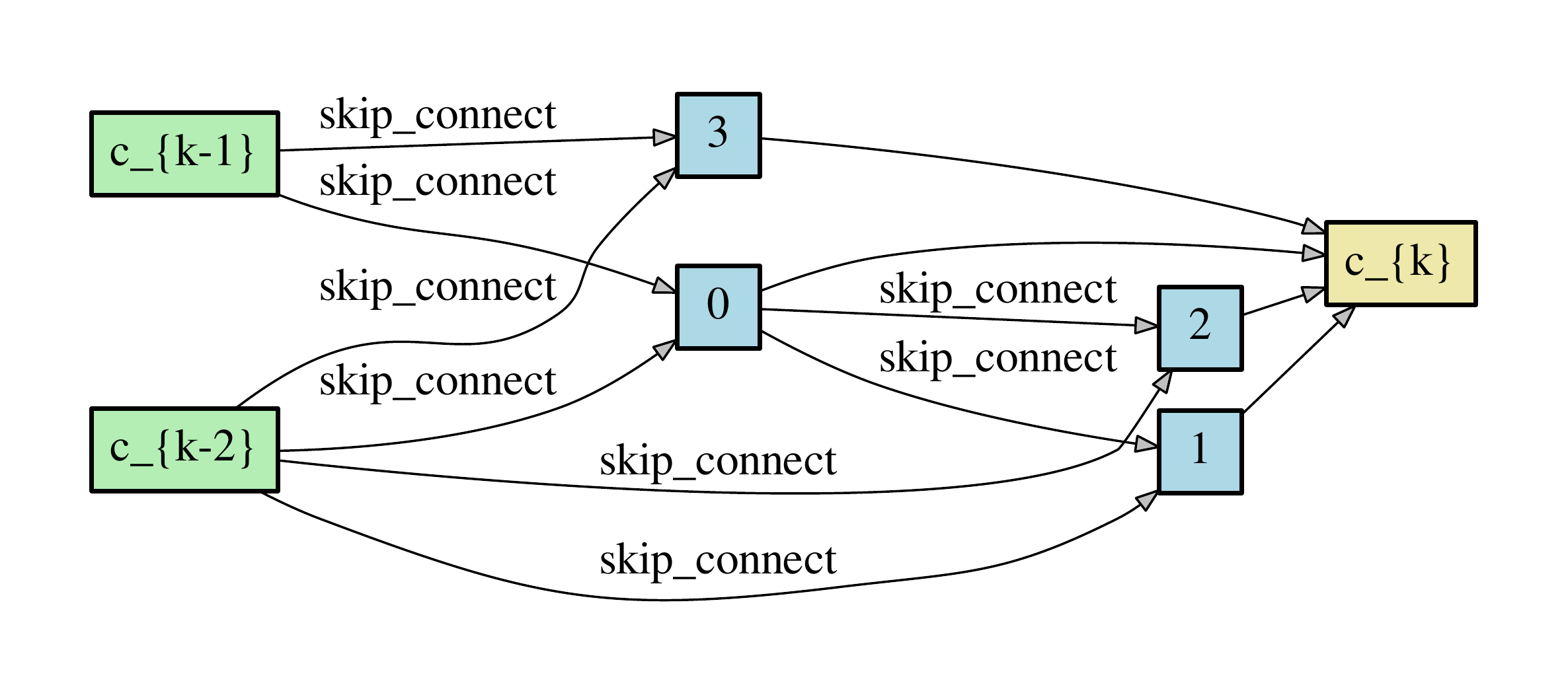}
    \end{minipage}
    }
    \subfigure[Reduction cell at epoch 150]{
    \begin{minipage}[h]{0.23\linewidth}
    \centering
      \includegraphics[width= 1\linewidth]{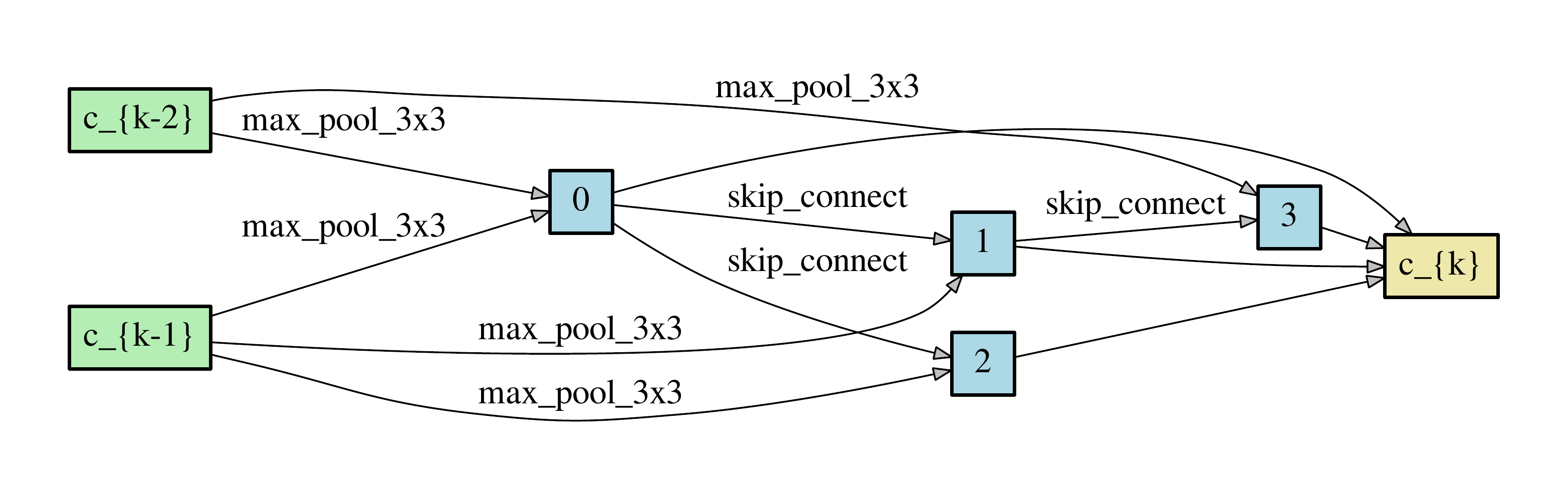}
    \end{minipage}
    }\\
    \subfigure[Normal cell at epoch 175]{
    \begin{minipage}[h]{0.23\linewidth}
    \centering
      \includegraphics[width = 1\linewidth]{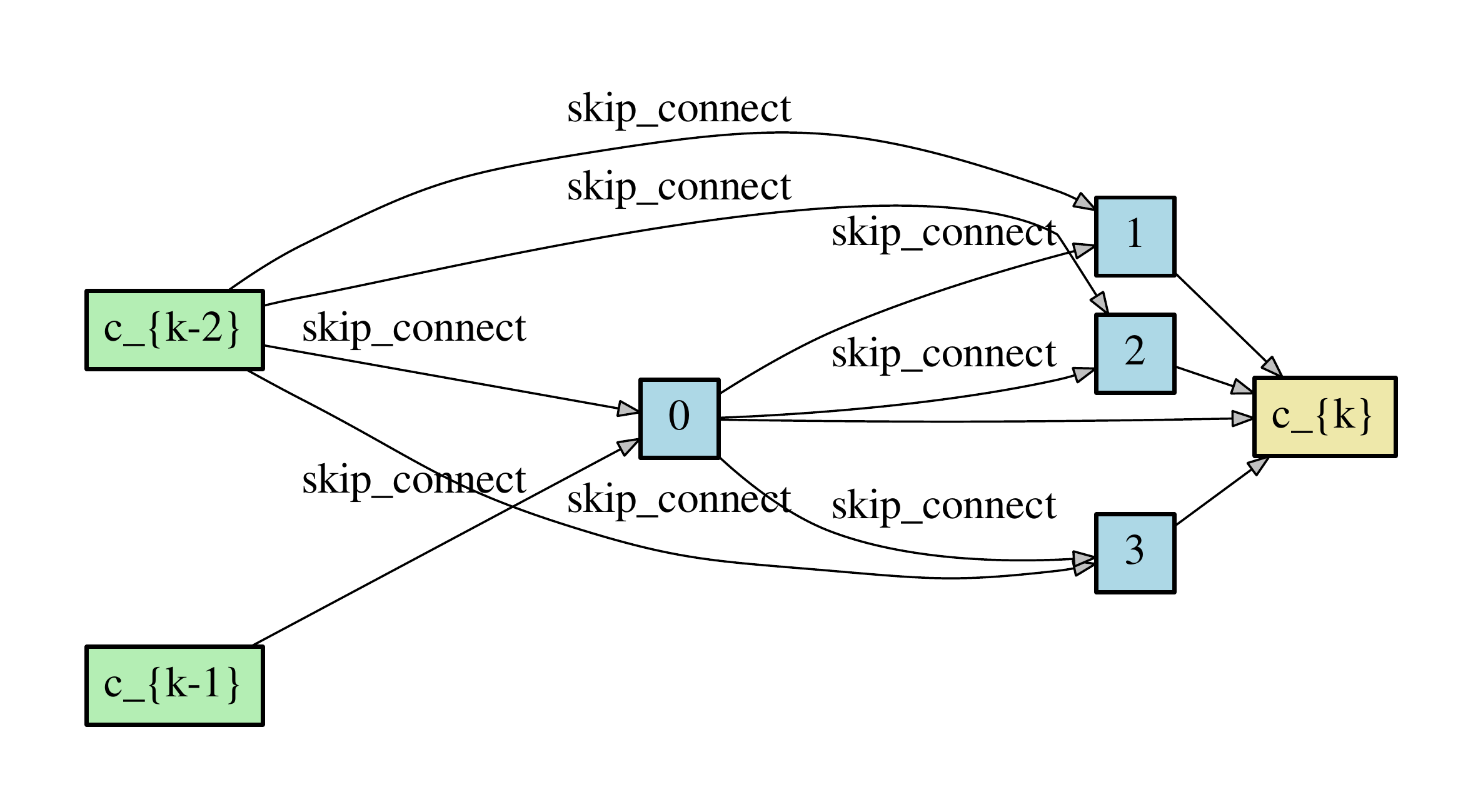}
    \end{minipage}
    }
    \subfigure[Reduction cell at epoch 175]{
    \begin{minipage}[h]{0.23\linewidth}
    \centering
      \includegraphics[width= 1\linewidth]{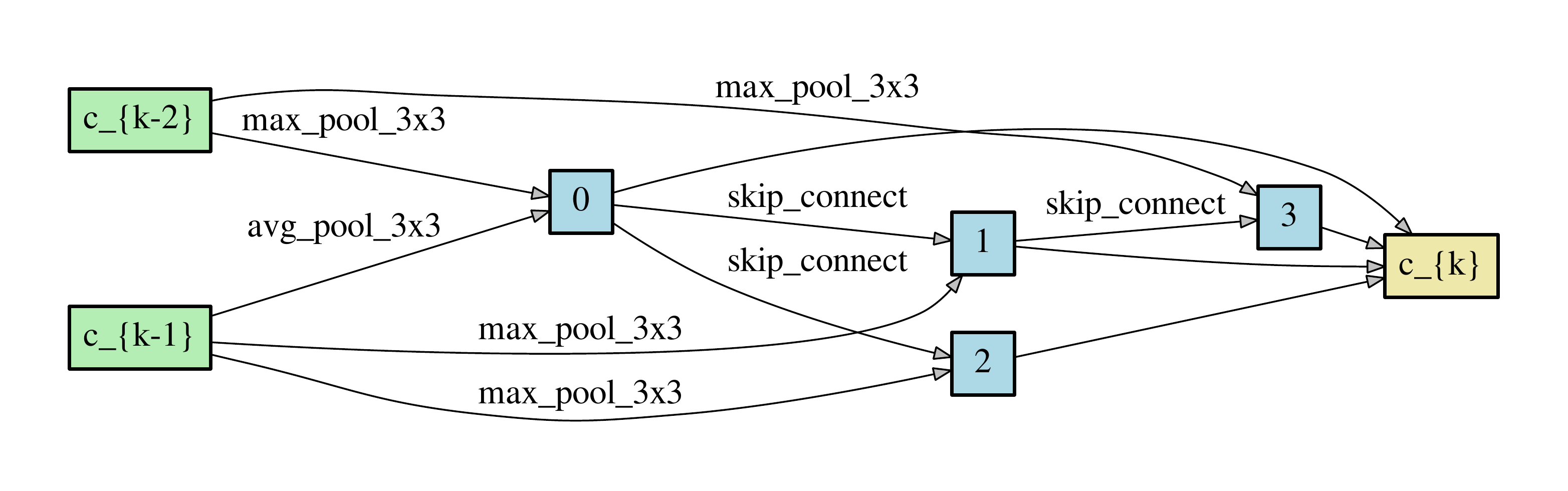}
    \end{minipage}
    }
    \subfigure[Normal cell at epoch 200]{
    \begin{minipage}[h]{0.23\linewidth}
    \centering
      \includegraphics[width = 1\linewidth]{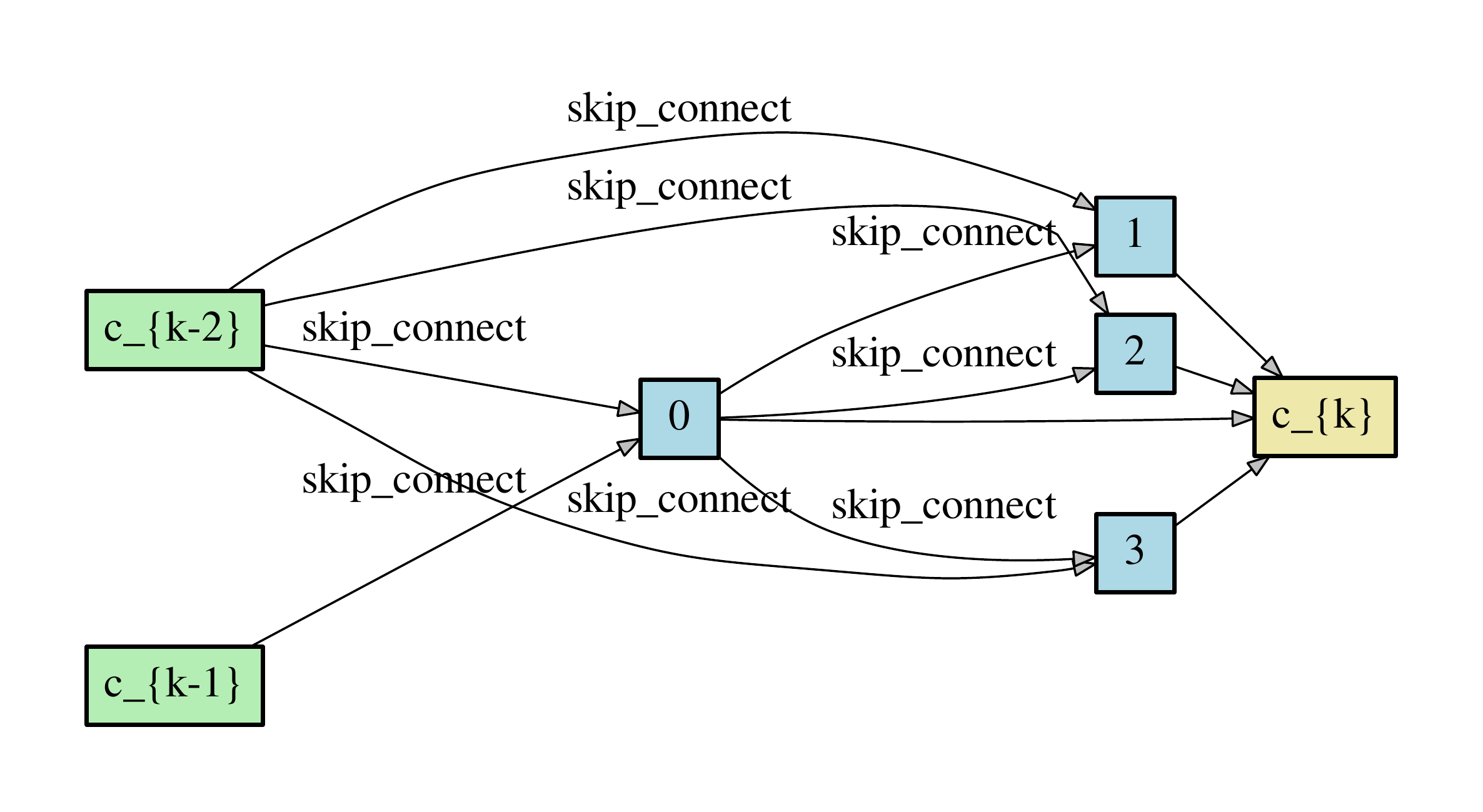}
    \end{minipage}
    }
    \subfigure[Reduction cell at epoch 200]{
    \begin{minipage}[h]{0.23\linewidth}
    \centering
      \includegraphics[width= 1\linewidth]{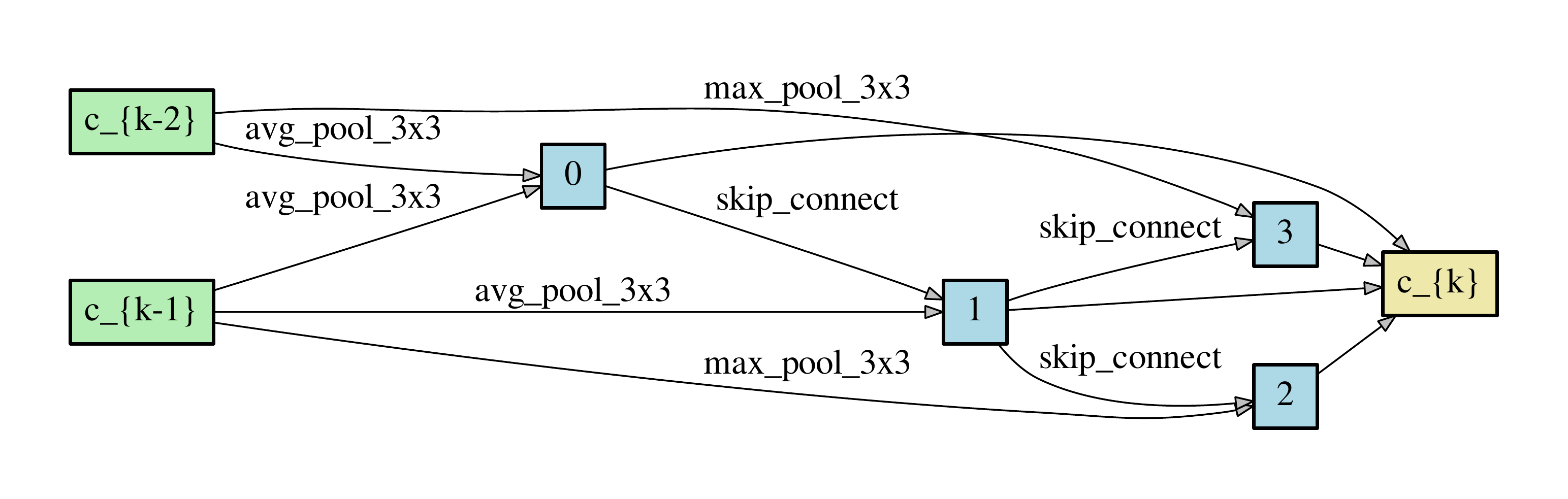}
    \end{minipage}
    }
   \caption{Normal and reduction cells given by DARTS on CIFAR-10 in $S1$ when searching for 200 epochs.}
\label{fig:darts200}
\end{figure*}

\appendix[More Visualization]
More architectures mentioned in Section \ref{EXP.Ablation} are visualized. The architectures in ablation study and robustness validation are illustrated in Fig. \ref{fig:ablation}, Fig. \ref{fig:idarts200} and Fig. \ref{fig:darts200} respectively. NN denotes node normalization and DD denotes decorrelation discretization.

%% insert where needed to balance the two columns on the last page with
%%\newpage


%\vfill


\end{document}